\documentclass[10pt,twocolumn,letterpaper]{article}

\usepackage{cvpr}              

\definecolor{cvprblue}{rgb}{0.21,0.49,0.74}
\definecolor{ForestGreen}{rgb}{0.13, 0.65, 0.23}
\newcommand{\antoine}[1]{\textcolor{black}{#1}}
\newcommand{\AMsup}[1]{\textcolor{black}{#1}}
\definecolor{brickred}{rgb}{0.8, 0.25, 0.33}
\newcommand{\david}[1]{\textcolor{black}{#1}}

\usepackage[pagebackref,breaklinks,colorlinks,allcolors=cvprblue]{hyperref}
\usepackage{textcomp}
\usepackage{colortbl}
\usepackage{array,multirow}
\usepackage{booktabs}
\usepackage{xcolor}
\usepackage{tikz}
\usepackage{pifont}
\usepackage{bbold}
\usepackage{bm}
\usepackage{threeparttable}
\usepackage{booktabs}
\usepackage{tikz}
\usepackage{graphicx}
\usepackage{caption}
\usepackage{pgfplots}
\pgfplotsset{compat=1.18}
\usepackage{tcolorbox} 
\usepackage{multicol}
\usepackage{algorithm}
\usepackage{algpseudocode}
\usepackage{placeins}
\usepackage{afterpage}
\newcommand*{\affaddr}[1]{#1} 
\newcommand*{\affmark}[1][*]{\textsuperscript{#1}}
\newcommand*{\email}[1]{\texttt{#1}}

\def\paperID{13904} 
\def\confName{CVPR}
\def\confYear{2025}

\title{Improved monocular depth prediction using distance transform over pre-semantic contours with self-supervised neural networks}
\author{%
Marwane Hariat\affmark[1], Antoine Manzanera\affmark[1], David Filliat \affmark[1]\\
\affaddr{\affmark[1]U2IS, ENSTA, Institut Polytechnique de Paris, Palaiseau, France}\\
\email{\{marwane.hariat, antoine.manzanera, david.filliat\}@ensta.fr}\\
}

\begin{document}
\twocolumn[{
\maketitle
\begin{center}
    \captionsetup{type=figure}
    \captionsetup[subfigure]{labelformat=empty}
    \addtocounter{figure}{-1}
    
    \begin{subfigure}[b]{0.23\textwidth}
        \centering
        \includegraphics[width=\textwidth, height=1.cm]{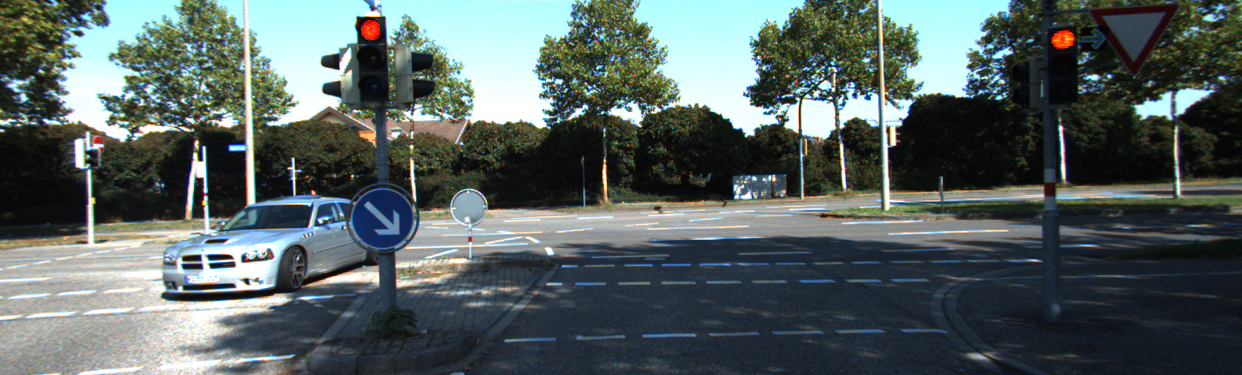}
    \end{subfigure}
    \hfill
    \begin{subfigure}[b]{0.23\textwidth}
        \centering
        \includegraphics[width=\textwidth, height=1.cm]{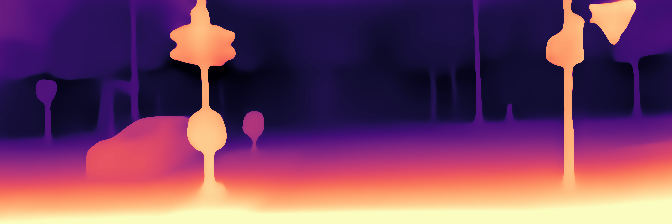}
    \end{subfigure}
    \hfill
    \begin{subfigure}[b]{0.23\textwidth}
        \centering
        \includegraphics[width=\textwidth, height=1.cm]{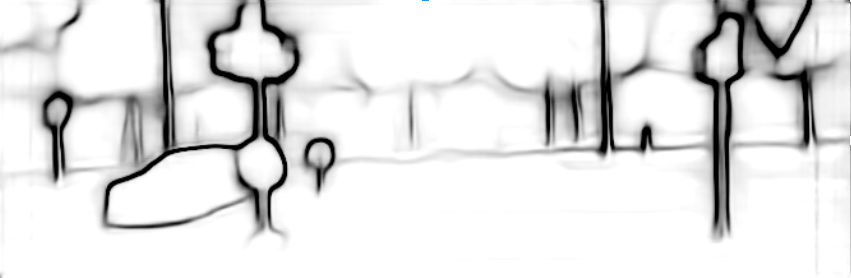}
    \end{subfigure}
    \hfill
    \begin{subfigure}[b]{0.23\textwidth}
        \centering
        \includegraphics[width=\textwidth, height=1.cm]{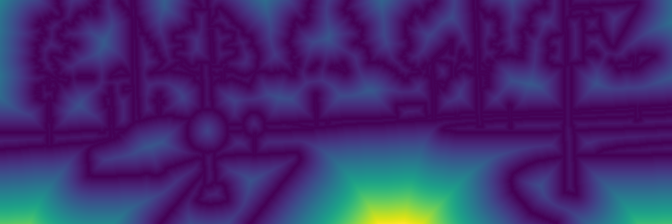}
    \end{subfigure}

    \vspace{0.3cm}  

    \begin{subfigure}[b]{0.23\textwidth}
        \centering
        \includegraphics[width=\textwidth, height=1.cm]{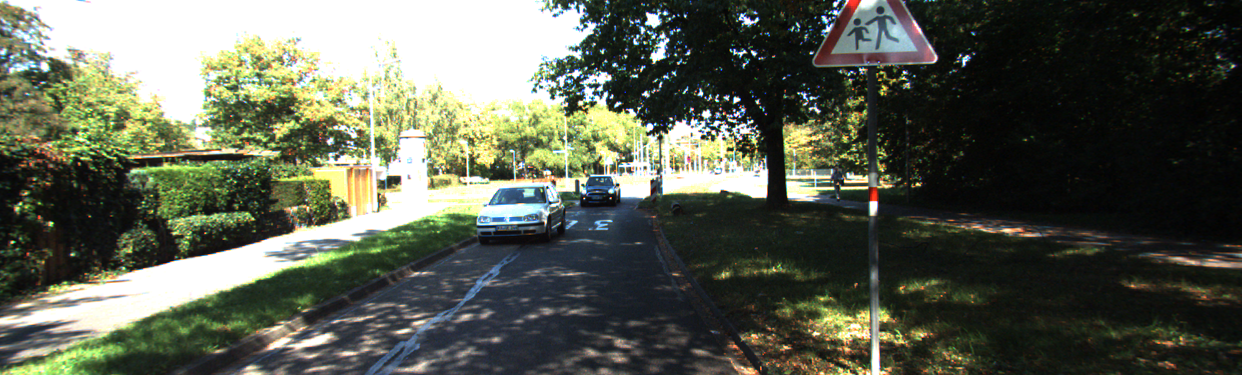}
        \label{figteaser:Orginalimage-anom}
    \end{subfigure}
    \hfill
    \begin{subfigure}[b]{0.23\textwidth}
        \centering
        \includegraphics[width=\textwidth, height=1.cm]{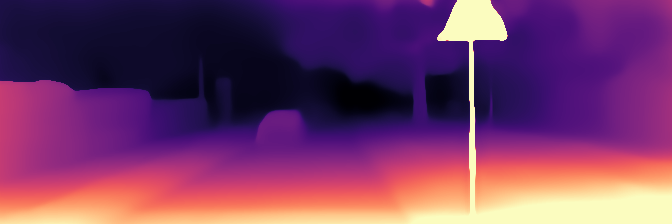}
        \label{figteaser:RobustaGeneration-anom}
    \end{subfigure}
    \hfill
    \begin{subfigure}[b]{0.23\textwidth}
        \centering
        \includegraphics[width=\textwidth, height=1.cm]{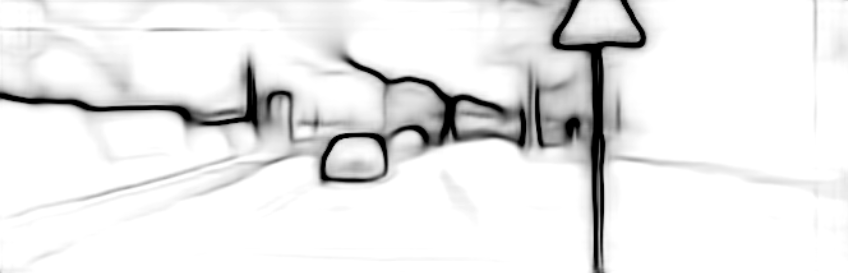}
        \label{figteaser:OasisGeneration-anom}
    \end{subfigure}
    \hfill
    \begin{subfigure}[b]{0.23\textwidth}
        \centering
        \includegraphics[width=\textwidth, height=1.cm]{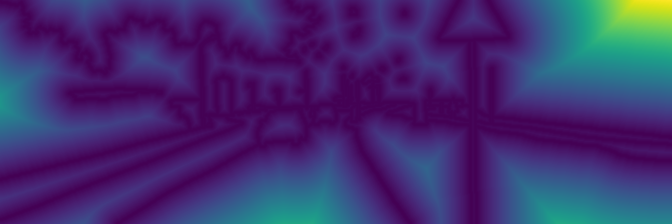}
        \label{figteaser:Label}
    \end{subfigure}

    \vspace{-0.1cm}  

    \begin{subfigure}[b]{0.23\textwidth}
        \centering
        \includegraphics[width=\textwidth, height=1.cm]{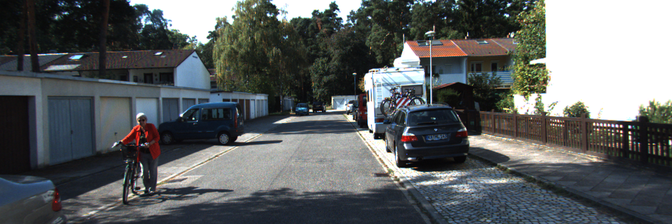}
        \caption{RGB images}
        \label{figteaser:Orginalimage-anom}
    \end{subfigure}
    \hfill
    \begin{subfigure}[b]{0.23\textwidth}
        \centering
        \includegraphics[width=\textwidth, height=1.cm]{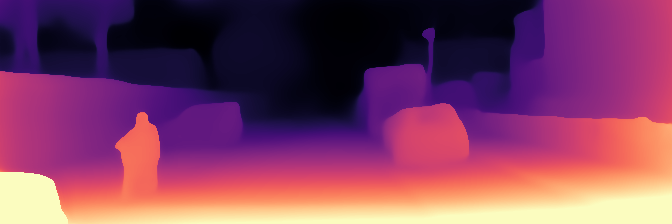}
        \caption{Depth}
        \label{figteaser:RobustaGeneration-anom}
    \end{subfigure}
    \hfill
    \begin{subfigure}[b]{0.23\textwidth}
        \centering
        \includegraphics[width=\textwidth, height=1.cm]{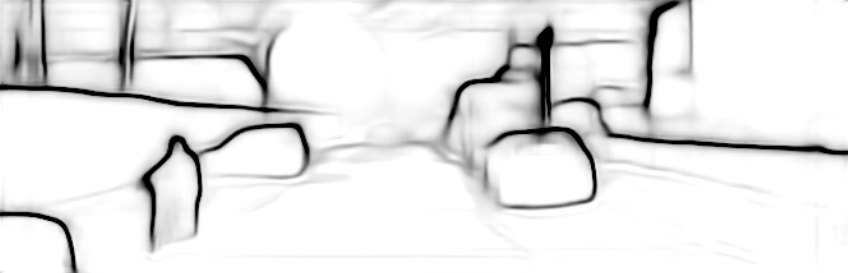}
        \caption{Edge}
        \label{figteaser:OasisGeneration-anom}
    \end{subfigure}
    \hfill
    \begin{subfigure}[b]{0.23\textwidth}
        \centering
        \includegraphics[width=\textwidth, height=1.cm]{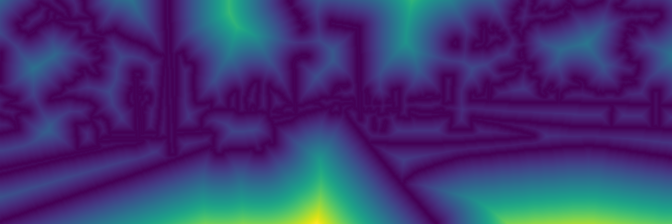}
        \caption{Distance Transform}
        \label{figteaser:Label}
    \end{subfigure}
    
    \captionof{figure}{The different modalities estimated in our self-supervised framework and used to reduce matching uncertainty between images.}
    \label{fig:qualitative_robustness}
\end{center}
}]
\maketitle

\begin{abstract}
Monocular depth estimation (MDE) with self-supervised training approaches struggles in low-texture areas, where photometric losses may lead to ambiguous depth predictions. To address this, we propose a novel technique that enhances spatial information by applying a distance transform over pre-semantic contours, augmenting discriminative power in  low texture regions. 
Our approach jointly estimates pre-semantic contours, depth and ego-motion. The pre-semantic contours are leveraged to produce new input images, with variance augmented by the distance transform in uniform areas. This approach results in more effective loss functions, enhancing the training process for depth and ego-motion.  We demonstrate theoretically that the distance transform is the optimal variance-augmenting technique in this context. Through extensive experiments on KITTI, Cityscapes, Waymo, NYUv2 and ScanNet our model demonstrates robust performance, surpassing competing self-supervised methods in MDE.
\end{abstract}
\section{Introduction}
\label{sec:intro}

The goal of monocular depth estimation (MDE) is 
to determine the distance of a pixel from the camera focal plane using only a single input image, unlike stereo depth estimation \cite{ha2016high, badki2020bi3d, godard2017unsupervised}, which relies on pairs of images. This eliminates the need for expensive sensors that require high-power resources \cite{li2020rtm3d, lu2021sgtbn}, such as active depth camera or LiDAR, while maintaining low latency and ensuring real-time compliance \cite{wofk2019fastdepth, elkerdawy2019lightweight}. 
\antoine{This makes MDE}
a core module in applications such as robot navigation 
\cite{shenoi2020jrmot, yuan2020sad}, 
autonomous driving 
\cite{natan2022end, wang2019pseudo}
or virtual/augmented reality 
\cite{el2018survey, rasla2022relative}.
\antoine{Deep Neural Networks (DNNs) are currently the state of the art for depth maps prediction thanks to their ability to exploit all}
contextual cues of the environment \cite{hu2019visualization, dijk2019neural}.\\
\noindent
Supervised methods for MDE rely on ground-truth annotations. 
\antoine{In spite of}
huge progress \cite{eigen2014depth, fu2018deep, teed2018deepv2d, chen2020improving}, 
\antoine{these methods}
are constrained by the cost and quality of the annotations. In this paper, we focus on self-supervised 
\antoine{MDE, that uses}
the structure-from-motion framework \cite{garg2016unsupervised, zhou2017unsupervised} and leverages depth and camera ego-motion estimations from DNNs to reconstruct the next frame from the current one. The supervision signal is then provided by
\antoine{the photometric loss, i.e. the difference between the reconstructed frame and the observed one, based on a combination of direct colour comparison and}
Structural Similarity Index Measure (SSIM) \cite{nilsson2020understanding}. 
Although more challenging, the self-supervised paradigm offers greater training flexibility. First, it can be trained using data from unknown cameras \cite{gordon2019depth}, as long as data acquisition provides successive frames, such as in YouTube videos \cite{chen2019self}. This is particularly relevant given the growing interest in foundation models, which can now leverage large datasets to produce generalizable outputs \cite{yang2024depth}. In addition, it is compatible with continual learning, allowing the model to continuously expand its knowledge as more data become available \cite{chen2019self, kuznietsov2021comoda}.\\
\noindent
However, the problem of MDE using the 
\antoine{photometric loss}
proposed by Zhou et al. \cite{zhou2017unsupervised} is ill-posed due to its reliance on SSIM and colour-based cost functions, which fail to discriminate effectively in low-texture areas. In these regions, multiple candidates can yield similarly low cost function values, making it difficult to determine the correct correspondence. Although interesting work has been done to address various issues such as scale ambiguity \cite{wang2018learning}, gradient locality \cite{jiang2019linearized, godard2019digging}, occlusion \cite{wang2018occlusion, godard2019digging}, moving objects \cite{ranjan2019competitive, li2021unsupervised}, and infinite depth holes \cite{guizilini2020semantically, godard2019digging}, the ill-posed nature of this problem has not been thoroughly and rigorously studied. We 
\antoine{aim}
to provide this essential foundation.\\
\noindent
\antoine{The}
key challenge lies in reducing ambiguity to build a more robust foundation for learning depth and ego-motion. To this end, we propose introducing sufficient variance in texture-less areas, while adhering  as closely as possible to \textquotesingle the constancy assumption\textquotesingle: any change introduced in image \(t\) should be reproduced identically in image \(t+1\),
\antoine{essential as the photometric loss}
relies on correspondences between successive frames. 
We propose to do this by assigning to each pixel the distance to its nearest 
\antoine{edge in the image},
commonly known as the \textquotesingle distance transform\textquotesingle\ \cite{strutz2021distance, niblack1992generating}. We demonstrate mathematically that the distance transform is the most effective way of adding variance while respecting the constancy assumption, and that it also leads to improved convergence properties. 
\antoine{Since the distance transform is calculated on binary contours, we propose a self-supervised pre-semantic contour estimation method, jointly learned with the depth estimation model,} and add this distance map as an additional channel to the input image 
to produce a \textquotesingle variance-augmented\textquotesingle\ image for reconstruction.
\\
In summary, our contributions are as follows.
\begin{itemize}
    \item \textbf{(1)} A self-supervised approach for 
    \antoine{pre-semantic}
    contour estimation, which 
    \antoine{(i)}
    generates the edge information necessary for computing the distance transform, and 
    \antoine{(ii)}
    simultaneously reinforces the depth estimation network.
    \item \textbf{(2)} A novel variance-augmented image that incorporates the distance transform, enhancing discriminative power in texture-less areas and improving convergence properties
    \item \textbf{(3)} A 
    theoretical foundation demonstrating the effectiveness of the distance transform for adding variance under the constancy assumption
    \item \textbf{(4)} An extensive experimental validation of our new pipeline, achieving competitive performance on 
    \antoine{different}
    benchmark datasets: KITTI \cite{geiger2013vision}, Cityscapes \cite{cordts2016cityscapes}, Waymo \cite{sun2020scalability}, NYUv2 \cite{silberman2012indoor} and ScanNet \cite{dai2017scannet}
\\
\end{itemize}
\section{Related works}
\label{sec:realated_works}

\textbf{Self-supervised framework.} The work of Zhou et al. \cite{zhou2017unsupervised} is 
pioneer in self-supervised monocular depth estimation. Their method utilizes depth and ego-motion to find the corresponding pixels between two successive frames and relies on the bilinear sampler \cite{jaderberg2015spatial} to reconstruct the upcoming scene from the current frame. The supervision signal is then derived from the photometric loss, which is a weighted sum of the SSIM and colour cost functions.
The effectiveness of this 
method relies on the constancy assumption, which asserts that 
\antoine{object colours remain}
consistent from frame to frame.
\antoine{This assumption is bound to the hypothesis of constant lighting over Lambertian surfaces, and do not hold in the case of occlusions and moving objects.}
This method has
\antoine{then}
been refined in several ways. Kim et al. \cite{kim2021revisiting} provide a comprehensive study of the main issues and solutions found in the literature. In \cite{wang2018occlusion}, occlusions are addressed using a forward warping module that counts the number of bilinear sampling operations for each pixel. Godard et al. \cite{godard2019digging} compute the occlusion mask by determining the per-pixel minimum of the photometric loss over time. Moving objects are identified through the differences between optical flow and depth/ego-motion in \cite{ranjan2019competitive, liu2019unsupervised, lee2019learning, hariat2023rebalancing, yin2018geonet, zou2018df}, while off-the-shelf semantic algorithms are utilized to localize potential moving objects in \cite{casser2019depth, guizilini2020semantically}. In addition, a residual map is introduced to correct the ego-motion in \cite{li2021unsupervised, gordon2019depth}. Infinite depth holes are filtered through a two-stage training process \cite{guizilini2020semantically} and are removed using an adaptive cost-volume architecture in \cite{watson2021temporal}. Finally, Godard et al. \cite{godard2019digging} propose a unified solution that effectively addresses all the aforementioned issues. However, the ill-posed nature of the problem
\antoine{due to unstructured objects}
remains unaddressed.
\\
\textbf{Contour estimation.} 
Semantic contours 
\antoine{should follow}
the boundaries derived from instance segmentation, outlining distinct entities within an image. 
\antoine{Such contours would naturally help improve the accuracy of depth maps.}
\cite{wang2016surge} and \cite{ramamonjisoa2019sharpnet} leverage ground-truth annotations to learn depth, normals, and contours, incorporating regularization terms to enforce constraints derived from their geometric relationships. In contrast, \cite{li2024devil} employs an off-the-shelf edge prediction algorithm to refine initial depth estimations as a post-processing step.
\cite{xian2020structure}, \cite{saeedan2021boosting} and \cite{li2023learning} utilize a pre-trained instance segmentation network to compute edges, sampling points on both sides of the boundary to enhance depth sharpness around those edges. \cite{zhu2020edge} introduced a morphing strategy to align depth borders computed with a threshold gradient on the depth with semantic borders obtained from a pre-trained semantic algorithm. While these techniques demonstrate significant improvements, they assume the edges as 
\antoine{provided}
and are therefore not fully self-supervised, in addition to not being GPU friendly.
Significant advances have been made in supervised edge map estimation \cite{xie2015holistically, deng2018learning}; however, the self-supervised approach remains an ambitious and challenging task. \cite{li2016unsupervised} introduces a reinforcement training procedure in which an optical flow network 
\antoine{is trained to predict}
the displacement
\antoine{of each pixel}
between two frames, 
\antoine{based on ground truth}
point correspondences.
Subsequently, an edge detector is updated to align its predicted edges of the image with the predicted edges of the estimated flow map. Still, this method 
\antoine{is not}
self-supervised. 
A 
work 
closer
to ours \cite{yang2018lego} proposes a fully self-supervised edge prediction framework that utilizes the positive part of the second derivatives of depth and the first-order gradient of normal to surface to feed an edge network. However, this approach employs suboptimal loss functions, leading to biased contours that favour shorter distances. Furthermore, it does not utilize the edges to enhance depth, nor does it include any post-processing to create refined contours. We propose new cost functions for training the edge network, resulting in significant performance gains. Additionally, we introduce a post-processing step to refine contours and implement an extended edge-aware diffusion smoothness loss \cite{heinrich2011improved} that encourages sharp depth discontinuities around estimated boundaries, in the same spirit as \cite{xian2020structure, zhu2020edge, li2023learning}.
\\
\textbf{Targeting the ill-posed optimization problem}. As previously mentioned, our objective is to minimize the photometric loss by establishing correspondences between pixels in the current frame and those in the next frame, achieved through the depth and ego-motion parameters. However, this correspondence procedure is not unique,
\antoine{since many pixels}
share the same colour and exhibit similar local colour statistics as 
\antoine{their neighbourhood}.
This ambiguity is especially pronounced in areas with low texture.
The challenge is to 
\antoine{compensate for}
the lack of variation by introducing 
\antoine{time-consistent}
variance across frames, 
\antoine{to satisfy}
the constancy assumption. This reduces the ill-posed nature of the optimization problem and creates better conditions for learning depth and ego-motion. \cite{zhan2018unsupervised, xu2021self} addressed this issue by 
\antoine{applying the reconstruction loss to deep features in addition to the usual colour images}.
Shu et al. \cite{shu2020feature} took this approach further and designed custom deep features by training an auto-encoder that encourages large and smooth gradient values in feature maps, yielding deep features with both semantic richness and substantial 
\antoine{structure}.
The constancy assumption is rather optimistic for these methods, as demonstrated in our experiments. 
\antoine{Indeed, due}
to their large receptive fields, deep features generally lack consistency from frame to frame. Additionally, the reduced resolution of feature maps relative to the input image significantly limits 
\antoine{the benefit of these methods in terms of accuracy.}
We show that the distance transform, not only 
\antoine{proves to be experimentally}
more consistent 
across frames, 
\antoine{but also stands as the theoretically best method for increasing variance}
under the constancy 
\antoine{constraint, while preserving}
the resolution of the input image.
Our work employs a similar methodology 
\antoine{as}
\cite{hafner2013census}, which theoretically prove the strong invariant properties induced by the census transform used in robust optical flow estimation \cite{meister2018unflow} and validate these properties through experiments.
\cite{chen2018estimating} also uses the distance transform, but as an additional input to infer dense depth from sparse depth. However, using the distance transform of edge maps as a training signal is a novel approach.

\newtheorem{definition}{Definition}
\newtheorem{theorem}{Theorem}
\section{Preamble}
\label{sec:preambule}
Our goal in this section is to propose a mathematical formalism that closely aligns with experimental scenarios to address the following question: How can we introduce sufficient variance in low-texture areas while adhering 
\antoine{as much as possible}
to the constancy assumption? The 
\antoine{purpose is to}
provide a better basis than the usual framework \cite{zhou2017unsupervised} for learning depth and ego-motion.\\
\noindent
We start by introducing some key theoretical properties of the distance transform that we illustrate with toy experiments.
The following definitions and theorems provide the theoretical foundation of our analysis.
\subsection{Maximal variance under constraints}
\textbf{Notations} Let \(\Omega\) be a compact convex set of \(\mathbb{R}^n\) with smooth boundary \(\partial\Omega\). Let \(\mathcal{A}\) be the set of affine transformation. And let \(\mathcal{I}_{\mathcal{A}}\) the set of positive functions invariant by affine transformation (see Definition~\ref{def:definition_2}).\\
\noindent
Here, 
\antoine{$\Omega$}
represents an object within a scene, while 
\antoine{$\mathcal{A}$}
models 
\antoine{a reasonable subset of reprojected 3d transformations likely to affect the objects from frame to frame}.

\label{sec:preambule:max_variance}

\begin{definition}\label{def:definition_1}
\begin{itshape}
The distance transform 
\antoine{$\delta_{\Omega}: \Omega \rightarrow \mathbb{R}^{+}$}
is defined 
as:
\begin{equation}
\antoine{\delta_{\Omega}(x) =~}
    d(x, \partial\Omega) = \inf_{y \in \partial\Omega} \|x - y\|
\end{equation}
The set of points having more than one closest point to the boundary is called the medial axis. By Rademacher’s Theorem \cite{nekvinda1988simple} the distance transform is differentiable almost everywhere (away of the medial axis) and satisfies the Eikonal equation:
\begin{equation}
    \begin{aligned}
        \|\nabla d\| = 1
    \end{aligned}
\end{equation}
\end{itshape}
\end{definition}
\begin{definition}\label{def:definition_2}
\begin{itshape}
Let \(f:\Omega \rightarrow \mathbb{R}^{+}\) a smooth function on \(\Omega\). Then f is said to be invariant by affine transformation if: 
\begin{equation}
    \begin{aligned}
        \antoine{\forall A \in \mathcal{A}, \forall x \in \Omega,~}  f(A(x)) = f(x) 
    \end{aligned}
\end{equation}
\end{itshape}
\end{definition}
Definition~\ref{def:definition_2} restates the constancy assumption mathematically: if an affine transformation \(A\) is applied to a convex shape \(\Omega\), then any changes introduce by \(f\) on \(\Omega\) should manifest identically in \(A(\Omega)\).
\begin{theorem}\label{theo:theorem_1}
\antoine{$\delta_{\Omega}$}
is the unique solution, up to an isomorphism, 
\antoine{to}
the following optimization problem:
\begin{equation}
\begin{aligned}
    \max_{f \in \mathcal{I}_{\mathcal{A}}} \quad & \int_{\Omega} \left(f\antoine{(x)} - \overline{f}\right)^2 \antoine{dx} 
    \quad & \text{s.t. } 
    \|\nabla f\| = 1
\end{aligned}
\end{equation}
\end{theorem}
\antoine{$\overline{f}$ being the average value of $f$ over $\Omega$.}
A proof of Theorem~\ref{theo:theorem_1} is given in Appendix \ref{sec:appendix:theoretical:max_variance}.
This property is particularly significant because it demonstrates that, under the constraint of the Eikonal equation, the optimal method for increasing variance within a compact convex shape while adhering to the constancy assumption is to use the distance transform. For this reason, in our practical experiments, we propose to introduce variance within image instances primarily via the distance transform; further details can be found in Section~\ref{sec:method:variance_augmented}.\\
\noindent
The Eikonal equation constraint acts as a standardization condition, promoting smooth variation. Without this constraint, 
\antoine{partitioning the shape in uniform black and white regions would also maximize the variance}.
Our goal
\antoine{is instead}
to ensure a more even spread of variance throughout the shape, making the variance-augmented image reprojection more informative for better depth and ego-motion learning.\\
\noindent
We also explore carefully relaxing this constraint to investigate well-chosen functions of the distance transform that could allow for greater variance within instances while still maintaining an acceptable level of smoothness:
\begin{equation}
    \begin{aligned}
        \mathcal{F} = \left\{
        \antoine{f: x \mapsto g \circ \delta_{\Omega}(x)} \;\middle|\; g \in \mathcal{C}^{\infty}(\mathbb{R}^+, \mathbb{R}^+), \; \|g\|_{\infty} \leq 1 \right\}
    \end{aligned}
    \label{eq:family_f}
\end{equation}

\noindent
A thorough comparison of depth estimation results across different functions is provided in Appendix~\ref{sec:appendix:discussion_distance:study_g}. The
\david{bound on the norm}
ensures that variance is not artificially amplified by larger values, maintaining consistency and fairness in comparisons.\\
We show that the maximal variance property can be
retrieved with a simple toy experiment in \ref{sec:appendix:theoretical:toy_variance}.

\subsection{Convergence properties}
\label{sec:premabule:convergence_properties}
\textbf{Notations} Let \((I, \tilde{I})\) be two successive images of a camera in movement and \((x,  y)\) a training instance where \(x\) represents a pixel of \(I\) and \(y\) be the corresponding pixel in \(\tilde{I}\).
Let us assume also that all pixels lie inside a convex shape \(\Omega\) defined by a smooth and closed contour \(\partial \Omega\). Our goal is to learn the model \(\Phi(x; \theta)\) to predict the new pixel location  \(\hat{y}\).\\
Let us define the cost function $l$ using functions from $\mathcal{F}$ (Equation \ref{eq:family_f}):
\begin{equation}
\begin{aligned}
    l : (\hat{y}, y) & 
    \antoine{ ~\mapsto \left( f(\hat{y}) - f(y) \right)^2}
    \\
    \text{with} \quad & f \in \mathcal{F}
\end{aligned}
\label{eq:loss_distance}
\end{equation}

\begin{definition}\label{def:definition_3}
\begin{itshape}
A function \(f: \Omega \rightarrow \mathbb{R}^+\) is \(\alpha\)-Lipschitz if:
\begin{equation}
    \begin{aligned}
       \forall u, v \quad |f(u) - f(v)| \leq \alpha \, \|u - v\|    
    \end{aligned}
\end{equation}
\end{itshape}
\end{definition}

\begin{definition}\label{def:definition_4}
\begin{itshape}
A function \(f: \Omega \rightarrow \mathbb{R}^+\) is \(\beta\)-smooth if:
\begin{equation}
    \begin{aligned}
      \forall u, v \quad  \|\nabla f(u) - \nabla f(v)\| \leq \beta \|u - v\|      
    \end{aligned}
\end{equation}
\end{itshape}
\end{definition}
\noindent
\david{The fact that $f$ is a function of the distance transform makes it possible to prove:}
\begin{theorem}\label{theo:theorem_2}
The loss function \(l\) is, with respect to the first argument \(\hat{y}\):
\begin{itemize}
    \item \(\alpha\)-Lipschitz
    \item \(\beta\)-smooth
    \item strongly convex if a regularization term  \( \eta \|y\|^2  \)  is added.
\end{itemize}
\end{theorem}
The constants \david{$\alpha, \beta, \eta$} depend on the curvature at the projection onto the boundary  \david{\(\partial \Omega\)}. A proof of Theorem~\ref{theo:theorem_2} is given in Appendix~\ref{sec:appendix:theoretical:convergence}. Those properties \david{brought by the distance transform} induce great convergence properties as described in \cite{akbari2021does, hardt2016train, kawaguchi2021recipe}. Amongst others: a bounded generalization error and a uniform stability of the stochastic gradient descent. \david{Note that the} SSIM + colour function can also be Lipschitz and/or uniform under very strict conditions. However their constants, unlike with the distance transform, depend on the size of the shape \(\Omega\) and can thus exceeds all bounds as mentioned in \cite{otero2021optimization}.\\
We show that the convergence property can be
retrieved with a simple toy experiment in \ref{sec:appendix:theoretical:toy_convergence}.

\section{Method}
\label{sec:method}
\begin{figure*}[htb]  
    \centering
     \includegraphics[width=1.0\textwidth]{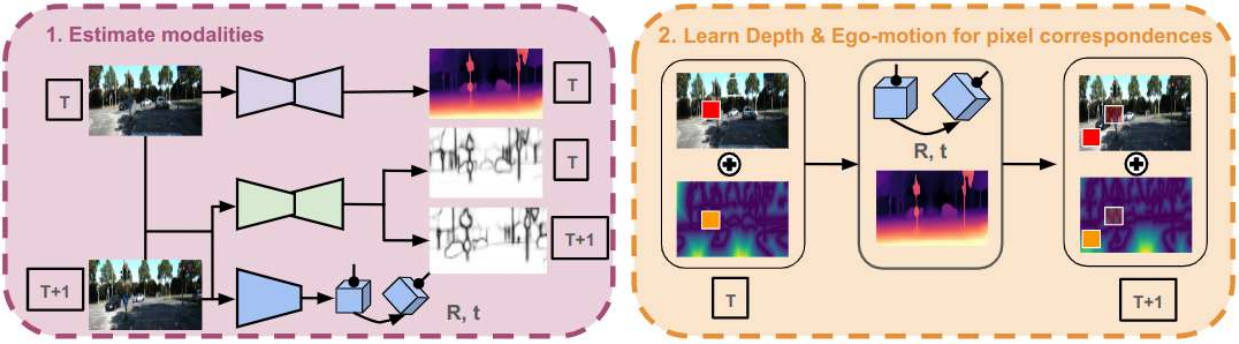}
    \caption{Diagram showing the two main steps of our pipeline: estimation of the different modalities followed by the re-projection on the variance augmented image using depth and ego-motion. T indicates the time frame. \(\text{R , t}\) respectively rotation and translation predictions.}
    \label{fig:graphic}
\end{figure*}

Our framework \david{for self-supervised MDE} (Figure~\ref{fig:graphic}) consists of three networks: Depth \(D_{\theta}\), Ego-motion \(T_{\alpha}\) and edge \(E_{\delta}\). For the sake of readability, we will omit the parameter symbols \(\theta, \alpha, \delta\) in the remainder of the section. Hyper-parameters in the loss functions will also be omitted here and detailed in Appendix~\ref{sec:appendinx:hyper_parameters}.

\subsection{Pre-semantic contours}

Our edge network renders values \( E(p) \) in the range \( [0, 1] \), where a higher value indicates a greater likelihood that the pixel \(p\) corresponds to an edge. To generate pseudo-labels for supervising the edge network, we utilize depth and surface normal estimation, which is derived from depth following the methods described in \cite{yang1711unsupervised, yin2019enforcing}. We discuss complementary insights of depth and surface normal for edge detection in Appendix~\ref{sec:appendix:discussion_contour:complementary_depth_normal}.\\
\noindent
\textbf{Depth \& Normal Pseudo-Labels:} Depth and surface normals are smooth and thus prone to exhibiting 
\antoine{zero-crossings of their Laplacian $\Delta D$ (resp. $\Delta N$)}
around semantic boundaries, 
\antoine{defined}
as follows:
\begin{equation}
\begin{aligned}
Z_{D}(p) &= 
\begin{cases} 
1 & \text{if } \Delta D(p) \Delta 
\antoine{D \left( p + (1,0) \right)}
< 0 \\ 
1 & \text{if } \Delta D(p) \Delta 
\antoine{D \left( p + (0,1) \right)}
< 0 \\ 
0 & \text{otherwise} 
\end{cases} \\[10pt]
\end{aligned}
\end{equation}
\noindent
\antoine{Same applies to normal Laplacian $\Delta N$.}
The zero-crossing mask is then 
\antoine{dilated ($3 \times 3$ structuring element), and}
multiplied by the magnitude of the gradient to create the pseudo-labels. For this, the depth gradient is normalized to avoid bias towards large depth values:
\begin{equation}
\begin{aligned}
\nabla D_{x}(p) &= \frac{
\antoine{D \left( p + (0,1) \right)}
- D(p)}{\max \big(
\antoine{D \left( p + (0,1) \right)}, 
D(p)\big)}, \\
\antoine{\left\| \nabla D(p) \right\|}
&= \big|\nabla D_{x}(p)\big| + \big|\nabla D_{y}(p)\big|.
\end{aligned}
\end{equation}
\noindent
\noindent
The usual normal gradient, which is often too noisy, is replaced with a dot-product measure:
\begin{equation}
\begin{aligned}
\big|\nabla N_{x}(p)\big| &= 1 - \big|\langle \vec{N}(p + 
\antoine{(0,1)}
), \vec{N}(p) \rangle\big|,\\
\antoine{\left\| \nabla N(p) \right\|}
&= \big|\nabla N_{x}(p)\big| + \big|\nabla N_{y}(p)\big|.
\end{aligned}
\end{equation}
\noindent
The 
weighted depth and normal pseudo-labels
employed to supervise the edge network are then defined as follows:
\begin{equation} 
    \begin{aligned} 
        w_{d}(p) &= \frac{Z_{D}(p) \big\|\nabla D(p)\big\|}{\sum_{p} Z_{D}(p) \big\|\nabla D(p)\big\|} 
    \end{aligned} 
\end{equation} 
\begin{equation} 
    \begin{aligned} 
        w_{n}(p) &= \frac{Z_{N}(p) \big\|\nabla N(p)\big\|}{\sum_{p} Z_{N}(p) \big\|\nabla N(p)\big\|} 
    \end{aligned} 
\end{equation} 
\noindent
\textbf{A contrastive loss} is also applied to encourage edge values to move toward either zero or one. This is achieved using a binary cross-entropy loss, where pseudo ground truth \(Y(p) \in \{0,1\}\) are created by thresholding edge estimations \(E(p)\) at \(0.5\):
\noindent
\begin{equation}
\begin{aligned}
\mathcal{L}_{c} &= -\sum_{p} \Big(Y(p) \log\big(E(p)\big) + \big(1 - Y(p)\big)\log\big(1 - E(p)\big)\Big)
\end{aligned}
\end{equation}
Finally, the loss function employed to supervise the edge network is defined as follows:
\begin{equation} 
    \begin{aligned} 
        \mathcal{L}^{\text{edge}} &= \sum_{p} \left( w_{d}(p) + w_{n}(p) \right) \left( 1 - E(p) \right) + \mathcal{L}_{c} + \mathcal{L}_{e} \\ 
        & \text{with} \  \mathcal{L}_{e} = \sum_{p} 
        \antoine{E(p)^2}
        \ \text{to avoid trivial solutions.} 
    \end{aligned}
\end{equation}
\textbf{Discontinuity Preserving Smoothing:} Incorporating an edge-aware smoothness loss is a widely used practice in training depth networks \cite{wang2018learning, yang1711unsupervised}:
\begin{equation}
    \begin{aligned}
        \mathcal{L}_s = \sum_{p} \left( |\nabla D_{x}(p)| e^{-|\nabla I_{x}(p)|} + |\nabla D_{y}(p)| e^{-|\nabla I_{y}(p)|} \right)
    \end{aligned}
\label{eq:bad_smooth}
\end{equation}
\noindent
However, this approach has two drawbacks: first, boundaries are estimated using the image gradient, which is not optimal; second, there is no mechanism to effectively manage discontinuities. To address these issues, we propose a new smoothness loss \david{exploiting our pre-semantic contours} defined as follows:
\begin{equation}
\begin{aligned}
    \mathcal{L}_{s_{1}} &= \sum_{p}
    \antoine{\left\| \nabla D (p) \right\|}
    \antoine{e^{-\alpha E(p)}}, \\ 
    \mathcal{L}_{s_{2}} &= \sum_{p}\log\left(1 + 
    \antoine{e^{- \left\| \nabla D (p) \right\|}}
    \right) \left(1 - 
    \antoine{e^{-\alpha E(p)}} \right), \\
    \mathcal{L}_{s} &= \mathcal{L}_{s_{1}} + \mathcal{L}_{s_{2}}.
\end{aligned}    
\end{equation}
\label{eq:smoothness_loss}
The component \(\mathcal{L}_{s_1}\) replace the former 
\antoine{equation}
\ref{eq:bad_smooth} to encourage depth smoothness inside semantic instances, while \(\mathcal{L}_{s_2}\) prevents diffusion 
and promotes significant depth changes \david{around boundaries}. In this manner, both the edge and depth networks benefit from each other: the depth estimation supervises the edge network, while the edge network enhances depth estimation through the smoothness loss.\\
\noindent
\textbf{Post-processing:} The map produced by the edge network consists of continuous values ranging from zero to one. To obtain 
\antoine{thin pre-semantic}
contours \(\widehat{\mathcal{C}}(I)\), we apply a post-processing pipeline that includes hysteresis thresholding, non-maximum suppression, morphological closing, contour filtering, and contour 
closing. More details can be found in Appendix~\ref{sec:appendix:discussion_contour:post} and Appendix~\ref{sec:appendix:discussion_contour:comparison_lego} discusses the advantages of our approach compared to the 
competitive
work \cite{yang2018lego}.

\subsection{A variance augmented image for depth learning}
\label{sec:method:variance_augmented}

With \(\widehat{\mathcal{C}}(I)\) the 
\antoine{thin pre-semantic}
contour of \(I\) obtained after edge network estimation and subsequent post-processing, we then compute the distance transform map 
\(\delta_{\text{I}}\)
defined as:
\begin{equation}
    \begin{aligned}
        \delta_{\text{I}}(p) = \min_{x \in \widehat{\mathcal{C}}(I)}(\|x - p\|)      
    \end{aligned}
\end{equation}
\noindent
See Appendix~\ref{sec:appendix:discussion_distance:algo} for the details of the algorithm. The resulting distance transform map is concatenated with the input image to create a \textquotesingle variance-augmented\textquotesingle\ image. The objective is then to learn depth and ego-motion in order to identify corresponding pixels across successive frames, and reconstruct the upcoming structure-augmented image in the most effective manner, as illustrated on Figure~\ref{fig:graphic}. The reconstruction of the original input image \(\widehat{\text{I}}\) is evaluated using 
\antoine{$\Psi$, the standard combination of colour difference and SSIM},
while the 
reconstruction
of the distance transform map is assessed with a \(\text{L}_1\) loss:

\begin{equation}
\begin{aligned}
\mathcal{L}_{\text{photo}} &=  \Psi\left(\widehat{\text{I}}, \text{I}\right), \text{and }\\
\mathcal{L}_{\text{dist}} &= 
\left\| \widehat{\delta_{\text{I}}} - \delta_{\text{I}} \right\|_1
\end{aligned}
\end{equation}
\noindent
The smoothness loss is added to get the final supervision signal of the depth network:
\begin{equation}
    \begin{aligned}
        \mathcal{L}^{\text{depth}} = \mathcal{L}_{\text{dist}} + \mathcal{L}_{\text{photo}} + \mathcal{L}_{s}
    \end{aligned}
\end{equation}
\label{eq:total_loss}
\noindent
\textbf{An improved distance transform:} This approach can be readily extended to  the family of one dimensional functions 
\(\mathcal{F} = \left\{ p \mapsto g \circ \delta_{\text{I}}(p) \right\} \) (as defined in Equation~\ref{eq:family_f}):

\begin{equation}
\begin{aligned}
\mathcal{L}_{\text{dist}} &= 
\antoine{\left\|g \circ \widehat{\delta_{\text{I}}} - g \circ \delta_{\text{I}}\right\|_1}.
\end{aligned}
\end{equation}
\noindent
We provide an analysis of the depth estimation results for various functions in 
Appendix~\ref{sec:appendix:discussion_distance:study_g}.
\noindent
To take this a step further, we propose extending the distance transform into the \(n\)-dimensional space by mapping the distance transform values to an \(n\)-dimensional random walk process denoted as $\text{RW}_n$:
\begin{equation}
\begin{aligned}
    \mathcal{L}_{\text{dist}} &= \sum_{p} \left\|\text{RW}_n \circ 
    \widehat{\delta_{\text{I}}} 
    (p) - \text{RW}_n \circ \delta_{\text{I}}(p)\right\|_1
\end{aligned}
\end{equation}
\noindent
 Using this approach, we can maintain the validity of the constancy assumption while increasing the variance even more. Additionally, the dimensional parameter \(n\) determines the significance of the distance transform map reconstruction within the total loss function. It introduces a trade-off between \(\mathcal{L}_{\text{dist}}\) and \(\mathcal{L}_{\text{photo}}\): the higher the dimension, the greater the emphasis on the artificially introduced structure compared to the natural structure found in the image; see Appendix~\ref{sec:appendix:discussion_distance:random_walk} for more details on the random walk mapping. Our experiments in Appendix~\ref{sec:appendix:discussion_distance:study_g} show that \(n=3\) provides the best results.

\section{Experiments}
\label{sec:experiments}
\subsection{Datasets}
We conduct experiments on multiple datasets:\\
\noindent 
\textbf{KITTI \cite{geiger2013vision}} includes diverse urban, rural, and highway scenes. We follow the standard data split defined by \cite{eigen2014depth} and apply the pre-processing steps outlined in \cite{zhou2017unsupervised} to filter static frames, resulting in a training set of 39\,810 images and 697 images for testing. Results are reported using the evaluation protocol from \cite{zhou2017unsupervised}.\\ 
\noindent
\textbf{Cityscapes \cite{cordts2016cityscapes}} offers urban scenes with dynamic objects, challenging depth estimation. The training set consists of 22\,973 images, with evaluation on 1\,525 test images following \cite{casser2019unsupervised, li2021unsupervised}. Edge detection is assessed using the contour evaluation from \cite{yang2018lego} on 500 validation frames.\\
\noindent 
\textbf{Waymo \cite{sun2020scalability}} is a large and diverse autonomous driving dataset that captures dynamic urban scenes under various environmental conditions, including nighttime and different weather scenarios. We sample \(100\,000\) image pairs from \(1\,000\) front camera video sequences for training and evaluate on 1\,500 pairs from 150 sequences, following \cite{li2021unsupervised}.\\
\noindent 
\textbf{NYUv2 \cite{silberman2012indoor}} consists of 464 indoor video sequences captured with a Kinect sensor. We use the official splits (302 training, 33 validation) and evaluate on 654 densely annotated test images, resizing inputs to \(320\times256\) pixels as in \cite{li2022monoindoor++}.\\
\noindent 
\textbf{ScanNet \cite{dai2017scannet}} is a large-scale RGB-D dataset with 2.5 million images across 1\,500+ scenes. We use it for zero-shot evaluation, following \cite{fan2023deeper, li2022monoindoor++}, with test images resized to \(320\times256\) pixels.

\subsection{Implementation Details}
We emphasize that our method is compatible with any DNN architecture and input image size. For fair comparison, our experiments are done with a \(256\times 832\) resolution and a UNet \cite{ronneberger2015u} structure, following \cite{zhou2017unsupervised}, unless stated otherwise. The depth and edge networks share a ResNet \cite{he2016deep} encoder. Our baseline, CoopNet \cite{hariat2023rebalancing}, handles moving objects—especially useful in Cityscapes—via self-supervised flow. The pose network, based on a ResNet encoder, outputs a 6-DoF vector. All encoders use ResNet-50 backbones initialized with ImageNet \cite{russakovsky2015imagenet} weights. Training runs for 30 epochs with batch size 4,  initial learning rate of \(10^{-4}\) (decreasing to \(10^{-5}\) after 20 epochs), and standard data augmentation from \cite{godard2019digging}. Loss weights, tuned via grid search, are detailed in Appendix~\ref{sec:appendinx:hyper_parameters}. Our PyTorch \cite{paszke2019pytorch} implementation trains on a single NVIDIA RTX A5000 GPU using Adam \cite{kingma2014adam} with \(\beta_{1} = 0.99\) and \(\beta_{2} = 0.999\)) and takes 12 hours.
\label{sec:experiments:implementation_details}
\subsection{Results}
\label{sec:experiments:depth_results}

\definecolor{lightblue}{rgb}{0.1, 0.4, 0.6} 

\begin{table}[ht]
\centering
\renewcommand{\arraystretch}{1.5} 
\fontsize{16}{18}\selectfont 
\resizebox{\linewidth}{!}{ 
\begin{tabular}{ r c c c c c c c }
\hline
\multirow{2}{*}{\textbf{Method}} 
& \multicolumn{4}{c}{\textit{Lower is better} \(\downarrow\)} 
& \multicolumn{3}{c}{\textit{Higher is better} \(\uparrow\)} \\  
\cmidrule(r){2-5} \cmidrule(l){6-8} 
& \textcolor{lightblue}{Abs Rel} & Sq Rel & RMSE & \textcolor{lightblue}{RMSE log} 
& \hspace{15pt}\textcolor{lightblue}{\(\delta_{1}\)} & \hspace{15pt}\(\delta_{2}\) &  \hspace{15pt}\(\delta_{3}\) \\
\hline
Monodepth2\cite{godard2019digging} & 0.110 & 0.831 & 4.642 & 0.187 &  \hspace{15pt}0.883 & \hspace{15pt} 0.962 &  \hspace{15pt}0.982\\
\(\text{Guizilini \textit{et al.}}^{\dagger}\)\cite{guizilini2020semantically} (R50) & 0.113 & 0.831 & 4.663 & 0.189 &  \hspace{15pt } 0.878 & \hspace{15pt}  \textbf{0.971} &  \hspace{15pt} 0.983\\
\(\text{Shu \textit{et al.}}^{*}\)\cite{shu2020feature}(R50)  & \underline{0.108} & \underline{0.792} & \underline{4.633} & \underline{0.184} &  \hspace{15pt}\underline{0.883} &  \hspace{15pt}0.961 & \hspace{15pt}\underline{0.983}\\
Ours (R50) & \textbf{0.104} & \textbf{0.725} & \textbf{4.453} & \textbf{0.180} & \hspace{15pt}\textbf{0.885} &  \hspace{15pt} \underline{0.962} & \hspace{15pt} \textbf{0.983}\\
\hline
MonoViT\cite{zhao2022monovit} & 0.099 & 0.708 & 4.372 & 0.175 &  \hspace{15pt } 0.900 & \hspace{15pt}  0.967 &  \hspace{15pt} 0.984\\
Ours + MonoViT & \textbf{0.092} & \textbf{0.674} & \textbf{4.300} & \textbf{0.165} & \hspace{15pt}\textbf{0.927} &  \hspace{15pt} \textbf{0.967} & \hspace{15pt} \textbf{0.984}\\
\hline
HR-Depth\cite{lyu2021hr} & 0.109 & 0.792 & 4.632 & 0.185 &  \hspace{15pt } 0.884 & \hspace{15pt}  0.962 &  \hspace{15pt} 0.983\\
RA-Depth\cite{he2022ra} & 0.096 & \underline{0.613} & \underline{4.216} & \underline{0.171} &  \hspace{15pt} \underline{0.903} & \hspace{15pt}\underline{0.968} &  \hspace{15pt} \underline{0.985}\\
DIFFNet\cite{zhou_diffnet} & 0.102 & 0.764 & 4.483 & 0.180 &  \hspace{15pt } 0.896 & \hspace{15pt}  0.965 &  \hspace{15pt} 0.983\\
\(\text{Guizilini \textit{et al.}}^{\dagger}\)\cite{guizilini2020semantically}& \underline{0.100} & 0.761 & 4.270 & 0.175 &  \hspace{15pt } 0.902 & \hspace{15pt}  0.965 &  \hspace{15pt} 0.982\\
Ours+DinoV2 & \textbf{0.082} & \textbf{0.604} & \textbf{4.108} & \textbf{0.162} &  \hspace{15pt}\textbf{0.928} & \hspace{15pt} \textbf{0.968} &  \hspace{15pt} \textbf{0.985}\\

\hline

Lego (paper) & 0.154 & 1.272 & 6.012 & 0.230 &  \hspace{15pt }N/A & \hspace{15pt} N/A &  \hspace{15pt} N/A\\

\(\text{Lego}^{\bigstar}\) (R50) & 0.115 & 0.855 & 4.789 & 0.195 &  \hspace{15pt} 0.876 & \hspace{15pt} 0.957 &  \hspace{15pt}0.979\\

\(\text{Lego}^{\bigstar}\) (R50) + \scalebox{1.2}{$\mathcal{L}_{\text{dist}}$} & 0.110 & 0.805 & 4.606 & 0.188 &  \hspace{15pt} 0.882 & \hspace{15pt} 0.960 &  \hspace{15pt} 0.982\\

Ours (R50) & \textbf{0.104} & \textbf{0.725} & \textbf{4.453} & \textbf{0.180} & \hspace{15pt}\textbf{0.885} &  \hspace{15pt}\textbf{0.962} & \hspace{15pt} \textbf{0.983}

\\
\hline
\(\text{Struct2Depth}^{\dagger}\)\cite{casser2019depth} & 0.145 & 1.737 & 7.28 & 0.205 &  \hspace{15pt}0.813 & \hspace{15pt}0.942 &  \hspace{15pt}0.978\\
\(\text{LearnK}^{\dagger}\)\cite{li2021unsupervised} & 0.127 & 1.330 & \underline{6.96} & 0.195 &  \hspace{15pt}0.830 &  \hspace{15pt}0.947 & \hspace{15pt}\textbf{0.981}\\
\(\text{Li \textit{et al.}}^{\dagger}\)\cite{li2021unsupervised} & \underline{0.119} & \underline{1.290} & 6.98 & \underline{0.190} &  \hspace{15pt}\underline{0.846} &  \hspace{15pt}\underline{0.951} & \hspace{15pt}\underline{0.980}\\
CoopNet\cite{hariat2023rebalancing} & 0.121 & 1.443 & 7.01 & \underline{0.190} &  \hspace{15pt}\underline{0.846} &  \hspace{15pt}\underline{0.951} & \hspace{15pt}\underline{0.980}\\
Ours (R50) & \textbf{0.115} & \textbf{1.221} & \textbf{6.79} & \textbf{0.186} &  \hspace{15pt}\textbf{0.850} &  \hspace{15pt}\textbf{0.955} & \hspace{15pt}\textbf{0.981} \\
\hline

\end{tabular}
}

\caption{\textbf{Results of depth estimations on KITTI 2015 (first four blocks) and Cityscapes (last block).}  For each metric the best result is displayed in bold and the second-best is underlined. Light blue metrics indicates the most challenging metrics. \(\bm{\dagger}\)\, denotes the use of an off-the-shelf semantic algorithm and *\, indicates results reproduced by us at the input resolution of \(256 \times 832\). Block\#1 and \#2 compares 
methods using similar encoder backbones: Resnet 50 (R50) in block\#1 and ViT in block\#2. Block\#3 compares SOTA depth models without restrictions on architecture or additional inputs, Block\#4 compares to Lego.}
\label{table:results_depth_kitti}
\end{table}

\begin{table}[ht]
\centering
\resizebox{\linewidth}{!}{ 
\renewcommand{\arraystretch}{1.2} 
\fontsize{16}{18}\selectfont 
\begin{tabular}{ r l c c c c }
\hline
\multirow{2}{*}{\textbf{Method}} 
& \multirow{2}{*}{\textbf{Dataset \antoine{(Train / Test)}}} 
& \multicolumn{4}{c}{\textit{Lower is better} \(\downarrow\)} \\  
\cmidrule(l){3-6} 
& & Abs Rel & Sq Rel & RMSE & RMSE log \\
\hline

LearnK\cite{li2021unsupervised} & Waymo \antoine{/ Waymo} & 0.157 & 1.531 & 7.090 & 0.205\\
Ours & Waymo \antoine{/ Waymo} & \textbf{0.125} & \textbf{1.278} & \textbf{5.990} & \textbf{0.189}\\
\hline
GLNet\cite{chen2019self}& \antoine{Cityscapes / KITTI} & 0.129 & 1.044 & 5.361 & 0.212\\
CoopNet\cite{hariat2023rebalancing} & \antoine{Cityscapes / KITTI} & 0.125 & 1.157 & 5.251 & 0.209\\
Ours & \antoine{Cityscapes / KITTI} & \textbf{0.116} & \textbf{0.892} & \textbf{4.892} & \textbf{0.195}\\
\end{tabular}
}
\caption{
Results on Outdoor datasets. Block\#1: in-domain evaluation. 
Block\#2: out-of-domain evaluation. 
}
\label{table:results_outdoor}
\end{table}

\begin{table}[ht]
\centering
\resizebox{\linewidth}{!}{ 
\renewcommand{\arraystretch}{1.2} 
\fontsize{16}{18}\selectfont 
\begin{tabular}{ r l c c c c c }
\hline
\textbf{Method} 
& \textbf{Dataset \antoine{(Train / Test)}} 
& Abs Rel & RMSE & \(\delta_1\) & \(\delta_2\) & \(\delta_3\) \\
\hline

MonoIndoor++\cite{li2022monoindoor++} & NYUv2 \antoine{/ NYUv2} & 0.132 & 0.517 & 0.834 & 0.961 & 0.990\\
IndoorDepth\cite{fan2023deeper} & NYUv2 \antoine{/ NYUv2} & 0.126 & 0.494 & 0.845 & 0.965 & 0.991\\
Ours & NYUv2 \antoine{/ NYUv2} & \textbf{0.115} & \textbf{0.458} & \textbf{0.859} & \textbf{0.970} & \textbf{0.992}\\
\hline
IndoorDepth\cite{fan2023deeper} & NYUv2 / ScanNet & 0.153 & 0.373 & 0.786 & 0.950 & 0.988\\
MonoIndoor++\cite{li2022monoindoor++} & NYUv2 / ScanNet & 0.138 & 0.347 & 0.810 & 0.967 & 0.993\\
Ours & NYUv2 / ScanNet & \textbf{0.127} & \textbf{0.312} & \textbf{0.843} & \textbf{0.970} & \textbf{0.993}\\

\end{tabular}
}
\caption{Results on Indoor datasets. Block\#1: in-domain evaluation.
Block\#2: out-of-domain evaluation.
}
\label{table:results_indoor}
\end{table}

\newcommand{\thickcheck}{\scalebox{1.5}{\checkmark}}

\begin{table}[ht]
\centering
\resizebox{\linewidth}{!}{ 
\renewcommand{\arraystretch}{1.2} 

\small 
\renewcommand{\arraystretch}{1} 
\fontsize{8}{10}\selectfont 
\begin{tabular}{ c c c c c c c}
\hline
\(\text{RW}_{3}\) & \(\mathcal{L}_{\text{dist}}\) & \(\mathcal{L}_{s}\)
& \(E_{\theta}\)
& \textcolor{lightblue}{Abs Rel (\(\downarrow\))} & \textcolor{lightblue}{RMSE log (\(\downarrow\))} & \textcolor{lightblue}{\(\delta_{1}\) (\(\uparrow\))}\\
\hline
 & & & & 0.113 & 0.19 & 0.878\\
 & & & \ding{51} & 0.112 & 0.189 & 0.880\\
 & & \ding{51} & \ding{51} & 0.110 & 0.187 & 0.881\\
 & \ding{51} & \ding{51} & \ding{51} & 0.106 & 0.183 & 0.884\\
 \ding{51} & \ding{51} & \ding{51} & \ding{51} & \textbf{0.104} & \textbf{0.180} & \textbf{0.885}\\
\hline
\end{tabular}
}
\caption{Ablation study on KITTI 2015 of losses 
mentioned in Section~\ref{sec:method:variance_augmented}. Note that \(E_{\theta}\) is part of the ablation as it does improve the depth metrics 
\antoine{and}
shares the same encoder as the depth network. \(\text{RW}_{3}\) is for 
\antoine{3d random walk encoding of distance transform}
as defined in Section~\ref{sec:method:variance_augmented}.} 
\label{table:results_ablation}
\end{table}

\noindent
\textbf{Depth} Quantitative results on KITTI and Cityscapes are presented in Table~\ref{table:results_depth_kitti}, where we compare against the most competitive methods. Blocks \#1 and \#2 demonstrate that, when using the same encoder backbone, our method outperforms others by a significant margin across nearly all metrics. When no architectural constraints are imposed, we employ a pre-trained DinoV2 \cite{oquab2023dinov2} encoder (further details in Appendix~\ref{sec:appendix:hyper_parameters:dino}), yielding the results in block \#3. 
For fairness, we compare against \(\text{LEGO}^{\bigstar}\) in block \#4, an enhanced version of LEGO that incorporates a ResNet-50 backbone and up-to-date training techniques. Adding our novel distance transform loss \(\mathcal{L}_\text{dist}\) into \(\text{LEGO}^{\bigstar}\) greatly improves results, showing once again the great importance of this loss component, but still underperforms ours, likely due to the superior quality of our contours. 

We also provide strong out-of-domain results in Table~\ref{table:results_outdoor} and Table~\ref{table:results_indoor}, demonstrating competitive performance on both indoor and outdoor datasets. Finally, an ablation study in Table~\ref{table:results_ablation} evaluates the impact of the different loss components introduced in our approach \antoine{and in Appendix~\ref{sec:appendix:discussion_distance:study_g} for different functions from the family} 
\(\mathcal{F}\) as defined in Section~\ref{sec:preambule:max_variance}.
It can be seen that the reprojection on the distance transform via \(\mathcal{L}_{\text{dist}}\) brings a decisive improvement. 
We provide a qualitative comparison of results in Figure~\ref{fig:qualitative_comparison_depth}. We can see that our method gives better results: it provides thinner object, is sharper around moving objects, does not suffer from smoothness issues for large objects and is not noisy in textured areas. More qualitative and quantitative results can be found in Appendix~\ref{sec:appendix:extra_results:depth}.\\

\noindent
\textbf{Flow \& Odometry} As discussed in Section~\ref{sec:experiments:implementation_details} we use an optical flow to handle moving objects following \cite{hariat2023rebalancing}. 
\antoine{Its results}
can be found in Appendix~\ref{sec:appendix:extra_results:flow}. Results 
\antoine{for}
odometry can be found in Appendix~\ref{sec:appendix:extra_results:odometry}. Our method significantly outperforms the baseline~\cite{hariat2023rebalancing} on both optical flow and odometry metrics.

\subsection{Contour evaluation}
We evaluate contour detection in Table~\ref{table:results_edges} using the same metrics and data as \cite{yang2018lego}: Optimal Dataset Scale, Optimal Image Scale, and Average Precision, computed on 500 Cityscapes validation images. Our method significantly outperforms LEGO, leveraging (1) depth \& normal pseudo-labels with Laplacian zero-crossing insights for better contour alignment, (2) the discontinuity-preserving smoothing loss \(\mathcal{L}_{s}\), and (3)  the contrastive loss \(\mathcal{L}_{c}\). Qualitative examples are in Figure~\ref{fig:qualitative_robustness}, with more in Appendix~\ref{sec:appendix:discussion_contour}.

\twocolumn[{
\begin{center}
    \captionsetup{type=figure}
    \captionsetup[subfigure]{labelformat=empty}
    \addtocounter{figure}{-1}
    
    \begin{subfigure}[b]{0.24\textwidth}
        \centering
        \includegraphics[width=\textwidth, height=1.2cm]{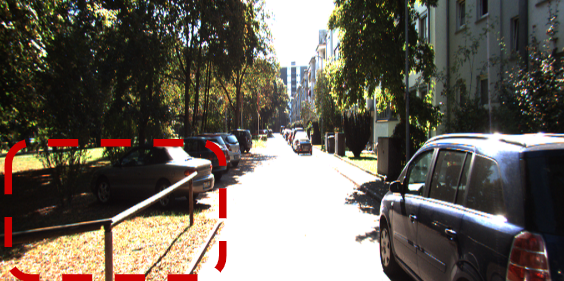}
    \end{subfigure}
    \hfill
    \begin{subfigure}[b]{0.24\textwidth}
        \centering
        \includegraphics[width=\textwidth, height=1.2cm]{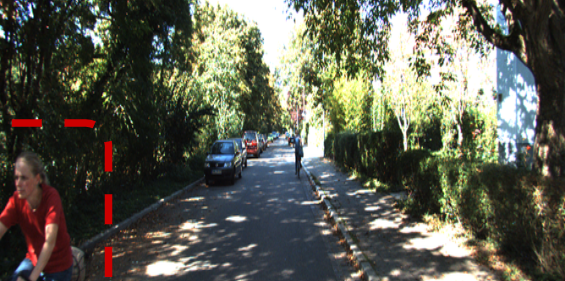}
    \end{subfigure}
    \hfill
    \begin{subfigure}[b]{0.24\textwidth}
        \centering
        \includegraphics[width=\textwidth, height=1.2cm]{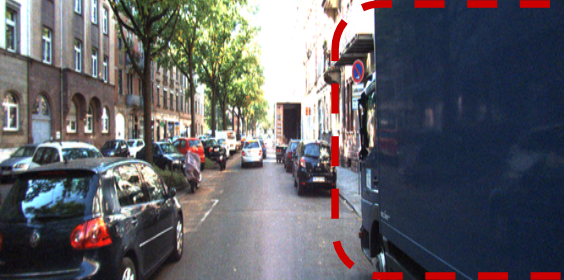}
    \end{subfigure}
    \hfill
    \begin{subfigure}[b]{0.24\textwidth}
        \centering
        \includegraphics[width=\textwidth, height=1.2cm]{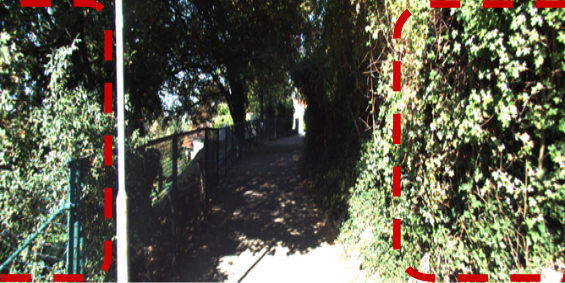}
    \end{subfigure}

    \vspace{0.2cm}  

    \begin{subfigure}[b]{0.24\textwidth}
        \centering
        \includegraphics[width=\textwidth, height=1.2cm]{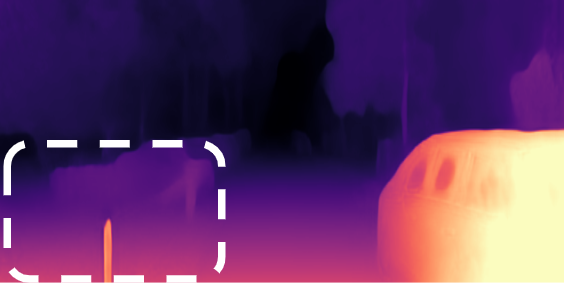}
        \label{figteaser:Orginalimage-anom}
    \end{subfigure}
    \hfill
    \begin{subfigure}[b]{0.24\textwidth}
        \centering
        \includegraphics[width=\textwidth, height=1.2cm]{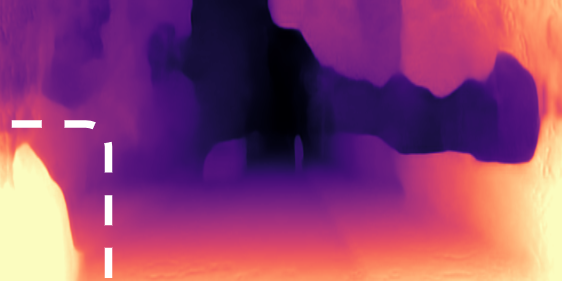}
        \label{figteaser:RobustaGeneration-anom}
    \end{subfigure}
    \hfill
    \begin{subfigure}[b]{0.24\textwidth}
        \centering
        \includegraphics[width=\textwidth, height=1.2cm]{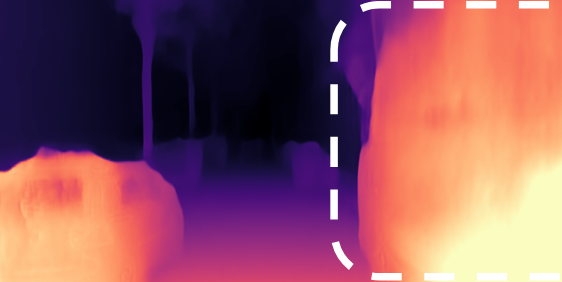}
        \label{figteaser:OasisGeneration-anom}
    \end{subfigure}
    \hfill
    \begin{subfigure}[b]{0.24\textwidth}
        \centering
        \includegraphics[width=\textwidth, height=1.2cm]{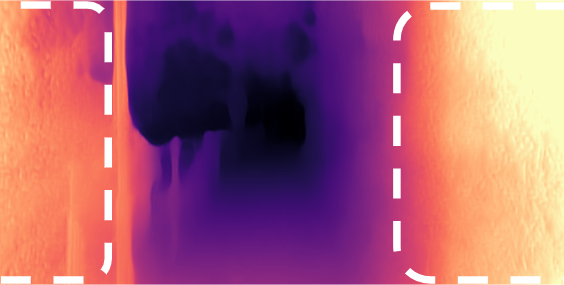}
        \label{figteaser:Label}
    \end{subfigure}

    \vspace{0.2cm}  

    \begin{subfigure}[b]{0.24\textwidth}
        \centering
        \includegraphics[width=\textwidth, height=1.2cm]{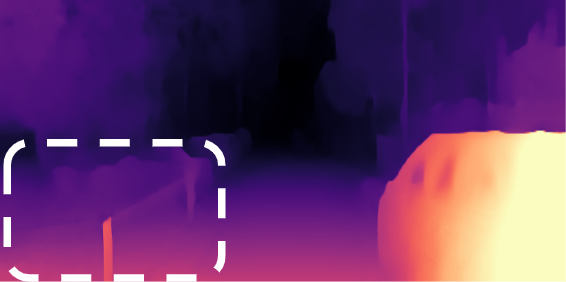}
        \label{figteaser:Orginalimage-anom}
    \end{subfigure}
    \hfill
    \begin{subfigure}[b]{0.24\textwidth}
        \centering
        \includegraphics[width=\textwidth, height=1.2cm]{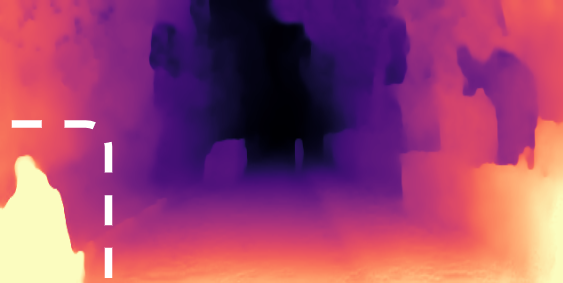}
        \label{figteaser:RobustaGeneration-anom}
    \end{subfigure}
    \hfill
    \begin{subfigure}[b]{0.24\textwidth}
        \centering
        \includegraphics[width=\textwidth, height=1.2cm]{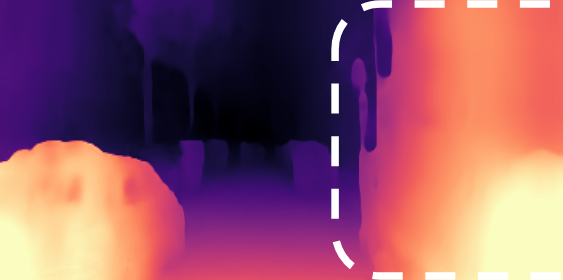}
        \label{figteaser:OasisGeneration-anom}
    \end{subfigure}
    \hfill
    \begin{subfigure}[b]{0.24\textwidth}
        \centering
        \includegraphics[width=\textwidth, height=1.2cm]{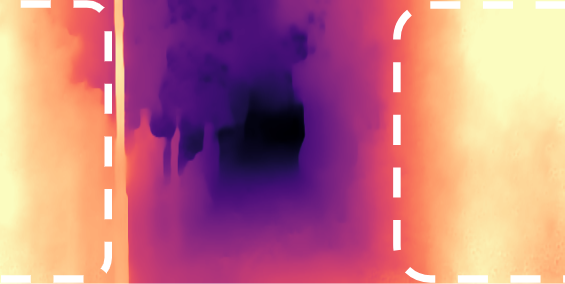}
        \label{figteaser:Label}
    \end{subfigure}

    \vspace{0.2cm}  

    \begin{subfigure}[b]{0.24\textwidth}
        \centering
        \includegraphics[width=\textwidth, height=1.2cm]{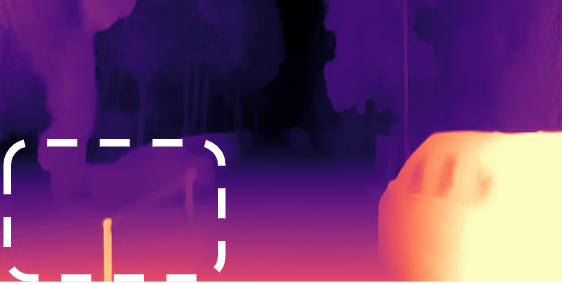}
        \label{figteaser:Orginalimage-anom}
    \end{subfigure}
    \hfill
    \begin{subfigure}[b]{0.24\textwidth}
        \centering
        \includegraphics[width=\textwidth, height=1.2cm]{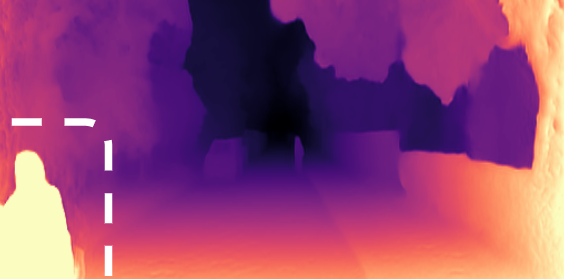}
        \label{figteaser:RobustaGeneration-anom}
    \end{subfigure}
    \hfill
    \begin{subfigure}[b]{0.24\textwidth}
        \centering
        \includegraphics[width=\textwidth, height=1.2cm]{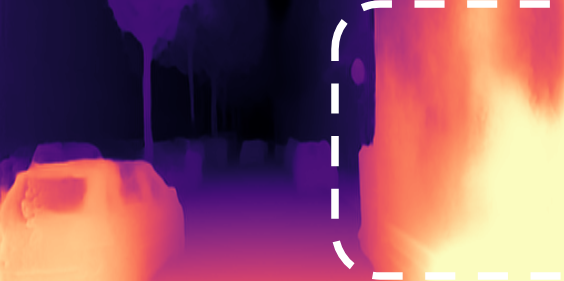}
        \label{figteaser:OasisGeneration-anom}
    \end{subfigure}
    \hfill
    \begin{subfigure}[b]{0.24\textwidth}
        \centering
        \includegraphics[width=\textwidth, height=1.2cm]{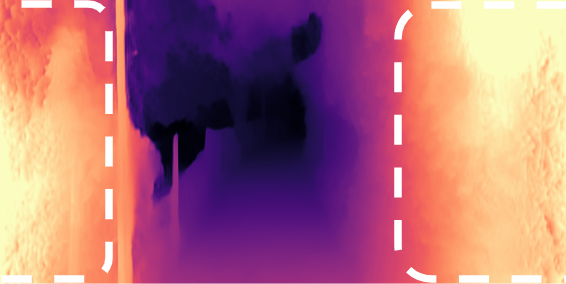}
        \label{figteaser:Label}
    \end{subfigure}

    \vspace{0.2cm}  

    \begin{subfigure}[b]{0.24\textwidth}
        \centering
        \includegraphics[width=\textwidth, height=1.2cm]{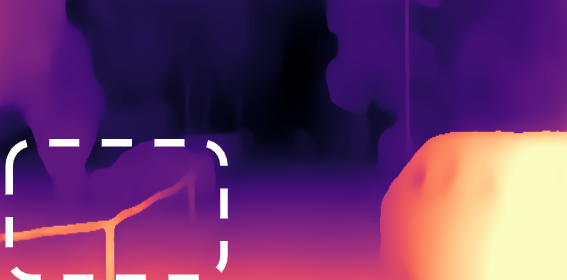}
        \label{figteaser:Orginalimage-anom}
    \end{subfigure}
    \hfill
    \begin{subfigure}[b]{0.24\textwidth}
        \centering
        \includegraphics[width=\textwidth, height=1.2cm]{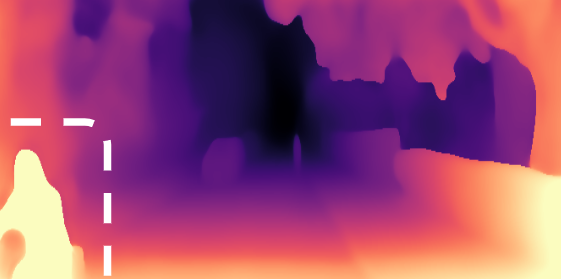}
        \label{figteaser:RobustaGeneration-anom}
    \end{subfigure}
    \hfill
    \begin{subfigure}[b]{0.24\textwidth}
        \centering
        \includegraphics[width=\textwidth, height=1.2cm]{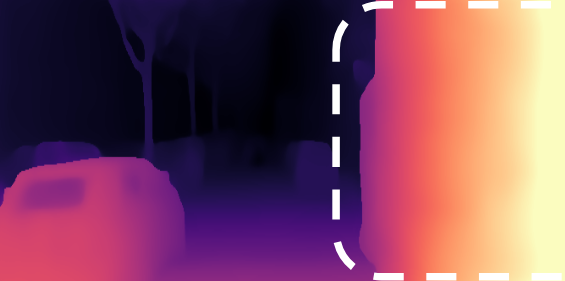}
        \label{figteaser:OasisGeneration-anom}
    \end{subfigure}
    \hfill
    \begin{subfigure}[b]{0.24\textwidth}
        \centering
        \includegraphics[width=\textwidth, height=1.2cm]{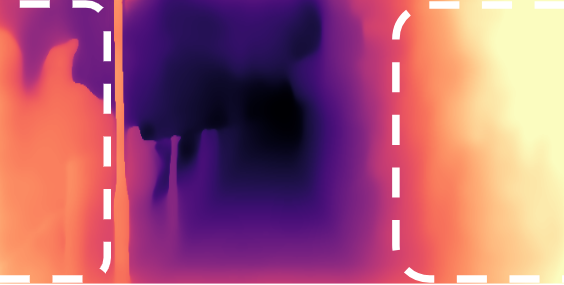}
        \label{figteaser:Label}
    \end{subfigure}
    
    \captionof{figure}{Qualtitative comparisons for different methods. First row: colour images. Second row: Monodepth2~\cite{godard2019digging}. Second row: CoopNet~\cite{hariat2023rebalancing}. Third row: FeatDepth~\cite{shu2020feature}. Last row: Ours. We compare qualitative results in four complex situations. First column: a thin object. Second column: a moving object. Third column: a large object with smoothness issue. Last column: Textured areas.}
    \label{fig:qualitative_comparison_depth}
\end{center}
}]

\begin{table}[ht]
\centering
\resizebox{\linewidth}{!}{ 
\renewcommand{\arraystretch}{1.2} 
\setlength{\tabcolsep}{12pt} 
\begin{tabular}{| c | c | c | c |}
\hline
\textbf{Methods} & \textbf{ODS} & \textbf{OIS} & \textbf{AP} \\
\hline
Lego (paper) & 0.710 & 0.731 & 0.729 \\
\(\text{Lego}^{\bigstar}\) & 0.709 & 0.736 & 0.734 \\
\hline
Depth \& Normal pseudo-labels & 0.755 & 0.775 & 0.760 \\
Depth \& Normal pseudo-labels + \(\mathcal{L}_s\) & 0.760 & 0.782 & 0.766 \\
Ours = Depth \& Normal pseudo-labels + \(\mathcal{L}_s\) + \(\mathcal{L}_c\)& \textbf{0.762} & \textbf{0.789} & \textbf{0.770} \\
\hline
\end{tabular}
}
\vspace{-5pt} 
\caption{Comparison for contour estimation.
Block\#1: Lego results. Block\#2: Ablation study of losses in contour estimation training. Experiments done on Cityscapes.}
\label{table:results_edges}
\end{table}

\subsection{Constancy assumption validity}
\label{sec:discussions:constancy_assumption}
We evaluate the constancy assumption validity of our method, 
\antoine{compared to}
\cite{shu2020feature} 
\antoine{that uses deep features from the ResNet-50 encoder in the re-projection.}
To do so, we use the KITTI MOTS \cite{Voigtlaender19CVPR_MOTS} dataset to track a specific point of an instance along different trajectories and compute the normalized temporal variance. We show that our distance transform satisfies much better the constancy assumption than any layers of the ResNet-50 
(see Figure~\ref{fig:constancy_assumption}). More details on the evaluation protocol are given in Appendix~\ref{sec:appendix:discussion_constancy}.

\begin{figure}[ht]
\centering
  \includegraphics[width=\linewidth]{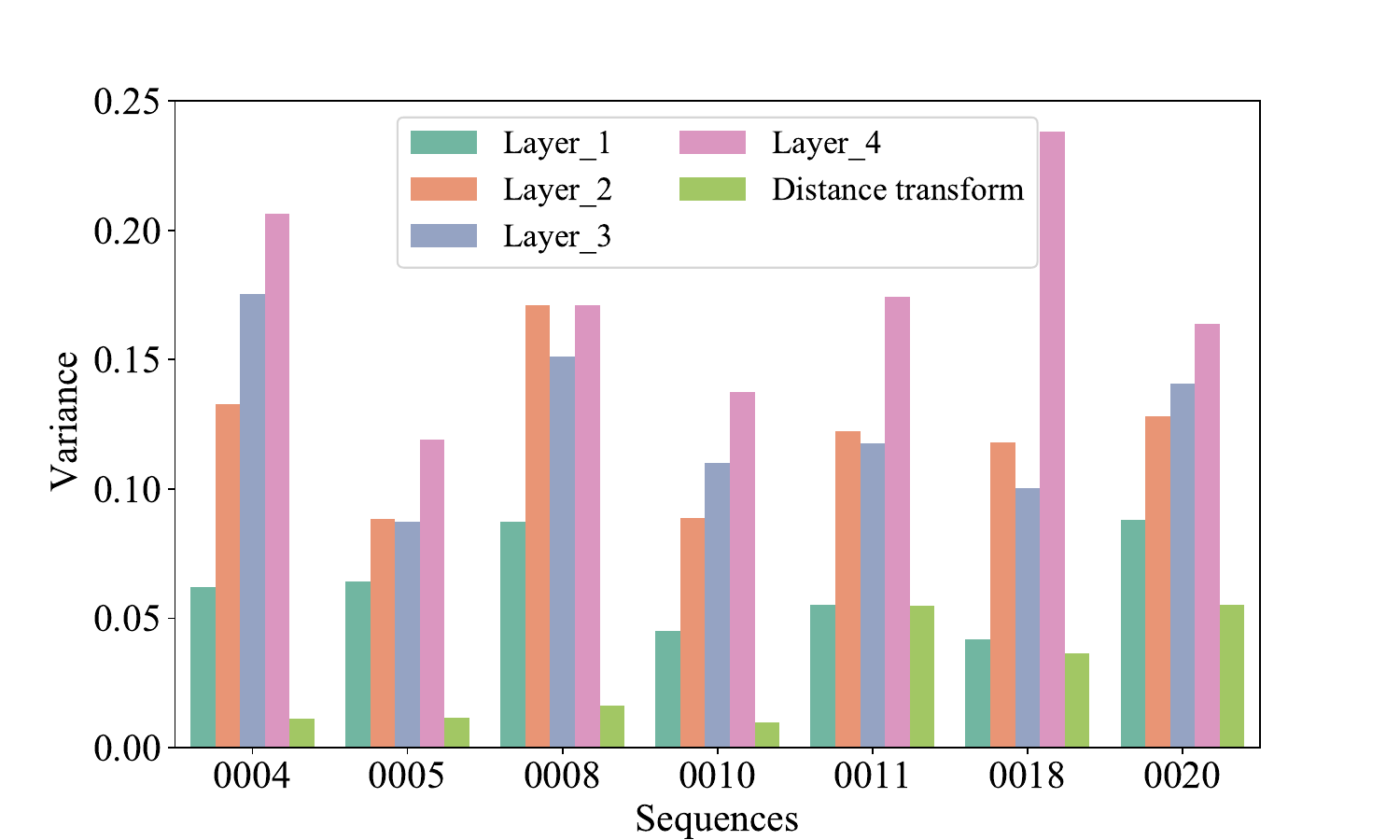}
  \caption{Comparison of normalized temporal variance of different features for the centre pixel of an instance tracked along different sequence of KITTI MOTS \cite{Voigtlaender19CVPR_MOTS}.}
  \label{fig:constancy_assumption}
\end{figure}

\section{Conclusion}

As a conclusion, this paper presents a novel self-supervised approach to monocular depth estimation, addressing challenges in low-texture regions by applying a distance transform over pre-semantic contours. This technique enhances depth prediction accuracy by increasing discriminative power where traditional photometric losses struggle.  Both theoretical analysis and extensive experiments validate the effectiveness of the proposed method, which outperforms conventional self-supervised approaches.\\
In future work, we aim to introduce consistent variance in low-texture regions without pre-semantic contours. Additionally, we plan to explore whether our framework can leverage variance-augmented images to learn from diverse datasets, ultimately developing a foundation model in a fully self-supervised manner.

{   
    \small
    \bibliographystyle{ieeenat_fullname}
    \bibliography{main}
}

\appendix
\clearpage
\setcounter{page}{1}
\setcounter{section}{0}
\maketitlesupplementary
\renewcommand\thesection{\Alph{section}}
\newtheorem{mytheorem}{Theorem}
\setcounter{figure}{0}

\numberwithin{figure}{section}
\numberwithin{equation}{section}
\numberwithin{table}{section}

\noindent
The supplementary material includes multiple details and insights that complement the main paper. 

\section{Theoretical Analysis}
\label{sec:appendix:theoretical}
In this section, we elaborate 
on the mathematical formalism of the distance transform. We also give more details on the toy experiments that corroborate the theoretical properties.
\subsection{Maximal variance under constraints}
\label{sec:appendix:theoretical:max_variance}
\begin{figure}[t]
\centering
\setlength{\tabcolsep}{0pt}
\renewcommand{\arraystretch}{1.0}
\includegraphics[width=0.4\linewidth]{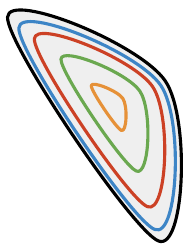}\includegraphics[width=0.4\linewidth]{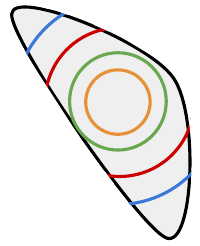}\\
\footnotesize{Level sets of distance transform (left) and distance to the centre (right) }\\
\hspace{0.5cm}\includegraphics[width=0.33\linewidth]{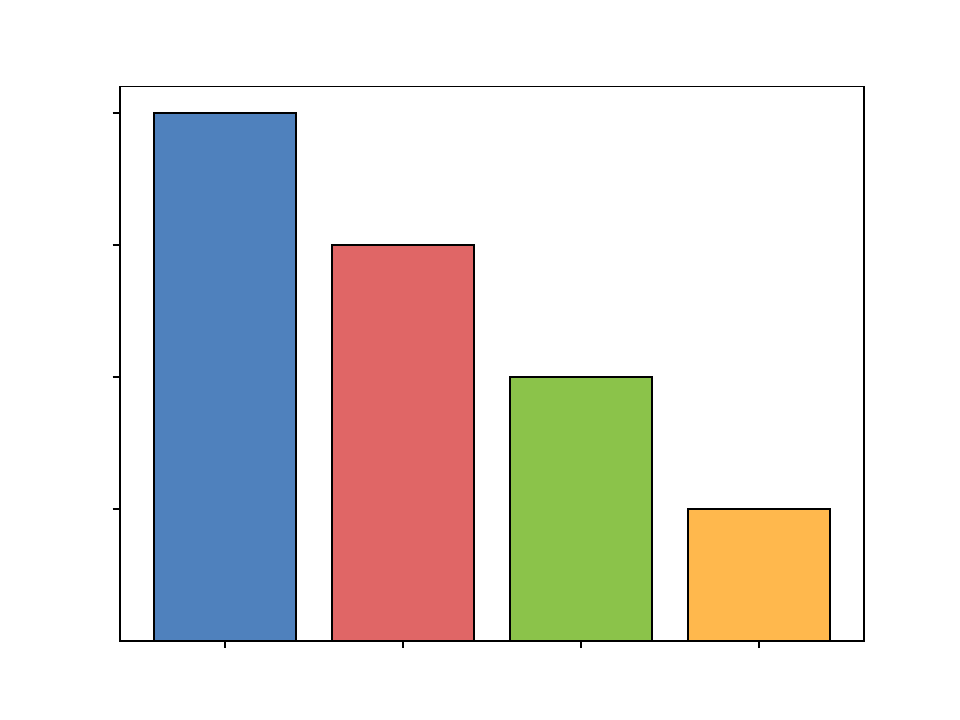} \hspace{0.5cm}\includegraphics[width=0.33\linewidth]{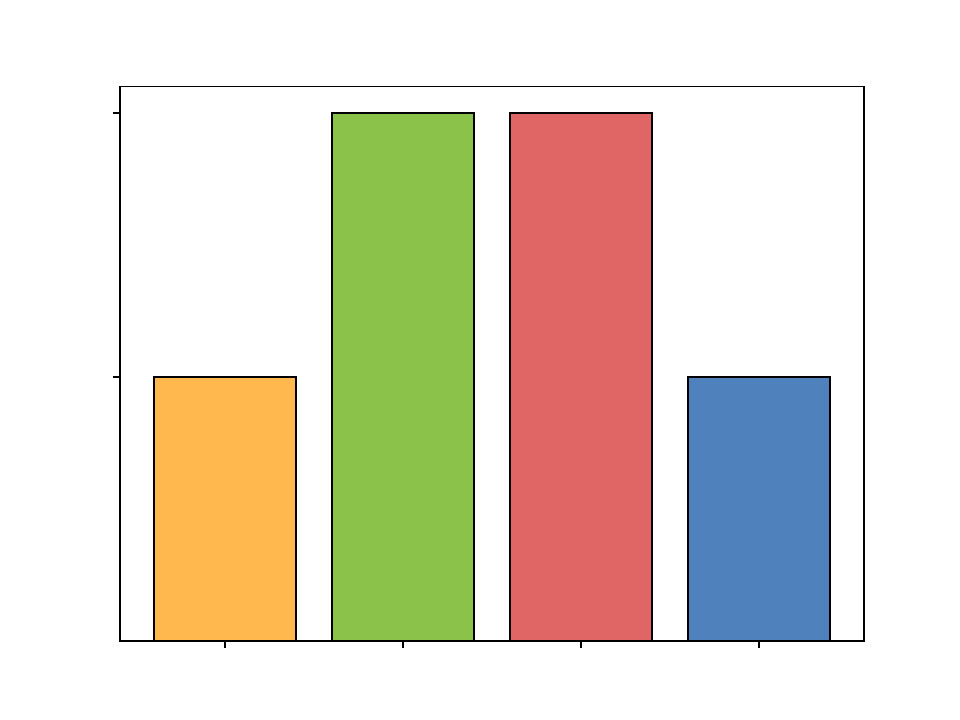}\\
\footnotesize{Histogram of distance transform values (left) and distance to centre (right)}
\caption{Illustration of the histogram of the number of pixels for each value for different level set. }
\label{fig:appendix:level_set}
\end{figure}

Let us give a proof of Theorem 1.
As a reminder:
\begin{mytheorem}
The distance transform is the unique solution, up to an isomorphism, of the following optimization problem:
\begin{equation}
\begin{aligned}
\max_{f \in \mathcal{I}_{\mathcal{T}}} \quad  \int_{\Omega} 
\AMsup{\Big(f(x) - \overline{f}\Big)^2 dx} \quad
\textrm{s.t.} \quad \|\nabla f\|=1\\
\end{aligned}
\end{equation}
\end{mytheorem}
Proof:\\
\noindent
We only provide a sketch of the proof to keep things simple. Level sets of a function 
\AMsup{$f$ are the $\{f^{-1}(v)\}$ for each specific value $v$}. 
For example, a function defined as the distance to a fixed point has circular level sets (Figure~\ref{fig:appendix:level_set}, right). In 
\AMsup{contrast, in}
the case of the distance transform, the level sets conform to the shape. Let us decompose the shape \(\Omega\) into a countably infinite number of level set slices. Then the level set lengths form a 
\AMsup{decreasing series}, \(\{f_n\}_{n \in \mathbb{N}}\),
as shown in Figure~\ref{fig:appendix:level_set} left, with:

\begin{equation} 
    \begin{aligned} 
        f_n \xrightarrow{n \to \infty} 0
    \end{aligned}
\end{equation}


\noindent
This unique property of the level sets in the distance transform case 
enables the existence of an isomorphism \(\varPhi\) that can 
be defined as an increasing function of the inverse of \(f_{n}\)

\begin{equation} 
    \begin{aligned} 
        \Phi(n) &\sim h\left( \frac{1}{f_n} \right) \quad \text{such that} \\
        &\sum_{n=1}^{\infty} \phi(n)f_n < \infty, \\
        &\sum_{n=1}^{\infty} \phi(n)^2f_n < \infty\\
        &\text{where } h \text{ is an increasing function.}
    \end{aligned}
\end{equation}
\noindent
In this way the mean value as well as the variance can be increased drastically. This is not possible for circular level sets, where the 
\AMsup{histogram}
is not monotonic as illustrated in Figure~\ref{fig:appendix:level_set}.

\subsection{Convergence properties}
\label{sec:appendix:theoretical:convergence}
Let us give a proof of theorem 2.
As a reminder:
\begin{mytheorem}
The loss function \(l\) \AMsup{(See Equation \ref{eq:loss_distance})}
is, with respect to the first argument \(\hat{y}\):
\begin{itemize}
    \item \(\alpha\)-Lipschitz
    \item \(\beta\)-smooth
    \item strictly convex if a regularization term  \( \eta \|y\|^2  \)  is added 
\end{itemize}
\end{mytheorem}
Proof:\\
\noindent
Let \(y\) be the ground truth pixel 
as explained in \AMsup{Section}
\ref{sec:premabule:convergence_properties} and let \(g\) be the function as defined in 
\AMsup{Equation}
\ref{eq:family_f},  with \(g'\) and \(g''\) bounded respectively by \(K_1\) and \(K_2\).\\
\textbf{Lipschitz: } We want to prove that:
\begin{equation}
\begin{aligned}
\exists \alpha \text{ s.t } \forall u, v \in \Omega: |l(u) - l(v)| \leq \alpha \|u - v\|
\end{aligned}
\end{equation}

\begin{equation}
\begin{aligned}
|l(u) - l(v)|\\
= \Big\|\big(g \circ \delta_{\Omega}(u) - g \circ \delta_{\Omega}(y)\big)^2 
- \big(g \circ \delta_{\Omega}(v) - g \circ \delta_{\Omega}(y)\big)^2\Big\| \\
= \Big\| \big(g \circ \delta_{\Omega}(u) - g \circ \delta_{\Omega}(v)\big) \\
\underbrace{\big(g \circ \delta_{\Omega}(u) 
\AMsup{+}
g \circ \delta_{\Omega}(v) 
\AMsup{- 2}
g \circ \delta_{\Omega}(y)\big)}_{\leq 4 \text{ as } \|g\| \leq 1} \Big\|
\end{aligned}
\end{equation}
Then,
\begin{equation}
\begin{aligned}
|l(u) - l(v)| &\leq 4 \|g \circ \delta_{\Omega}(u) - g \circ \delta_{\Omega}(v)\| \\
|l(u) - l(v)| &\leq 4 \|g'\|_{\infty} \|\delta\|_{\infty} \|u - v\|
\end{aligned}
\end{equation}
We have \(\|\delta\|_{\infty} \leq 1\) by the Eikonal equation in Definition~\ref{def:definition_2}. Then:
\begin{equation}
\begin{aligned}
|l(u) - l(v)| &\leq \underbrace{4K_{1}}_{=\alpha} \|u - v\|
\end{aligned}
\end{equation}
\noindent
\textbf{Smooth:} We want to prove that:
\begin{equation}
\begin{aligned}
\exists \beta \text{ s.t } \forall u,v \in \Omega, \|\nabla l(u) - \nabla l(v)\| \leq \beta \|u-v\|
\end{aligned}
\end{equation}

Computing the gradient of $l$ gives:
\begin{equation}
\begin{aligned}
\nabla l(u) &= 2\big(g \circ \delta_{\Omega}(u) - g \circ \delta_{\Omega}(y)\big) \nabla \delta_{\Omega}(u) g'\big(\delta_{\Omega}(u)\big)
\end{aligned}
\end{equation}
Then we introduce intermediate values that cancel each other in the gradient differences to be able to compute a bound:
\begin{equation}
\begin{aligned}
\|\nabla l(u) - \nabla l(v)\| =\\
\Big\| 2\big(g \circ \delta_{\Omega}(u) - g \circ \delta_{\Omega}(y)\big) \nabla \delta_{\Omega}(u) g'\big(\delta_{\Omega}(u)\big) \\
\quad - 2\big(g \circ \delta_{\Omega}(v) - g \circ \delta_{\Omega}(y)\big) \nabla \delta_{\Omega}(u) g'\big(\delta_{\Omega}(u)\big) \\
\quad + 2\big(g \circ \delta_{\Omega}(v) - g \circ \delta_{\Omega}(y)\big) \nabla \delta_{\Omega}(u) g'\big(\delta_{\Omega}(u)\big) \\
\quad - 2\big(g \circ \delta_{\Omega}(v) - g \circ \delta_{\Omega}(y)\big) \nabla \delta_{\Omega}(v) g'\big(\delta_{\Omega}(u)\big) \\
\quad + 2\big(g \circ \delta_{\Omega}(v) - g \circ \delta_{\Omega}(y)\big) \nabla \delta_{\Omega}(v) g'\big(\delta_{\Omega}(u)\big) \\
\quad - 2\big(g \circ \delta_{\Omega}(v) - g \circ \delta_{\Omega}(y)\big) \nabla \delta_{\Omega}(v) g'\big(\delta_{\Omega}(v)\big) \Big\|
\end{aligned}
\end{equation}
We have, for each part of the sum:
\begin{equation}
\begin{aligned}
\Big\| 2\big(g \circ \delta_{\Omega}(u) - g \circ \delta_{\Omega}(y)\big) \nabla \delta_{\Omega}(u) g'\big(\delta_{\Omega}(u)\big)\\
- 2\big(g \circ \delta_{\Omega}(v) - g \circ \delta_{\Omega}(y)\big) \nabla \delta_{\Omega}(u) g'\big(\delta_{\Omega}(u)\big) \Big\|\\
\leq 2\alpha K_1 \|u - v\|\\
\Big\| 2\big(g \circ \delta_{\Omega}(v) - g \circ \delta_{\Omega}(y)\big) \nabla \delta_{\Omega}(u) g'\big(\delta_{\Omega}(u)\big)\\
- 2\big(g \circ \delta_{\Omega}(v) - g \circ \delta_{\Omega}(y)\big) \nabla \delta_{\Omega}(v) g'\big(\delta_{\Omega}(u)\big) \Big\|\\
\leq 4K_1 \|\nabla \delta_{\Omega}(u) - \nabla\delta_{\Omega}(v)\|\\
\Big\| 2\big(g \circ \delta_{\Omega}(v) - g \circ \delta_{\Omega}(y)\big) \nabla \delta_{\Omega}(v) g'\big(\delta_{\Omega}(u)\big)\\
- 2\big(g \circ \delta_{\Omega}(v) - g \circ \delta_{\Omega}(y)\big) \nabla \delta_{\Omega}(v) g'\big(\delta_{\Omega}(u)\big) \Big\|\\
\leq 4 K_2 \|u - v\|
\end{aligned}
\end{equation}
\noindent
In order to bound the second term, let \(x = \Pi(u)\) be the orthogonal projection of \(u\) on \(\Omega\). And let \(\kappa(x)\) be the mean curvature at \(x\). Note that \(\Omega\) being convex, we have \(0 \leq \kappa(x)\). Then the Hessian of \(\delta_{\Omega}\) at \(u\) is diagonalizable in the surface tangent and normal orthonormal basis :

\begin{equation}
\begin{aligned}
H_{\delta_{\Omega}}(u) = \begin{pmatrix}
0 & 0 \\
0 & \frac{\kappa(x)}{1 + \delta_{\Omega}(u)\kappa(x)}
\end{pmatrix}
\end{aligned}
\end{equation}
Let us define the maximal curvature:
\begin{equation}
\begin{aligned}
K_3 = \max_{x\in\partial\Omega} \kappa(x)
\end{aligned}
\end{equation}
Then,

\begin{equation}
\begin{aligned}
\frac{\kappa(x)}{1 + \delta_{\Omega}(u)\kappa(x)} \leq K_3
\end{aligned}
\end{equation}
We can therefore bound the second term:
\begin{equation}
\begin{aligned}
\|\nabla \delta_{\Omega}(u) - \nabla \delta_{\Omega}(v) \| \leq K_{3}\|u -v |\|
\end{aligned}
\end{equation}
and then bound the gradient different:
\begin{equation}
\begin{aligned}
\|\nabla l(u) - \nabla l(v)\| \leq \underbrace{(2\alpha K_1 + 4K_2 + 4K_1K_3)}_{=\beta} \|u - v\|
\end{aligned}
\end{equation}
\textbf{Strictly convex:} We want to prove that the Hessian of \(l\) is definite positive

\begin{equation}
\begin{aligned}
\nabla l(u) &= 2\underbrace{\big(g \circ \delta_{\Omega}(u) - g \circ \delta_{\Omega}(y)\big)}_{=\epsilon} \nabla \delta_{\Omega}(u) g'\big(\delta_{\Omega}(u)\big)\\
\end{aligned}
\end{equation}
and
\begin{equation}
\begin{aligned}
H_{l}(u) = 2g'(\delta_{\Omega}(u))^2\nabla\delta_{\Omega}^T \nabla \delta_{\Omega}\\
+ 2\epsilon g''(\delta_{\Omega}(u))\nabla\delta_{\Omega}^T \nabla \delta_{\Omega}\\
+ 2\epsilon g'(\delta_{\Omega}(u)) H_{\delta_{\Omega}}(u)
\end{aligned}
\end{equation}
\noindent
In the surface tangent, normal orthonormal basis, it can be written as follows:

\begin{equation}
\begin{aligned}
H_{l}(u) = \begin{pmatrix}
\underbrace{2g'(\delta_{\Omega}(u))^2 + 2\epsilon g''(\delta_{\Omega}(u))}_{\lambda_1} & 0 \\
0 & \underbrace{2\epsilon g'(\delta_{\Omega}(u))}_{\lambda_2}
\end{pmatrix}
\end{aligned}
\end{equation}
\noindent
If we add a regularization term, then it becomes:

\begin{equation}
\begin{aligned}
H_{l}(u) = \begin{pmatrix}
2\eta + \lambda_{1} & 0 \\
0 & 2\eta + \lambda_{2}
\end{pmatrix}
\end{aligned}
\end{equation}
\noindent
If the prediction \(u\) and the ground truth \(y\) are close enough, then \(\epsilon\) is low enough and:

\begin{equation}
\begin{aligned}
0 < 2\eta + 2g'(\delta_{\Omega}(u))^2 + \underbrace{2\epsilon g''(\delta_{\Omega}(u))}_{\sim 0}\\
0 < 2\eta + \underbrace{2\epsilon g'(\delta_{\Omega}(u))}_{\sim 0}
\end{aligned}
\end{equation}

\subsection{A toy experiment for maximal variance}
\label{sec:appendix:theoretical:toy_variance}
\begin{figure}[t]
\centering
\setlength{\tabcolsep}{1pt}  
\renewcommand{\arraystretch}{0.5}  
\begin{tabular}{ccccc}
    \begin{subfigure}[b]{0.19\linewidth}
        \includegraphics[width=\linewidth]{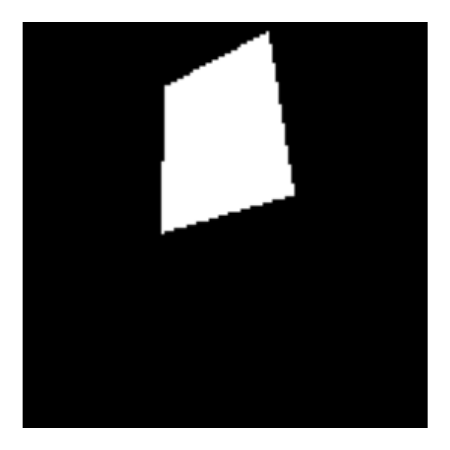}
    \end{subfigure} &
    \begin{subfigure}[b]{0.19\linewidth}
        \includegraphics[width=\linewidth]{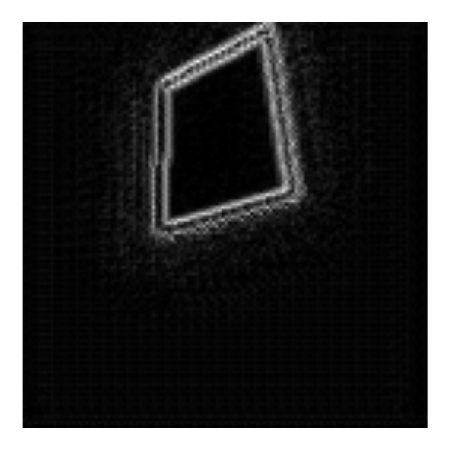}
    \end{subfigure} &
    \begin{subfigure}[b]{0.19\linewidth}
        \includegraphics[width=\linewidth]{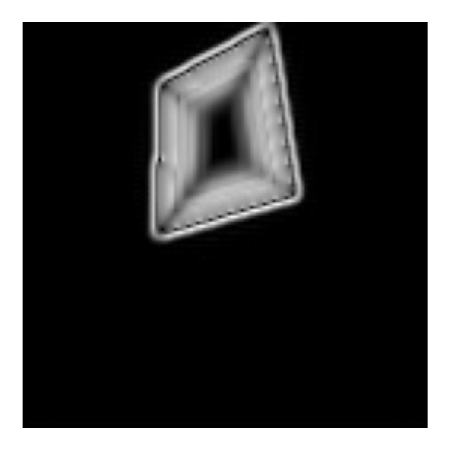}
    \end{subfigure}  &
    \begin{subfigure}[b]{0.19\linewidth}
        \includegraphics[width=\linewidth]{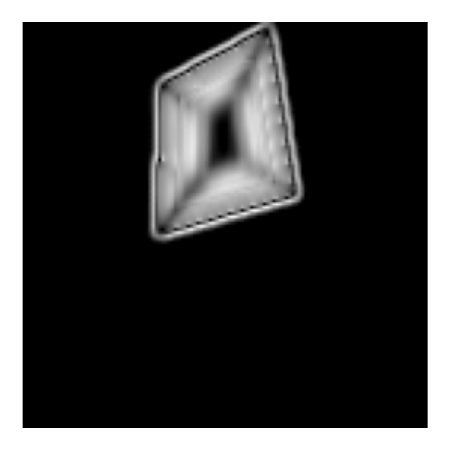}
    \end{subfigure}
    \begin{subfigure}[b]{0.19\linewidth}
        \includegraphics[width=\linewidth]{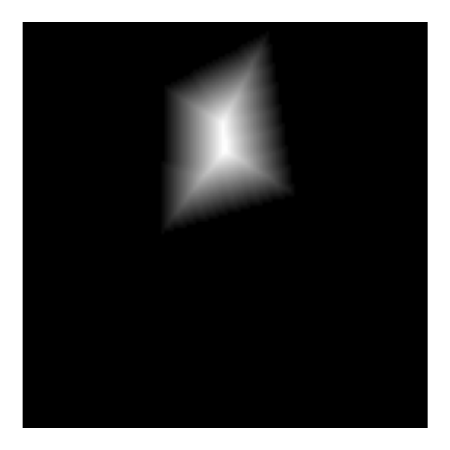}
    \end{subfigure} &\\

    \begin{subfigure}[b]{0.19\linewidth}
        \includegraphics[width=\linewidth]{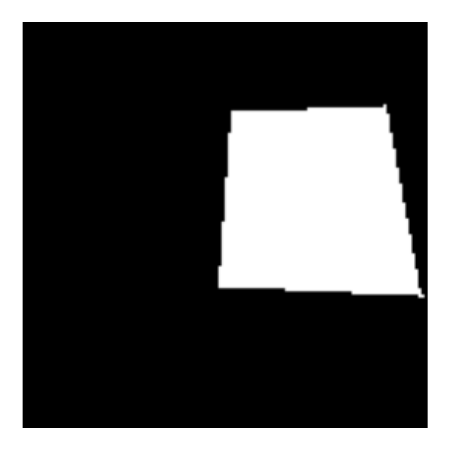}
    \end{subfigure} &
    \begin{subfigure}[b]{0.19\linewidth}
        \includegraphics[width=\linewidth]{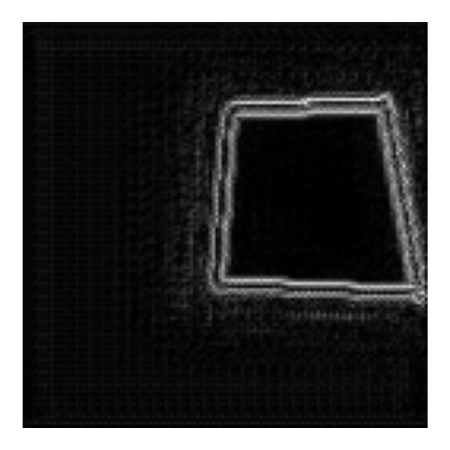}
    \end{subfigure} &
    \begin{subfigure}[b]{0.19\linewidth}
        \includegraphics[width=\linewidth]{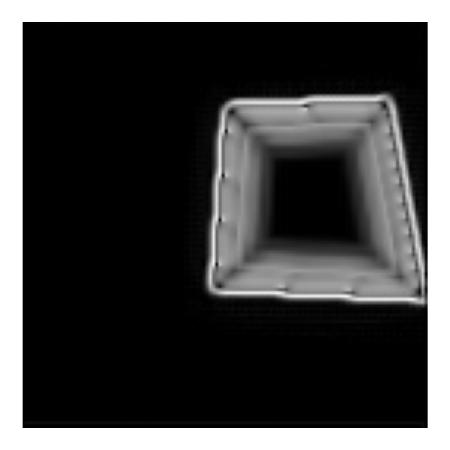}
    \end{subfigure}  &
    \begin{subfigure}[b]{0.19\linewidth}
        \includegraphics[width=\linewidth]{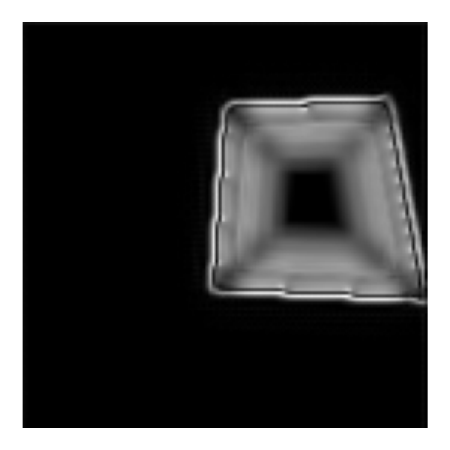}
    \end{subfigure}
    \begin{subfigure}[b]{0.19\linewidth}
        \includegraphics[width=\linewidth]{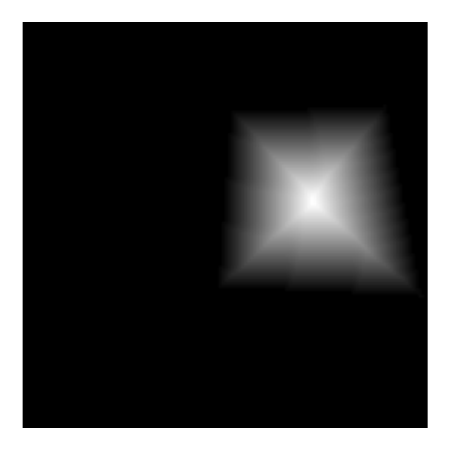}
    \end{subfigure} &\\

    \begin{subfigure}[b]{0.19\linewidth}
        \includegraphics[width=\linewidth]{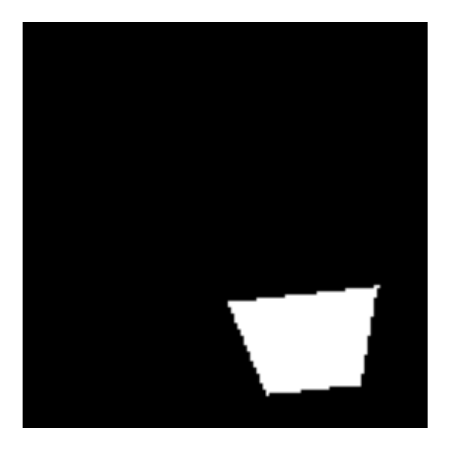}
    \end{subfigure} &
    \begin{subfigure}[b]{0.19\linewidth}
        \includegraphics[width=\linewidth]{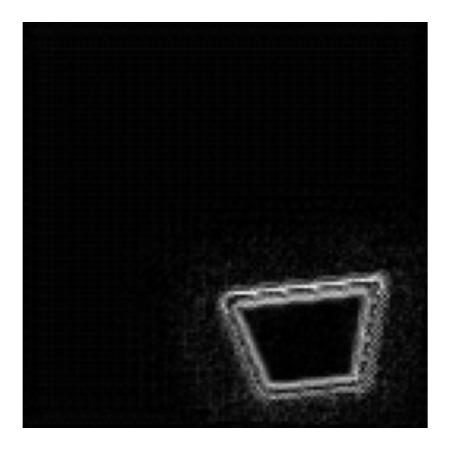}
    \end{subfigure} &
    \begin{subfigure}[b]{0.19\linewidth}
        \includegraphics[width=\linewidth]{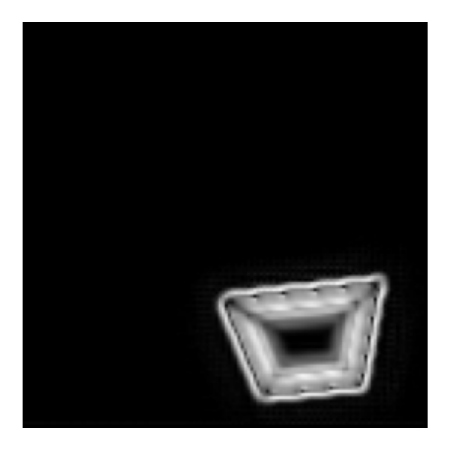}
    \end{subfigure}  &
    \begin{subfigure}[b]{0.19\linewidth}
        \includegraphics[width=\linewidth]{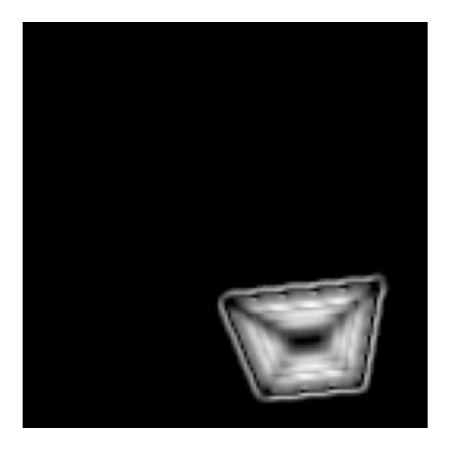}
    \end{subfigure}
    \begin{subfigure}[b]{0.19\linewidth}
        \includegraphics[width=\linewidth]{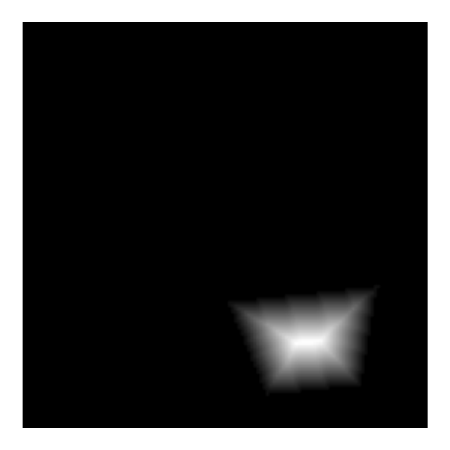}
    \end{subfigure} &\\

    \begin{subfigure}[b]{0.19\linewidth}
        \includegraphics[width=\linewidth]{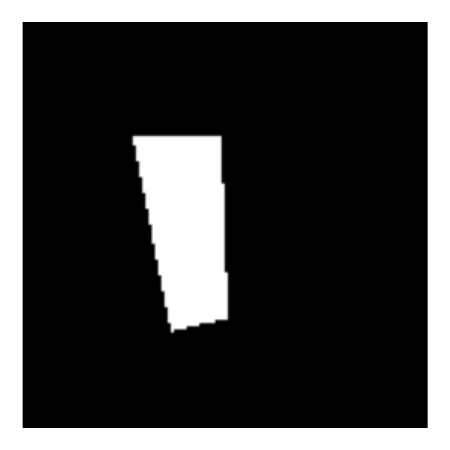}
    \end{subfigure} &
    \begin{subfigure}[b]{0.19\linewidth}
        \includegraphics[width=\linewidth]{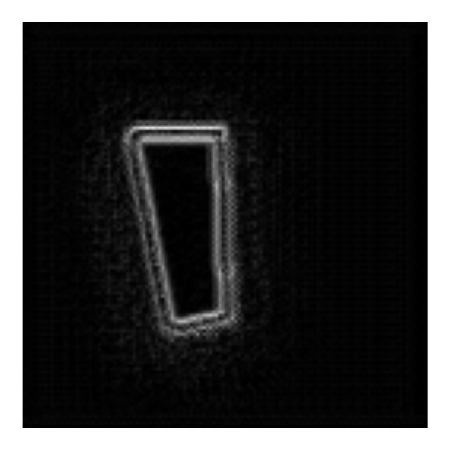}
    \end{subfigure} &
    \begin{subfigure}[b]{0.19\linewidth}
        \includegraphics[width=\linewidth]{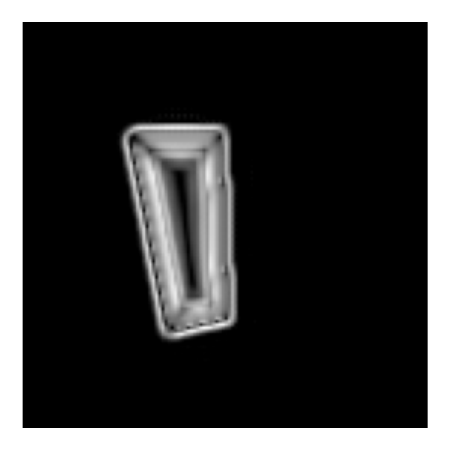}
    \end{subfigure}  &
    \begin{subfigure}[b]{0.19\linewidth}
        \includegraphics[width=\linewidth]{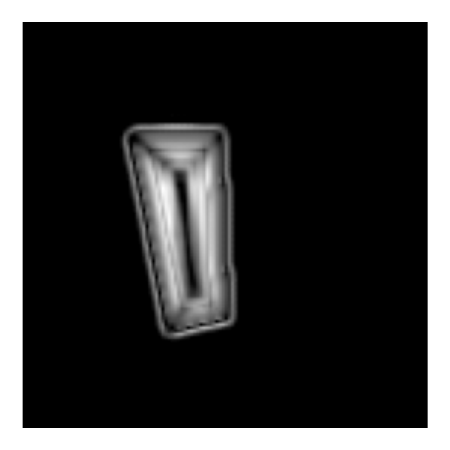}
    \end{subfigure}
    \begin{subfigure}[b]{0.19\linewidth}
        \includegraphics[width=\linewidth]{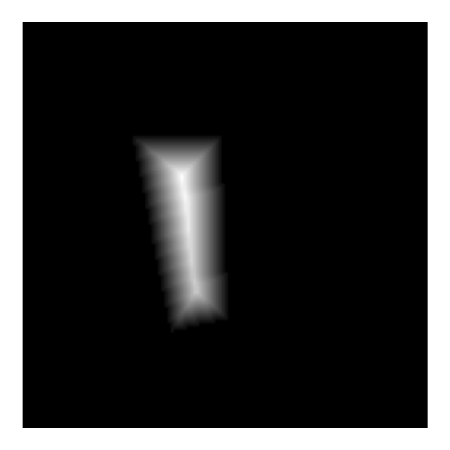}
    \end{subfigure} &\\

    \begin{subfigure}[b]{0.19\linewidth}
        \includegraphics[width=\linewidth]{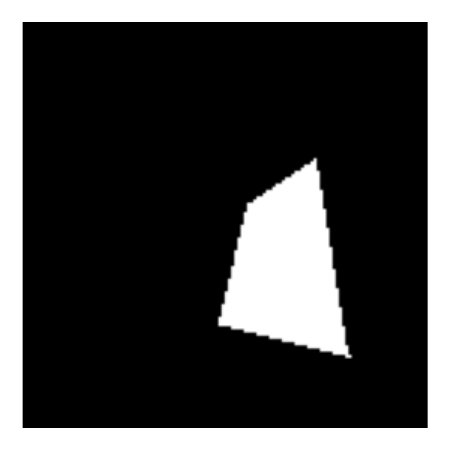}
    \end{subfigure} &
    \begin{subfigure}[b]{0.19\linewidth}
        \includegraphics[width=\linewidth]{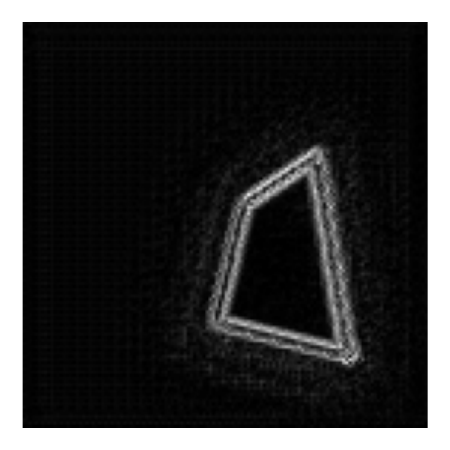}
    \end{subfigure} &
    \begin{subfigure}[b]{0.19\linewidth}
        \includegraphics[width=\linewidth]{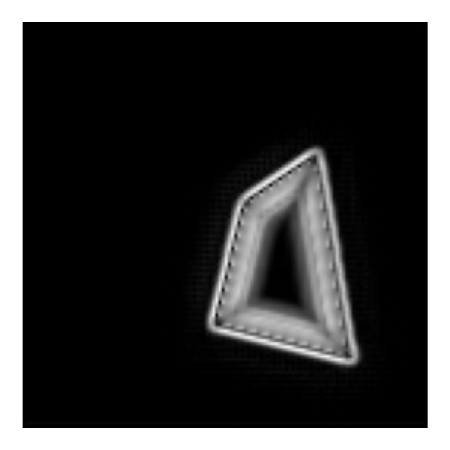}
    \end{subfigure} &
    \begin{subfigure}[b]{0.19\linewidth}
        \includegraphics[width=\linewidth]{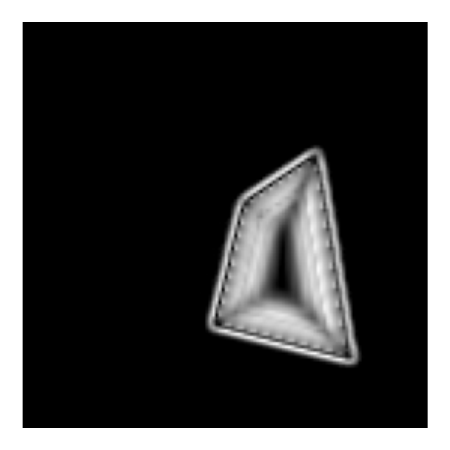}
    \end{subfigure}
    \begin{subfigure}[b]{0.19\linewidth}
        \includegraphics[width=\linewidth]{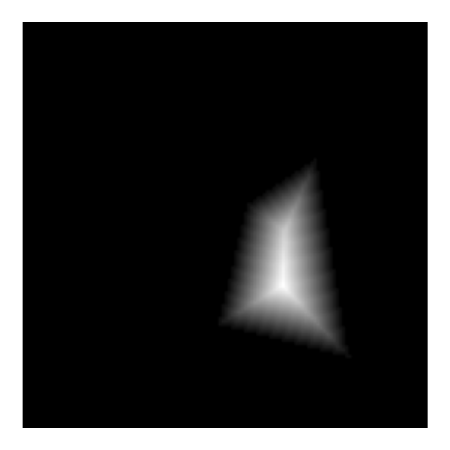}
    \end{subfigure} &\\

    \begin{subfigure}[b]{0.19\linewidth}
        \includegraphics[width=\linewidth]{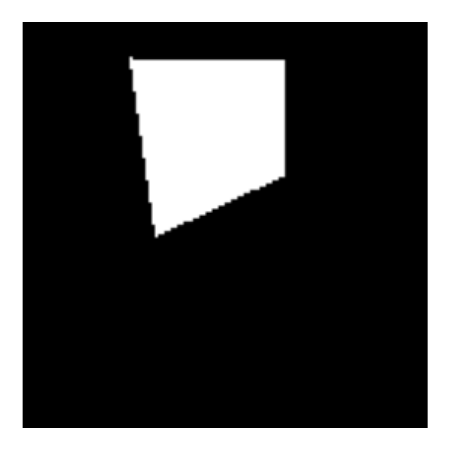}
    \end{subfigure} &
    \begin{subfigure}[b]{0.19\linewidth}
        \includegraphics[width=\linewidth]{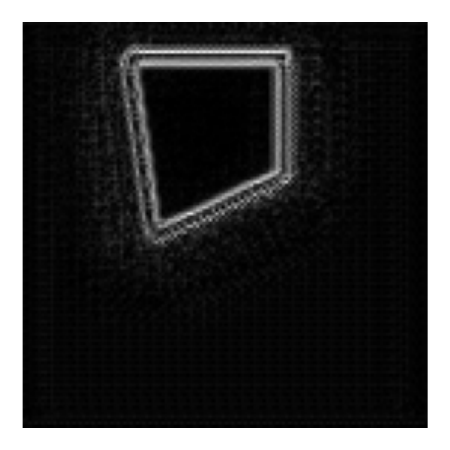}
    \end{subfigure} &
    \begin{subfigure}[b]{0.19\linewidth}
        \includegraphics[width=\linewidth]{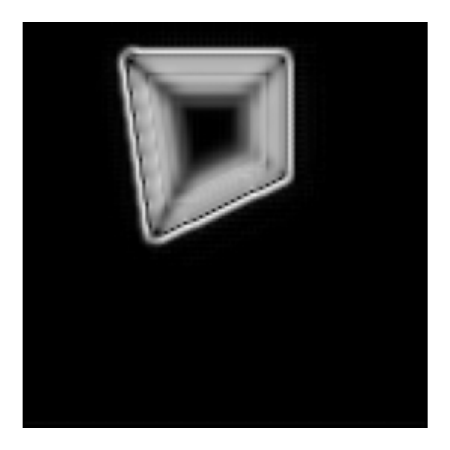}
    \end{subfigure} &
    \begin{subfigure}[b]{0.19\linewidth}
        \includegraphics[width=\linewidth]{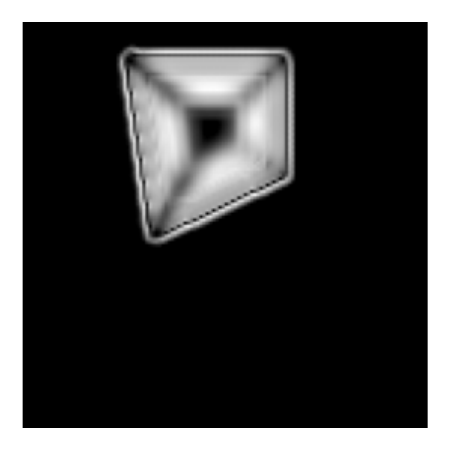}
    \end{subfigure}
    \begin{subfigure}[b]{0.19\linewidth}
        \includegraphics[width=\linewidth]{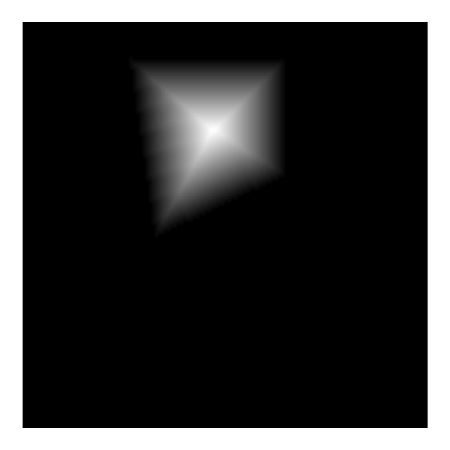}
    \end{subfigure}
    
\end{tabular}
\caption{Results of the U-Net training. From left to right: 
(i) Input shape, (ii) Gradient 
\antoine{norm}
of the U-Net output after 1 epoch, (iii) after 5 epochs, (iv) after 10 epochs, (v) Distance transform. We show the gradient 
\antoine{norm}
of the output for better visualization of the convergence process. The gradient of the resulting image is uniform except on the medial axis (a.k.a. skeleton), thus corresponding to the distance transform.}
\label{fig:appendix:convergence_polygons}
\end{figure}

\begin{figure*}[htb]  
    \centering
    \includegraphics[width=1.0\textwidth]{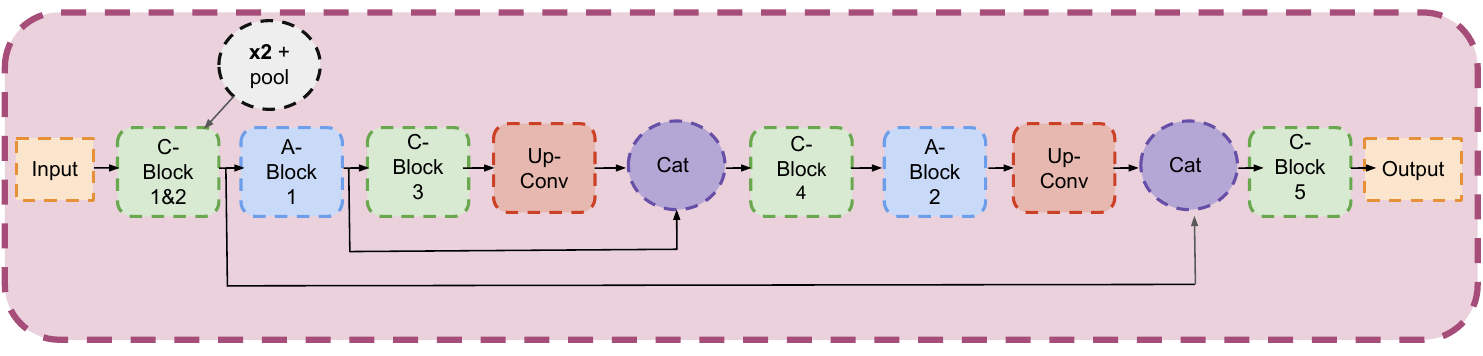}  
    \caption{U-Net architecture used for our toy experiment on the maximal variance. Details of C-Blocks and A-Blocks given in Table~\ref{tab:appendix:blocks}.}
\label{fig:appendix:u_net}
\end{figure*}

Let us now show that the maximal variance property mentioned in ~\ref{sec:preambule:max_variance} can be retrieved with a simple experiment.\\
We built a large dataset comprising 10\,000 polygons \(I\) of shape \(64\times64\) \AMsup{(see examples on Figure}
\ref{fig:appendix:convergence_polygons}). Then we trained a U-Net architecture \(\Theta\) detailed in Figure~\ref{fig:appendix:u_net} and Table~\ref{tab:appendix:blocks} \cite{ronneberger2015u} with self-attention \cite{vaswani2017attention} to maximize the variance in polygon areas while respecting translation, flip and rotation invariance. We also promoted low Laplacian values to satisfy the Eikonal equation and avoid clustering patterns as mentioned previously. Training was conducted for 50 epochs, with a learning rate of \(lr=0.01\) and a batch size of 16, optimizing the following loss function:

\begin{equation} 
    \begin{aligned} 
    \mathcal{L} = \lambda_{\text{repro}}\mathcal{L}_{\text{repro}} + \lambda_{\text{var}}\mathcal{L}_{\text{var}}\\
    + \lambda_{\text{laplacian}}\mathcal{L}_{\text{laplacian}} + \lambda_{\text{control}}\mathcal{L}_{\text{control}}
    \end{aligned}
\end{equation}
with
\begin{equation} 
    \begin{aligned} 
    \mathcal{L}_{\text{repro}} = \|\Theta(I) - T^{-1}\big(\Theta\left(T(I)\right)\big)\|
    \end{aligned}
\end{equation}
Here \(T\) is chosen to be either a translation, rotation, or flip. The objective of this loss is to best satisfy the constancy assumption.

\begin{equation} 
    \begin{aligned} 
    \mathcal{L}_{\text{var}} = \frac{\text{mean}(I)}{\text{var}(I)}\\
    \mathcal{L}_{\text{laplacian}} = \|\Delta I\|_1\\
    \mathcal{L}_{\text{control}} = \|I\|_1
    \end{aligned}
\end{equation}
\noindent
Here \(\mathcal{L}_{\text{var}} \) is used to promote variance within the polygon. A normalization by the mean is applied to prevent large values. \(\mathcal{L}_{\text{laplacian}}\) helps to approximate the Eikonal equation. Finally, we found that introducing \(\mathcal{L}_{\text{control}}\) improves training stability..\\
\noindent
We set the following values: \(\lambda_{\text{repro}}=1\), \(\lambda_{\text{var}}=1\), \(\lambda_{\text{laplacian}}=0.1\), and \(\lambda_{\text{control}}=0.01\). 
\noindent
Our experiment shows that \(\Theta\) indeed converges towards the distance transform solution as seen on Figure~\ref{fig:appendix:convergence_polygons}.

\begin{table}[ht]
\centering
\begin{minipage}{\columnwidth}  
\begin{tabular}{|p{\columnwidth}|}  
\hline
\rowcolor{green!20}  
\textbf{C-Block(in, out, dilation)} \\
\hline
Conv2D(in\_ch=in, out\_ch=out, kernel\_size=3, dilation=dilation, padding=dilation), \\
BatchNorm2d(out\_ch=out), \\
ReLU, \\
Conv2D(in\_ch=out, out\_ch=out, kernel\_size=3, dilation=dilation, padding=dilation), \\
BatchNorm2d(out\_ch=out), \\
ReLU \\
\hline
\end{tabular}
\end{minipage}

\vspace{-1pt}  

\begin{minipage}{\columnwidth}  
\begin{tabular}{|p{\columnwidth}|}  
\hline
\rowcolor{blue!20}  
\textbf{A-Block(in)} \\
\hline
f=Conv2D(in\_ch=in, out\_ch=in / 8, kernel\_size=1) \\
g=Conv2D(in\_ch=in, out\_ch=in / 8, kernel\_size=1) \\
h=Conv2D(in\_ch=in, out\_ch=in, kernel\_size=1) \\
s=Softmax(dim=-1) \\
\hline
\end{tabular}
\end{minipage}

\vspace{-1pt}  

\begin{minipage}{\columnwidth}  
\begin{tabular}{|p{\columnwidth}|}  
\hline
\textbf{Blocks} \\
\hline
C-Block 1 = C-Block(1, 32, 1) \\
C-Block 2 = C-Block(32, 64, 2) \\
C-Block 3 = C-Block(64, 128, 1) \\
C-Block 4 = C-Block(128, 64, 2) \\
C-Block 5 = C-Block(64, 32, 1) + Conv2D(in\_ch=32, out\_ch=1, kernel\_size=1)\\
\hline
A-Block 1 = A-Block(64)\\
A-Block 2 = A-Block(64)\\
\hline
\end{tabular}
\end{minipage}

\caption{Details of blocks used for our UNet architecture, see diagram of our architecture on Figure~\ref{fig:appendix:u_net}.}
\label{tab:appendix:blocks}
\end{table}

\subsection{A toy experiment for convergence}
\label{sec:appendix:theoretical:toy_convergence}

\begin{figure}[t]
\centering
\setlength{\tabcolsep}{0pt}
\renewcommand{\arraystretch}{1.0}
\includegraphics[width=0.44\linewidth]{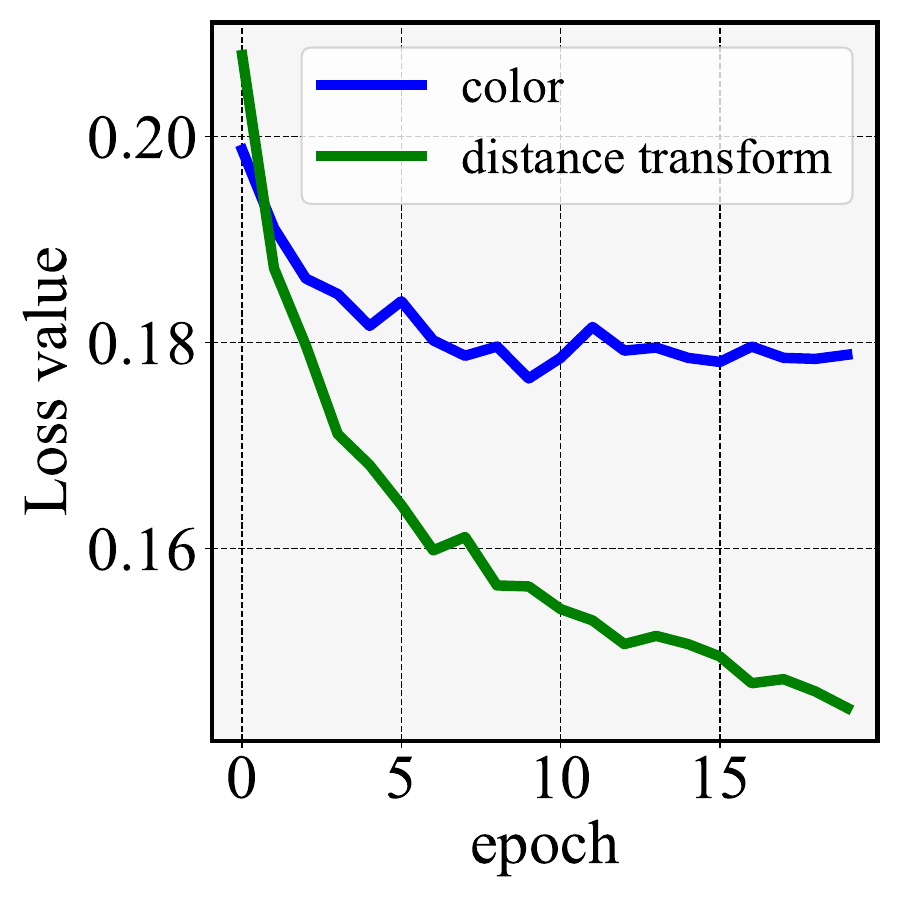}\includegraphics[width=0.44\linewidth]{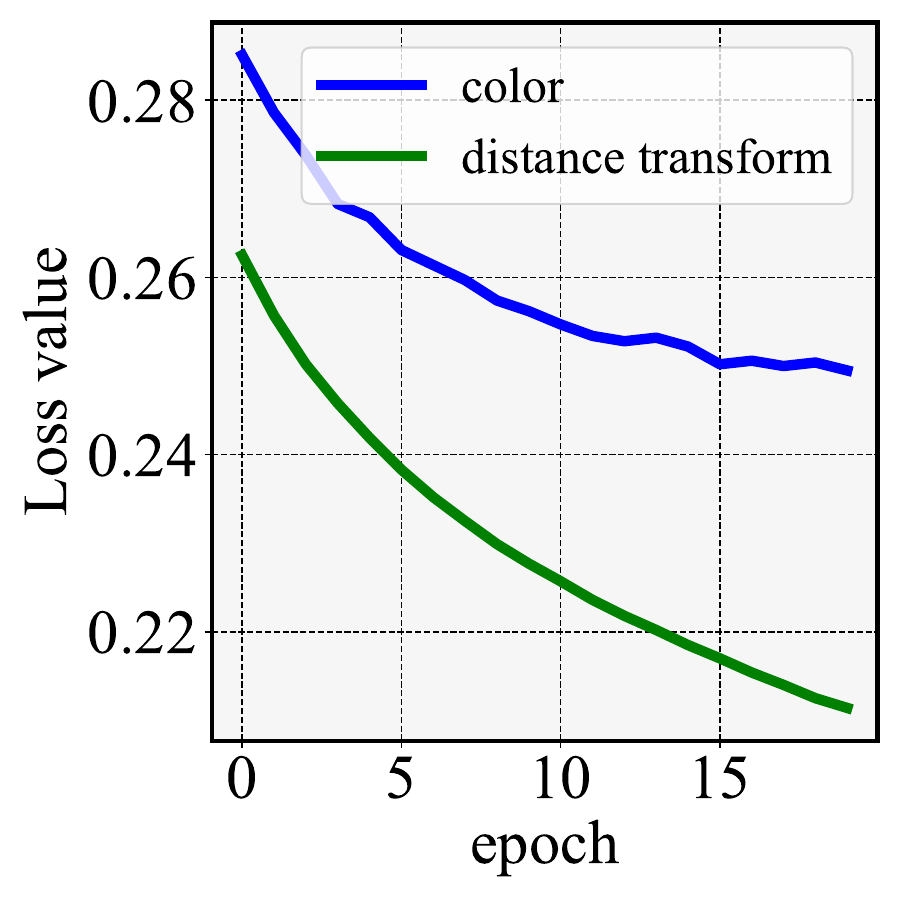}
\footnotesize{Training loss for translation prediction for rectangles (left) }\\
\footnotesize{and stars (right)}\\
\hspace{0.5cm}\includegraphics[width=0.33\linewidth]{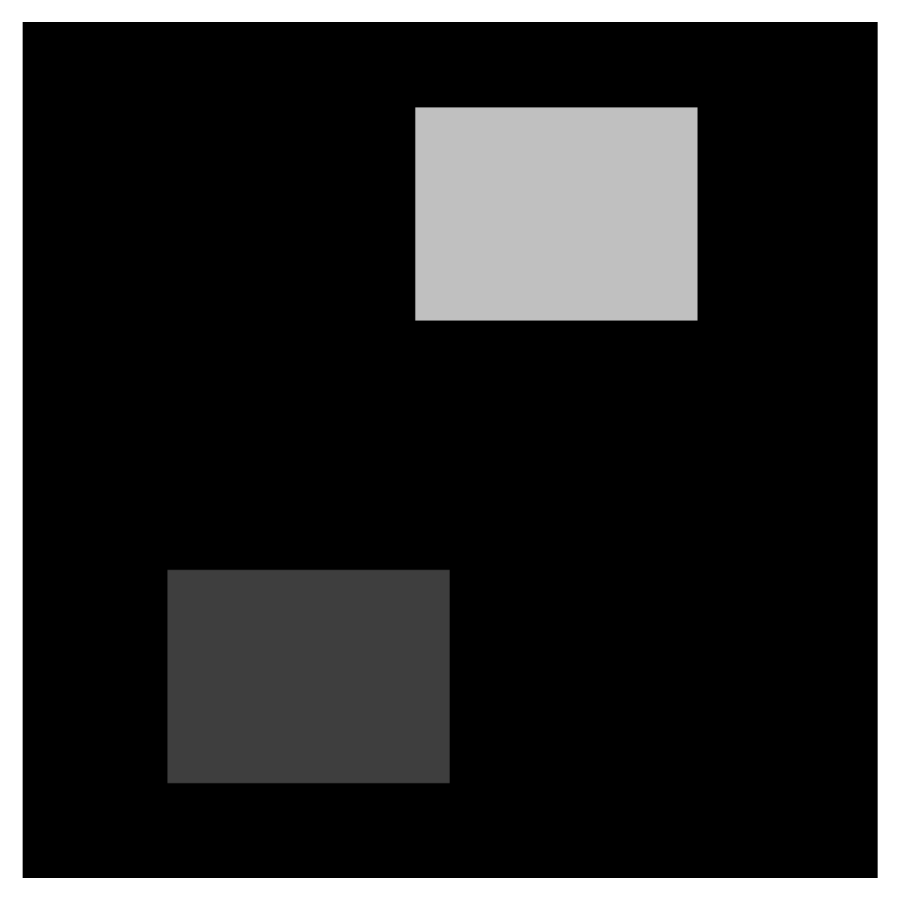} \hspace{0.5cm}\includegraphics[width=0.33\linewidth]{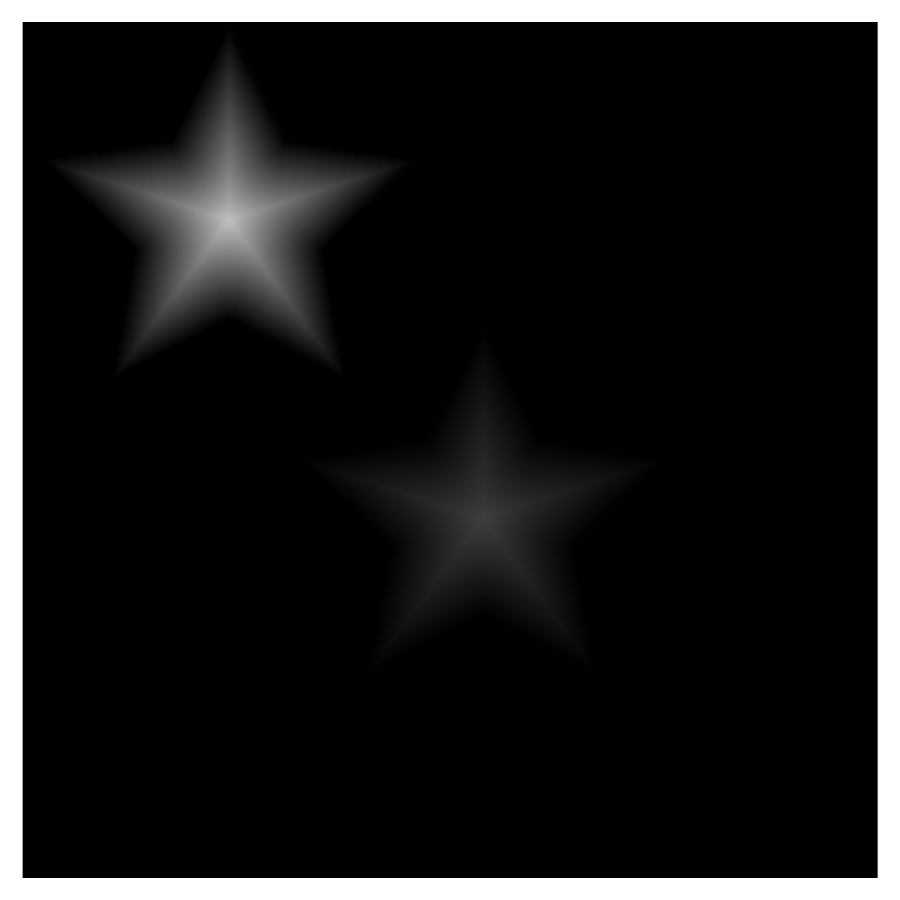}\\
\footnotesize{Translation of two uniform rectangles (left) and stars with distance transform (right)}\\
\caption{Convergence study for a translation prediction model with and without the distance transform.}
\label{fig:appendix:convergence_study}
\end{figure}

\begin{figure}[t]
\centering
\setlength{\tabcolsep}{0pt}
\renewcommand{\arraystretch}{1.0}
\hspace{0.5cm}\includegraphics[width=0.33\linewidth]{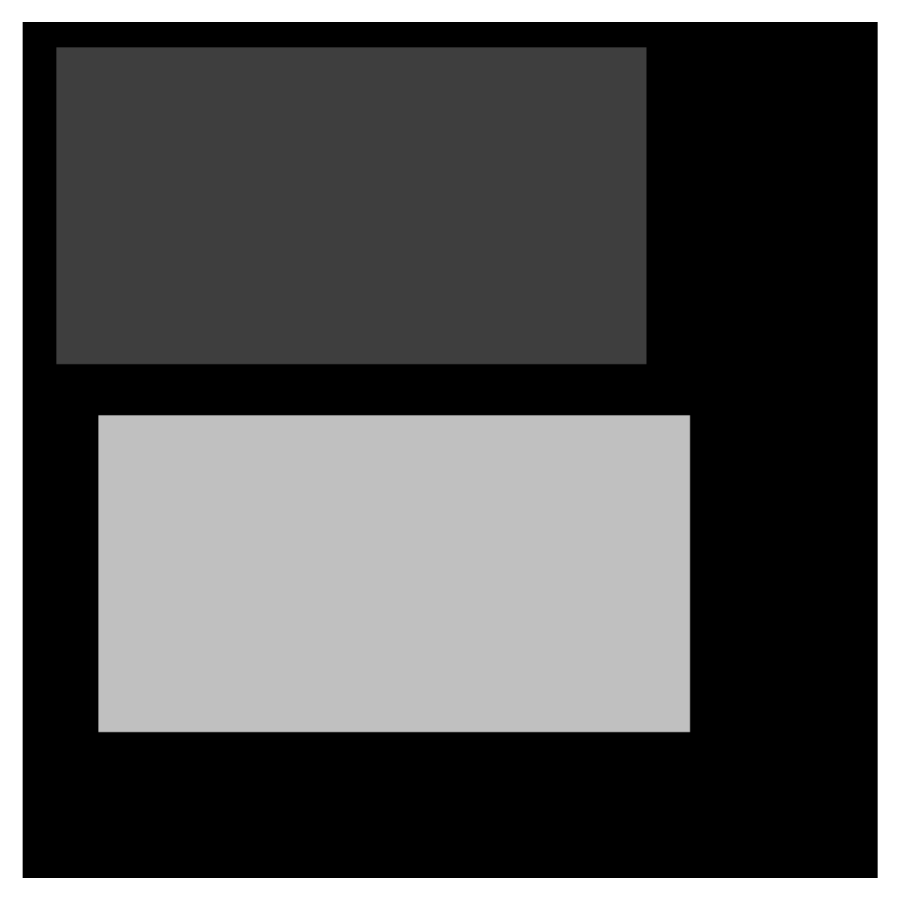} \hspace{0.5cm}\includegraphics[width=0.33\linewidth]{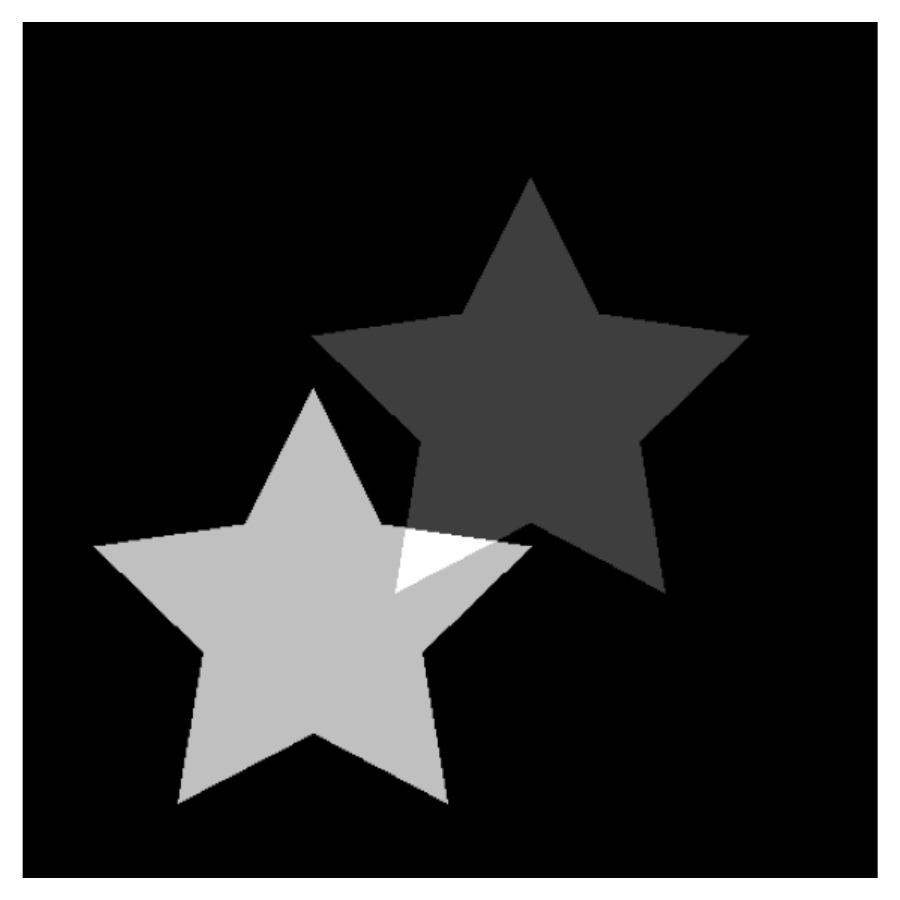}\\
\footnotesize{Colour dataset with rectangles (left) and stars (right). Lighter shapes are before the random translation.}\\
\hspace{0.5cm}\includegraphics[width=0.33\linewidth]{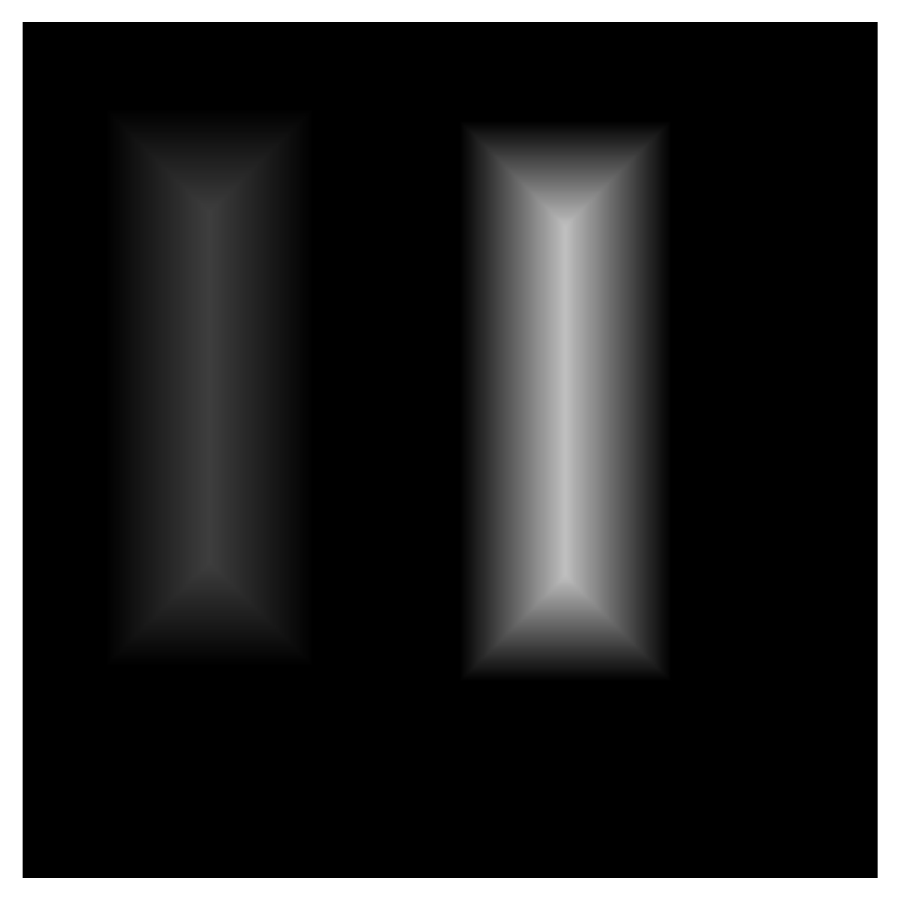} \hspace{0.5cm}\includegraphics[width=0.33\linewidth]{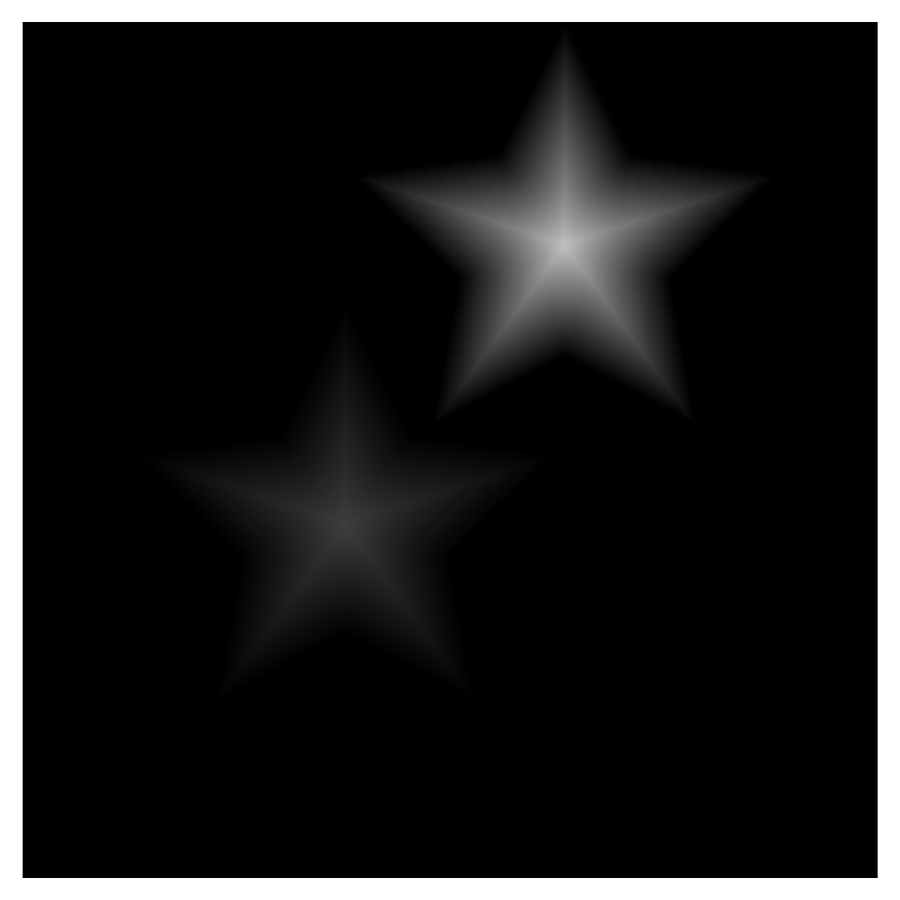}\\
\footnotesize{Distance transform dataset with rectangles (left) and stars (right).}
\caption{Example of images used in the experiment to analyse convergence properties.}
\label{table:appendix:convergence_dataset}
\end{figure}

Let us now show that the improved convergence property mentioned in ~\ref{sec:premabule:convergence_properties} can be retrieved with a simple experiment.\\
We built \(n=10\,000\) pairs of rectangles \(\big\{\big(R^{i}_{0}, R^{i}_{1}\big)\big\}_{i\in\{1,...,n\}}\) each of shape \(500\times500\) with \(R^{i}_{1}\) being  \(R^{i}_{0}\) after a random rigid translation \(\Delta_{i} = \left(u_{i}, v_{i}\right)\). We trained a multi-layer perceptron detailed in Table~\ref{table:appendix:perceptron} to take as input the pair of rectangle images and to predict the shift \(\Delta\). Training was done for 20 epochs, using a learning rate of 0.01 and a batch size of 32. We discarded any random translation that caused parts of the rectangle to fall outside the boundaries. We 
\AMsup{considered two}
scenarios: first, rectangles are only filled with a white colour. Second,
rectangles are filled with the distance transform values. We re-iterated this procedure with stars instead of rectangles. Some images of the dataset are displayed on Figure~\ref{table:appendix:convergence_dataset}. Figure~\ref{fig:appendix:convergence_study} show that the convergence on the training is much faster with the distance transform.

\begin{table}[ht]
\centering
\begin{tabular}{|p{\columnwidth}|}  
\hline
\rowcolor{orange!20}  
\textbf{Multi-layer perceptron} \\
\hline
Linear(in\_features=\(2\times500\times500\), out\_features=128) \\
Linear(in\_features=128, out\_features=64) \\
Linear(in\_features=64, out\_features=2) \\
\hline
\end{tabular}
\caption{Neural Network Layer Definitions for SelfAttention Class}
\label{table:appendix:perceptron}
\end{table}

\section{Extra Results}
\label{sec:appendinx:extra_results}
\subsection{Depth}
\label{sec:appendix:extra_results:depth}
We give extra quantitative results of our method in Table~\ref{table:appendix:results_depth}. As expected, when we increase the input resolution from \(256\times832\) to \(320\times1024\) and apply the online refinement procedure of \cite{casser2019depth}, we improve even more the metrics.

\definecolor{lightblue}{rgb}{0.1, 0.4, 0.6}

\begin{table}[ht]
\centering
\renewcommand{\arraystretch}{1.5} 
\fontsize{16}{18}\selectfont 
\resizebox{\linewidth}{!}{ 
\begin{tabular}{ r c c c c c c c }
\hline
\multirow{2}{*}{\textbf{Method}} 
& \multicolumn{4}{c}{\textit{Lower is better} \(\downarrow\)} 
& \multicolumn{3}{c}{\textit{Higher is better} \(\uparrow\)} \\  
\cmidrule(r){2-5} \cmidrule(l){6-8} 
& \textcolor{lightblue}{Abs Rel} & Sq Rel & RMSE & \textcolor{lightblue}{RMSE log} 
& \hspace{15pt}\textcolor{lightblue}{\(\delta_{1}\)} & \hspace{15pt}\(\delta_{2}\) &  \hspace{15pt}\(\delta_{3}\) \\
\hline

Ours HR\(\left(320 \times 1024\right)\) & 0.101 & 0.703 & 4.422 & 0.176 & \hspace{15pt}0.895 &  \hspace{15pt}0.963 & \hspace{15pt}0.984\\
\(\text{Ours HR}^{*}\) & 0.083 & 0.655 & 4.11 & 0.166 & \hspace{15pt}0.916 &  \hspace{15pt}0.965 & \hspace{15pt}
0.984\\
\hline
\end{tabular}
}
\caption{\textbf{Results of depth estimations on KITTI 2015}.\\
*: Using the online refinement technique of \cite{casser2019depth}.}
\end{table}
\noindent
\label{table:appendix:results_depth}

\noindent
We also provide more depth images in Figure~\ref{fig:appendix:qualitative_depth_results}. In general, our method renders very sharp depth images, which is the sign that our pipeline indeed reduces the ill-posed nature of the optimization problem.

\subsection{Flow}
\label{sec:appendix:extra_results:flow}
Our method uses the strategy of \cite{hariat2023rebalancing} to remove moving pixels from the computation of the photometric loss. It takes advantage of a self-supervised optical flow network. Quantitative results are given in Table~\ref{table:appendix:results_flow}. We can see that our improved framework also improves flow metrics. The flow is also trained using the variance augmented image. To assess optical flow we use the KITTI 2015 flow dataset containing 200 annotated training images as test images.

\begin{table}
\centering

\begin{tabular}{|c|c|c|}
\hline
Method&Noc&All\\
\hline
FlowNetS\cite{fischer2015flownet}&8.12&14.19\\
FlowNet2\cite{ilg2017flownet}&4.93&10.06\\
GeoNet\cite{yin2018geonet}&8.05&10.81\\
GLNet\cite{chen2019self}&4.86&8.35\\
CoopNet\cite{hariat2023rebalancing}&5.10&9.43\\
Ours&\textbf{4.59}&\textbf{7.82}\\
\hline
\end{tabular}
\caption{{\bf Optical Flow:} Average end point error (in pixels) for non occluded (Noc) and for all (All) pixels on the KITTI 2015 flow dataset.}
\end{table}
\label{table:appendix:results_flow}

\subsection{Odometry}
\label{sec:appendix:extra_results:odometry}
We give results for odometry in Table~\ref{table:appendix:results_odometry}. To assess odometry we use Sequence 9 and 10 of the KITTI Odometry dataset.

\begin{table*}[ht]
\centering
\begin{tabular}{|c|c|c|c|c|}
\hline
\multirow{2}{*}{Methods} & \multicolumn{2}{c|}{Seq. 09} & \multicolumn{2}{c|}{Seq. 10} \\
\cline{2-5}
& {\(t_{err}\left(\%\right)\)} & {\(r_{err}\left(^{\circ}/100m\right)\)}  & {\(t_{err}\left(\%\right)\)} & {\(r_{err}\left(^{\circ}/100m\right)\)} \\
\cline{1-5}
ORB-SLAM\cite{mur2015orb}&15.30&0.26&3.68&0.48\\
\hline
\text{Zhou et al.}\cite{zhou2017unsupervised}&17.84&6.78&37.91&17.78\\
\text{Bian et al.}\cite{bian2019unsupervised}&11.2&3.35&10.1&4.96\\
CoopNet\cite{hariat2023rebalancing}&8.42&2.66&7.29&\textbf{2.14}\\
Ours&\textbf{8.39}&\textbf{2.31}&\textbf{7.17}&2.81\\
\hline
\end{tabular}
\caption{\textbf{Odometry}: Average Translation and Rotation errors for sequence 09 and 10 of the KITTI Odometry Dataset. Bold indicate the best learning based method.}
\end{table*}
\label{table:appendix:results_odometry}

\section{Discussion on the distance transform}
\label{sec:appendix:discussion_distance}

\subsection{Algorithm}
\label{sec:appendix:discussion_distance:algo}

The
distance transform 
in the case of \(8\)-neighbours (\(d_{8}\)) 
is shown on Algorithm~\ref{alg:cap}.

\begin{algorithm}
\caption{Distance Transform \(d_8\)}\label{alg:cap}
\begin{algorithmic}[1]
\Require \( I \) image of size \( H \times W \), \( \mathcal{C} \) binary contour of \(I\)
\State Initialize \( F(i,j) = \infty \) for all \( (i,j) \in I \)
\State Forward pass:
\For{\(i=1\) \textbf{to} \(H\)}
    \For{\(j=1\) \textbf{to} \(W\)}
        \If{\(\mathcal{C}\left(i,j\right) = 1\)}
            \State \(F(i,j) \gets 0\)  \Comment{Set to 0 if contour is present}
        \Else
            \State \(F(i,j) \gets \min\big(F(i-1,j-1), F(i-1,j), F(i-1,j+1), F(i,j-1)\big)+1\)
            \Comment{Update distance using neighbours}  
        \EndIf
    \EndFor
\EndFor
\State Backward pass:
\For{\(i=1\) \textbf{to} \(H-1\)}
    \For{\(j=1\) \textbf{to} \(W-1\)}
        \If{\(\mathcal{C}\left(i,j\right) = 1\)}
            \State \(F(i,j) \gets \min\big(F(i,j), F(i+1,j+1)+1, F(i+1,j)+1, F(i+1,j-1)+1,F(i,j+1)+1\big)\)
            \Comment{Update distance using neighbours}  
        \EndIf
    \EndFor
\EndFor
\end{algorithmic}
\end{algorithm}

\subsection{Functions of the type $g(d)$}
\label{sec:appendix:discussion_distance:study_g}
Different
functions of the distance transform 
are illustrated
in Figure~\ref{fig:appendix:qualitative_dist_results},
while the corresponding quantitative results for depth are shown on Table~\ref{table:appendix:results_functions}.

\subsection{Random Walk}
\label{sec:appendix:discussion_distance:random_walk}
The
random walk
is shown on Algorithm~\ref{alg:appendix:rw},
\AMsup{provided here}
in the 2D case for simplicity.

\begin{algorithm}
\caption{Random Walk}\label{alg:appendix:rw}
\begin{algorithmic}[1]
\Require \(N\) number of steps, \(\varepsilon\) size of one step, \(\{\theta_{i}\}_{i=1}^{k}\)\, is a discrete partition of the unit disk in \(k\) angles.
\State Initialize \( X_0 = \left(0, 0\right)\)
\For{\(i=1\) \textbf{to} \(N\)}
    \State \(\theta = \text{random}\left(\theta_1, ..., \theta_k\right)\) \Comment{choose a random angle}
    \State \(X_i = X_{i-1} + \varepsilon e^{i\theta}\)
\EndFor
\end{algorithmic}
\end{algorithm}
\noindent
Once the random walk is performed, the mapping is done as follows:

\begin{equation}
\begin{aligned}
\text{RW}\left(i\right) = X_{i}
\end{aligned}
\end{equation}

\begin{table}[ht]
\centering
\resizebox{\linewidth}{!}{ 
\renewcommand{\arraystretch}{1.0} 
\small 
\begin{tabular}{ c c c c }
\hline
\textbf{Functions} &  \textcolor{lightblue}{Abs Rel (\(\downarrow\))} &  \textcolor{lightblue}{RMSE log (\(\downarrow\))} &  \textcolor{lightblue}{\(\delta_1\) (\(\uparrow\))} \\
\hline
\(x \mapsto x\) & 0.106 & 0.183 & 0.884 \\
\(x \mapsto \sin(\pi x)\) & 0.105 & 0.182 & 0.884 \\
\(x \mapsto 4x(1 - x)\) & 0.107 & 0.185 & 0.882 \\
\(\text{RW}_{2}\) & 0.105 & 0.181 & 0.885 \\
\(\text{RW}_{3}\) & \textbf{0.104} & \textbf{0.180} & \textbf{0.885} \\
\(\text{RW}_{4}\) & 0.107 & 0.185 & 0.880 \\
\hline
\end{tabular}
}
\caption{Results 
\antoine{on KITTI 2015}
for various functions as defined in Section~\ref{sec:preambule:max_variance}, 
\antoine{and different RW encoding of the distance transform}.
}
\label{table:appendix:results_functions}
\end{table}

One example 
of a random walk function 
is
shown on Figure~\ref{fig:appendix:qualitative_dist_results}. It was performed using \(\varepsilon=0.01\) and a partition number \(k=1000\) for polar and azimuthal angles. Quantitative results for different dimensions of random walks are shown on Table~\ref{table:appendix:results_functions}.

\section{Discussion on the contour}
\label{sec:appendix:discussion_contour}
\subsection{Complementarity Depth - Normal}
\label{sec:appendix:discussion_contour:complementary_depth_normal}
\begin{figure}[t]
\centering
\setlength{\tabcolsep}{0pt}
\renewcommand{\arraystretch}{1.0}
\hspace{0.5cm}\includegraphics[width=0.4\linewidth, height=2cm]{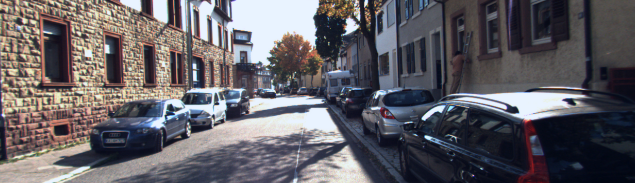} \hspace{0.5cm}\includegraphics[width=0.4\linewidth, height=2cm]{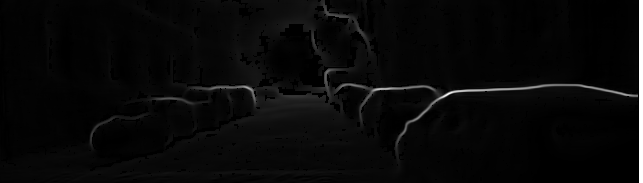}\\
\footnotesize{Image (left) and Laplacian activation of depth (right).}\\
\hspace{0.5cm}\includegraphics[width=0.4\linewidth, height=2cm]{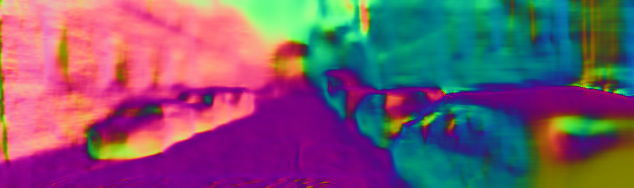} \hspace{0.5cm}\includegraphics[width=0.4\linewidth, height=2cm]{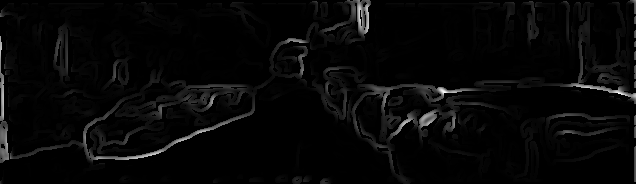}\\
\footnotesize{Normal (left) and gradient activation of normal (right).}
\caption{Illustration of the complementarity of depth and normals.}
\label{fig:appendix:complementary}
\end{figure}

Depth contours and normal to surface contours target different parts of the contour. The depth is better at localizing edge resulting from occlusions, 
\AMsup{i.e.}
between foreground instances and background, 
\AMsup{where a large distance gradient is expected. In contrast,}
the normal is better at 
\AMsup{locating edges related to sharp angle changes,} as shown on Figure~\ref{fig:appendix:complementary}.

\subsection{Comparison to Lego}
\label{sec:appendix:discussion_contour:comparison_lego}
In the Lego method~\cite{yang2018lego}, instead of considering zero-crossing to learn the edges, the authors consider the whole positive part of the second-derivative. This creates coarse edges that are not aligned on the true semantic borders but either towards the exterior or the interior of the instances. Besides, the loss derived from the depth includes a normalization term that poorly 
\AMsup{addresses}
the bias of large distances. Our proposed procedure solves all 
these issues. We give a comparison of qualitative results in Figure~\ref{fig:appendix:qualitative_contour}. We observe that our contour estimations perform better at large distances and with orientation changes. Additionally, the predictions are more tightly aligned with the objects, which is crucial for the effectiveness of our framework.

\subsection{Post-processing}
\label{sec:appendix:discussion_contour:post}
\begin{figure}[t]
\centering
\setlength{\tabcolsep}{0pt}
\renewcommand{\arraystretch}{1.0}
\hspace{0.5cm}\includegraphics[width=0.4\linewidth, height=2cm]{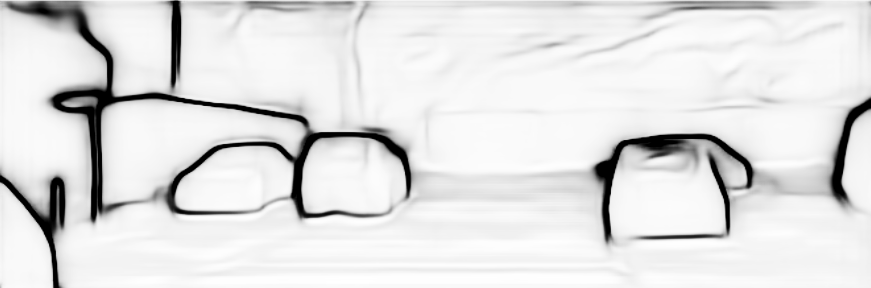} \hspace{0.5cm}\includegraphics[width=0.4\linewidth, height=2cm]{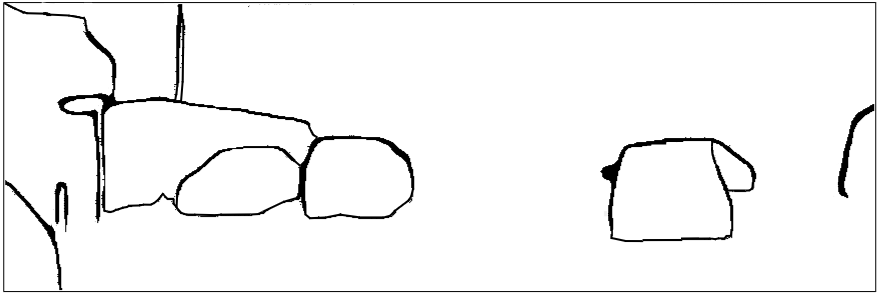}\\
\caption{Output of the edge neural network (left) and contour estimation after post-processing (right)}
\label{fig:appendix:post_processing}
\end{figure}

For post-processing, we first apply hysteresis thresholding to the output of the edge network, using a low threshold of 80 and a high threshold of 100 to obtain \(\text{edge}_{h}\). \\
Next, we perform non-maximum suppression along the gradient direction on the output of the edge network to obtain \(\text{edge}_{n}\). Finally we compute:

\begin{equation} 
    \begin{aligned} 
        \text{edge}_{\text{binary}} =
        \begin{cases}
            1 & \text{if } 0 < \text{edge}_{h} \times \text{edge}_{n}, \\
            0 & \text{otherwise}.
        \end{cases}
    \end{aligned}
\end{equation}
\noindent
The final estimated pre-semantic contour, \(\widehat{\mathcal{C}}(I)\), is derived from  \(\text{edge}_{\text{binary}}\) by applying morphological transformations using OpenCV's "morphologyEx" to fill holes, performing a contour closing procedure, and filtering out small, isolated contours. A post-processing result is shown on Figure~\ref{fig:appendix:post_processing}.

\section{Discussion on the constancy assumption}
\label{sec:appendix:discussion_constancy}

In this experiment, we focus on the constancy assumption, i.e., the fact that any change introduced in image~\(t\) should be reproduced identically in image \(t+1\)
We aim to evaluate the validity of the constancy assumption for our distance transform map and compare it to the deep features of a ResNet-18 pre-trained on ImageNet. To achieve this, we use the KITTI MOTS dataset \cite{Voigtlaender19CVPR_MOTS}, which provides several sequences of images with ground-truth instances, as shown in Figure~\ref{fig:appendix:instance_mots}. We considered a mask with a radius of 3 pixels around the center of the object tracked across the sequence, as illustrated in Figure~\ref{fig:appendix:qualitative_constancy}.

\begin{center}
    \captionsetup{type=figure}
    \captionsetup[subfigure]{labelformat=empty}
    \addtocounter{figure}{-1}
    
    \begin{subfigure}[b]{0.45\textwidth}
        \centering
        \includegraphics[width=\textwidth, height=2.5cm]{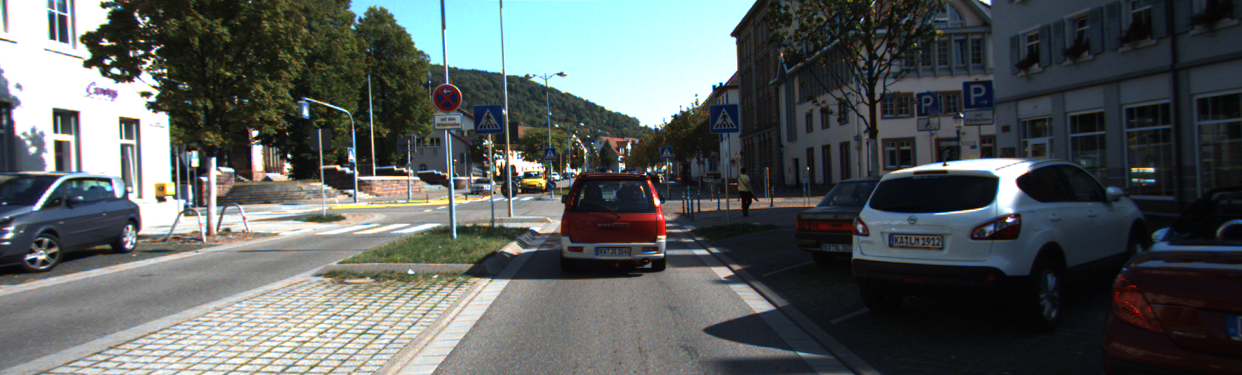}
    \end{subfigure}
    \hfill
    \begin{subfigure}[b]{0.45\textwidth}
        \centering
        \includegraphics[width=\textwidth, height=2.5cm]{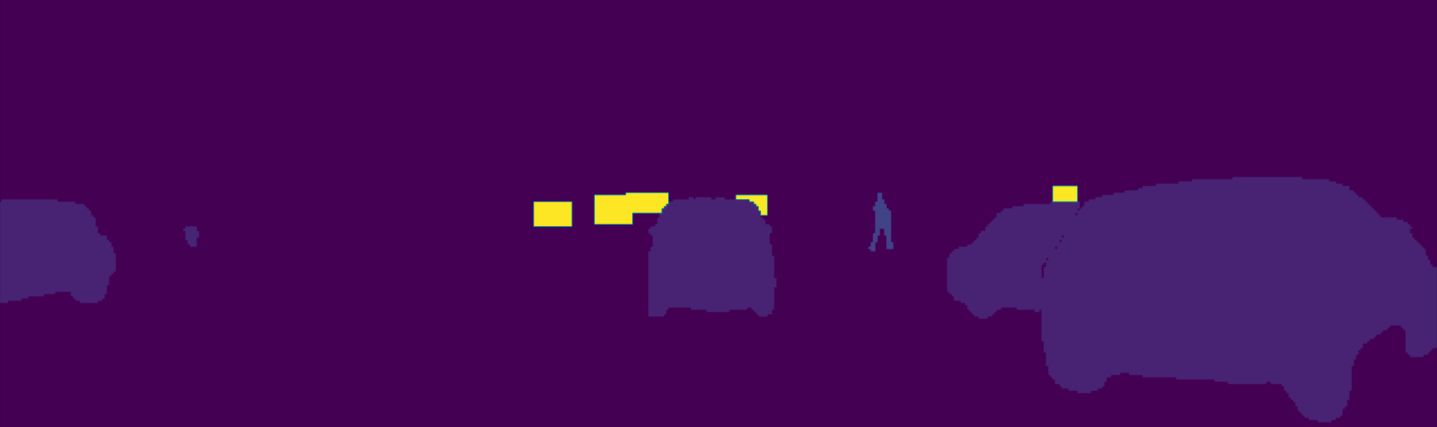}
    \end{subfigure}
    
    \captionof{figure}{One Image of sequence 11 of dataset KITI MOTS~\cite{Voigtlaender19CVPR_MOTS} (above) and the associated ground-truth instance segmentation map (below).}
\label{fig:appendix:instance_mots}
\end{center}

\begin{table}[ht]
\centering
\resizebox{0.65\linewidth}{!}{ 
\renewcommand{\arraystretch}{1.2} 
\setlength{\tabcolsep}{12pt} 
\begin{tabular}{| c | c |}
\hline
\textbf{Sequences} & Object Indexes \\
\hline
4&1002\\
5&1031\\
8&1008\\
10&1000\\
11&1000\\
18&1003\\
20&1012\\
\hline
\end{tabular}
}
\vspace{0.3cm} 
\caption{Sequences of KITTI MOT, with the corresponding object indexes, used for our experiment.}
\label{table:appendix:sequence_constancy}
\end{table}

\noindent
After tracking the position of the mask along the sequence, we computed the variance over time, normalized by the mean over time, for both our distance transform map and the deep feature maps. For the deep features, we considered the output of the four-layer blocks in the encoding part (referred to as layer1, layer2, layer3, and layer4 in PyTorch).
We conducted these experiments on 7 sequences, summarized in Table~\ref{table:appendix:sequence_constancy}, along with their corresponding object IDs. Figure~\ref{fig:constancy_assumption} illustrates the improved constancy achieved when using the distance transform.

\section{Training hyper-parameters}
\label{sec:appendinx:hyper_parameters}
\subsection{ResNet50 encoder}
As a reminder, the loss used to train the edge network is:

\begin{equation*} 
    \begin{aligned} 
        \mathcal{L}^{\text{edge}} &= \sum_{p} \left(\lambda_d w_{d}(p) + \lambda_n w_{n}(p) \right) \left( 1 - E(p) \right) \\
        &  + \lambda_c \mathcal{L}_{c}+ \lambda_e \mathcal{L}_{e} \\ 
        & \text{with} \  \mathcal{L}_{e} = \sum_{p} 
        \antoine{E(p)^2}
        \ \text{to avoid trivial solutions.} 
    \end{aligned}
\end{equation*}
\noindent
The loss weights are set as follows:
\(\lambda_d=0.5\), \(\lambda_n=0.25\), \(\lambda_e=1.0\) and \(\lambda_c=0.001\). Contrastive loss $\mathcal{L}_{c}$ is only introduced after 15 epochs.\\
\\
The loss used to train the depth network is:
\begin{equation*}
    \begin{aligned}
        \mathcal{L}^{\text{depth}} = \lambda_{\text{dist}}\mathcal{L}_{\text{dist}} + \lambda_{\text{photo}}\mathcal{L}_{\text{photo}} + \lambda_{s}\mathcal{L}_{s}
    \end{aligned}
\end{equation*}
\noindent
The weights are set as follows:
\(\lambda_{\text{photo}}=1\), \(\lambda_{\text{dist}}=1\) and \(\lambda_{s}=0.001\).
Normal smoothing is also added to the total loss to better predict edges with \(\lambda_{ns}=0.01\):

\begin{equation*}
    \begin{aligned}
        \mathcal{L}_{ns} = \sum_{p} \left( |\nabla N_{x}(p)| e^{-|\nabla I_{x}(p)|} + |\nabla N_{y}(p)| e^{-|\nabla I_{y}(p)|} \right)
    \end{aligned}
\end{equation*}

\subsection{Dino encoder}
\label{sec:appendix:hyper_parameters:dino}
We also considered a DinoV2~\cite{oquab2023dinov2} encoder following the implementation provided by Facebook Research. We followed the implementation of \cite{yang2024depth} for the depth head. We considered  the large ViT "dinov2\_vitl14" with \(304\)M parameters. 
For this experiment a OneCycle learning rate was chosen for the depth network with \(lr=2.10^{-6}\) for the encoder part and \(lr=10^{-5}\) for the decoder part and a weight decay of \(0.01\). Other network training parameters and hyper-parameters remained unchanged from the original implementation.

\twocolumn[{
\begin{center}
    \captionsetup{type=figure}
    \captionsetup[subfigure]{labelformat=empty}
    \addtocounter{figure}{-1}
    
    \begin{subfigure}[b]{0.24\textwidth}
        \centering
        \includegraphics[width=\textwidth, height=2.5cm]{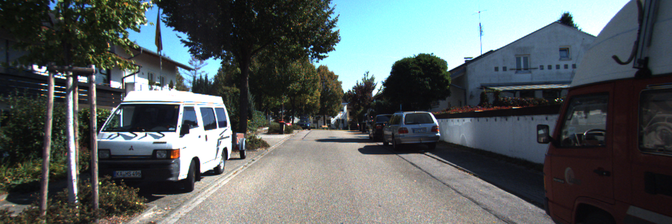}
    \end{subfigure}
    \hfill
    \begin{subfigure}[b]{0.24\textwidth}
        \centering
        \includegraphics[width=\textwidth, height=2.5cm]{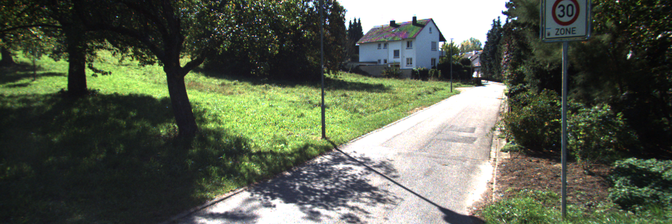}
    \end{subfigure}
    \hfill
    \begin{subfigure}[b]{0.24\textwidth}
        \centering
        \includegraphics[width=\textwidth, height=2.5cm]{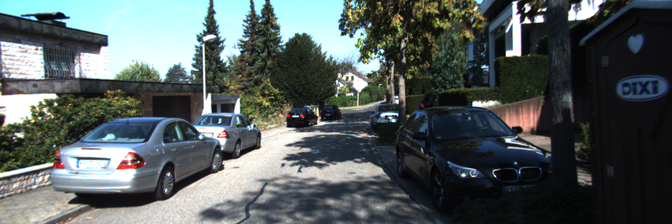}
    \end{subfigure}
    \hfill
    \begin{subfigure}[b]{0.24\textwidth}
        \centering
        \includegraphics[width=\textwidth, height=2.5cm]{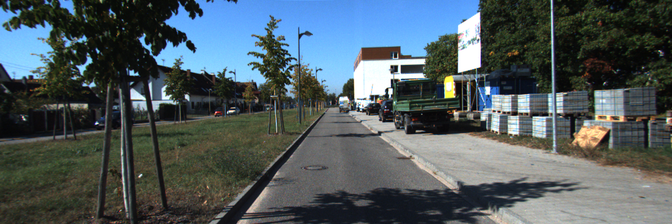}
    \end{subfigure}

    \vspace{0.2cm}  

    \begin{subfigure}[b]{0.24\textwidth}
        \centering
        \includegraphics[width=\textwidth, height=2.5cm]{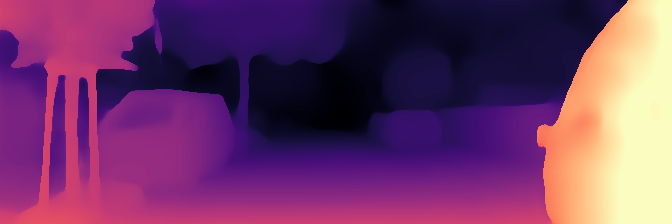}
        \label{figteaser:Orginalimage-anom}
    \end{subfigure}
    \hfill
    \begin{subfigure}[b]{0.24\textwidth}
        \centering
        \includegraphics[width=\textwidth, height=2.5cm]{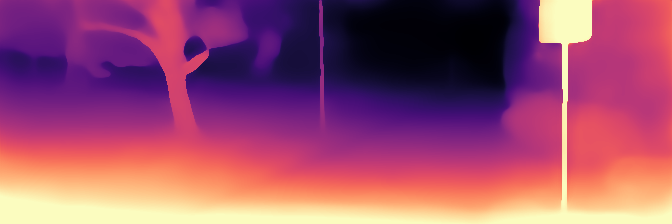}
        \label{figteaser:RobustaGeneration-anom}
    \end{subfigure}
    \hfill
    \begin{subfigure}[b]{0.24\textwidth}
        \centering
        \includegraphics[width=\textwidth, height=2.5cm]{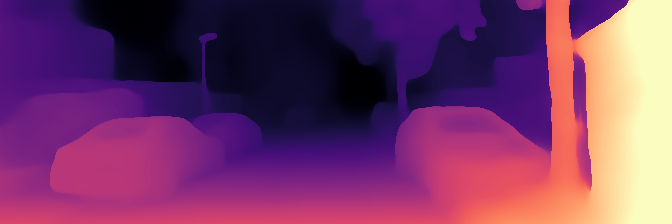}
        \label{figteaser:OasisGeneration-anom}
    \end{subfigure}
    \hfill
    \begin{subfigure}[b]{0.24\textwidth}
        \centering
        \includegraphics[width=\textwidth, height=2.5cm]{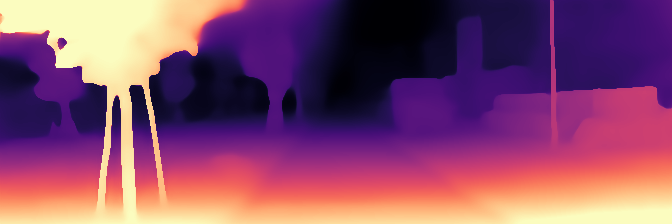}
        \label{figteaser:Label}
    \end{subfigure}

    \vspace{0.2cm}  

    \begin{subfigure}[b]{0.24\textwidth}
        \centering
        \includegraphics[width=\textwidth, height=2.5cm]{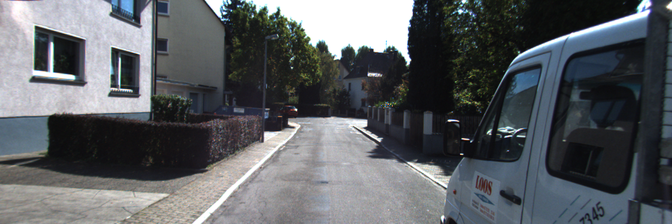}
        \label{figteaser:Orginalimage-anom}
    \end{subfigure}
    \hfill
    \begin{subfigure}[b]{0.24\textwidth}
        \centering
        \includegraphics[width=\textwidth, height=2.5cm]{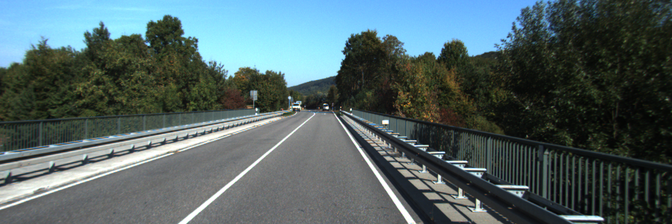}
        \label{figteaser:RobustaGeneration-anom}
    \end{subfigure}
    \hfill
    \begin{subfigure}[b]{0.24\textwidth}
        \centering
        \includegraphics[width=\textwidth, height=2.5cm]{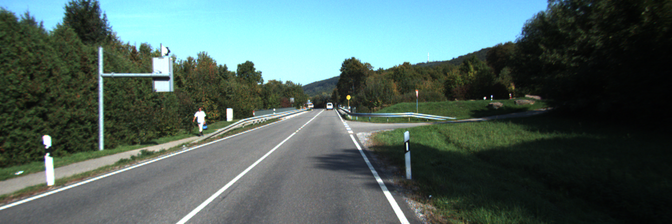}
        \label{figteaser:OasisGeneration-anom}
    \end{subfigure}
    \hfill
    \begin{subfigure}[b]{0.24\textwidth}
        \centering
        \includegraphics[width=\textwidth, height=2.5cm]{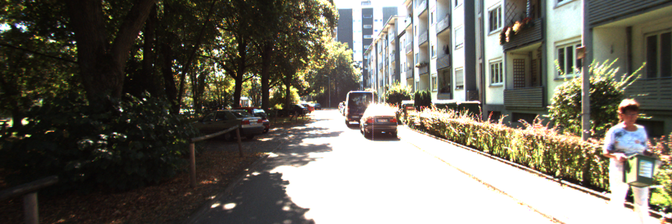}
        \label{figteaser:Label}
    \end{subfigure}

    \vspace{0.2cm}  

    \begin{subfigure}[b]{0.24\textwidth}
        \centering
        \includegraphics[width=\textwidth, height=2.5cm]{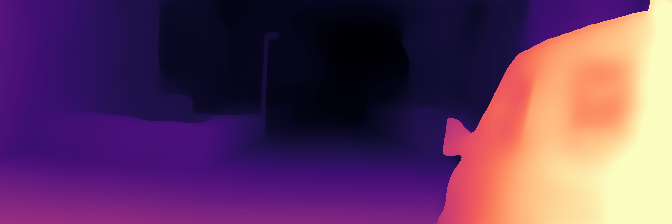}
        \label{figteaser:Orginalimage-anom}
    \end{subfigure}
    \hfill
    \begin{subfigure}[b]{0.24\textwidth}
        \centering
        \includegraphics[width=\textwidth, height=2.5cm]{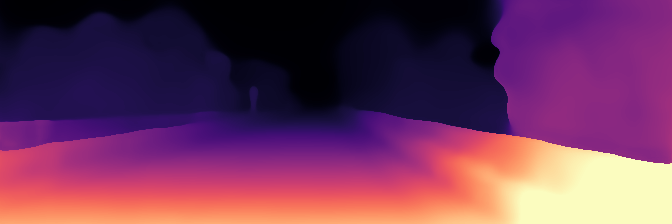}
        \label{figteaser:RobustaGeneration-anom}
    \end{subfigure}
    \hfill
    \begin{subfigure}[b]{0.24\textwidth}
        \centering
        \includegraphics[width=\textwidth, height=2.5cm]{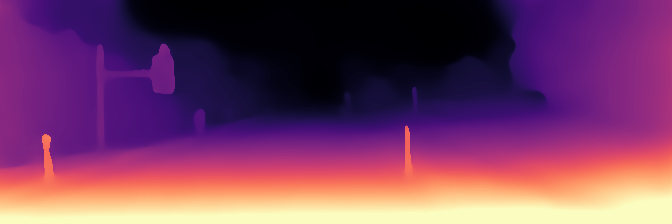}
        \label{figteaser:OasisGeneration-anom}
    \end{subfigure}
    \hfill
    \begin{subfigure}[b]{0.24\textwidth}
        \centering
        \includegraphics[width=\textwidth, height=2.5cm]{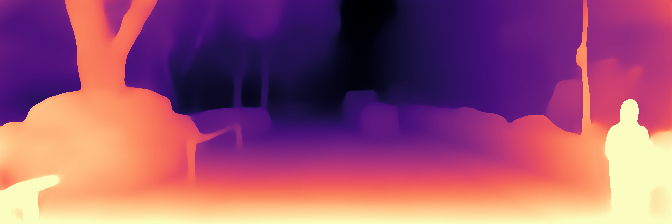}
        \label{figteaser:Label}
    \end{subfigure}

    \vspace{0.2cm}  

    \begin{subfigure}[b]{0.24\textwidth}
        \centering
        \includegraphics[width=\textwidth, height=2.5cm]{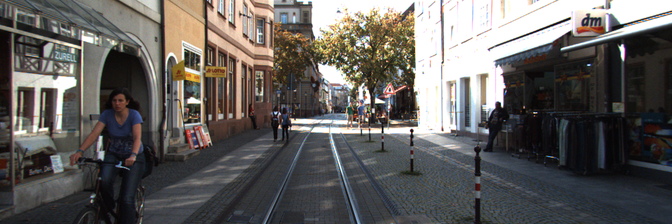}
        \label{figteaser:Orginalimage-anom}
    \end{subfigure}
    \hfill
    \begin{subfigure}[b]{0.24\textwidth}
        \centering
        \includegraphics[width=\textwidth, height=2.5cm]{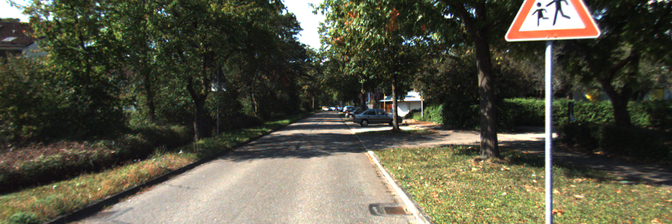}
        \label{figteaser:RobustaGeneration-anom}
    \end{subfigure}
    \hfill
    \begin{subfigure}[b]{0.24\textwidth}
        \centering
        \includegraphics[width=\textwidth, height=2.5cm]{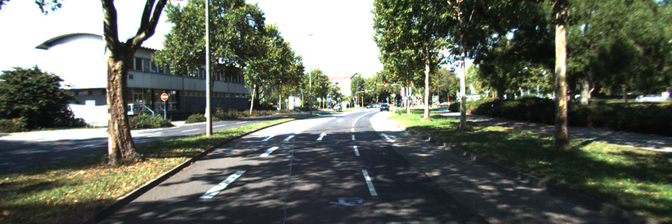}
        \label{figteaser:OasisGeneration-anom}
    \end{subfigure}
    \hfill
    \begin{subfigure}[b]{0.24\textwidth}
        \centering
        \includegraphics[width=\textwidth, height=2.5cm]{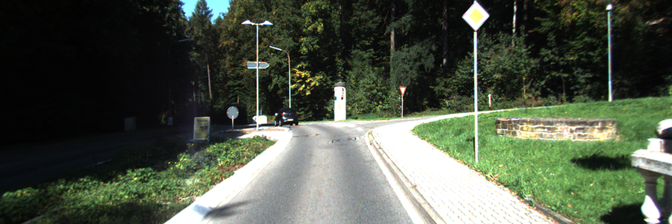}
        \label{figteaser:Label}
    \end{subfigure}

        \vspace{0.2cm}  

    \begin{subfigure}[b]{0.24\textwidth}
        \centering
        \includegraphics[width=\textwidth, height=2.5cm]{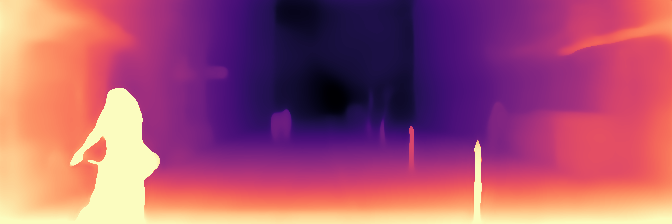}
        \label{figteaser:Orginalimage-anom}
    \end{subfigure}
    \hfill
    \begin{subfigure}[b]{0.24\textwidth}
        \centering
        \includegraphics[width=\textwidth, height=2.5cm]{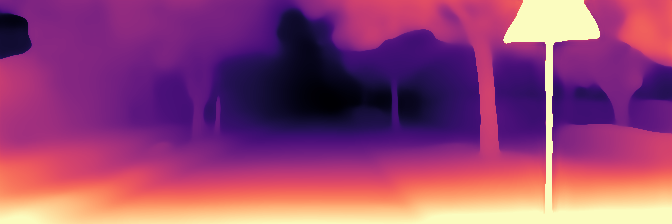}
        \label{figteaser:RobustaGeneration-anom}
    \end{subfigure}
    \hfill
    \begin{subfigure}[b]{0.24\textwidth}
        \centering
        \includegraphics[width=\textwidth, height=2.5cm]{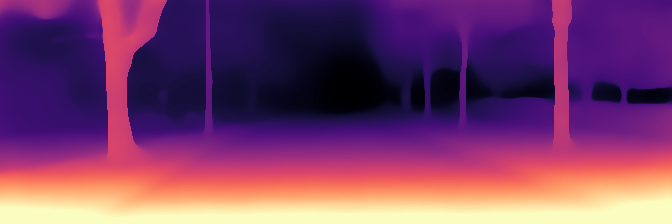}
        \label{figteaser:OasisGeneration-anom}
    \end{subfigure}
    \hfill
    \begin{subfigure}[b]{0.24\textwidth}
        \centering
        \includegraphics[width=\textwidth, height=2.5cm]{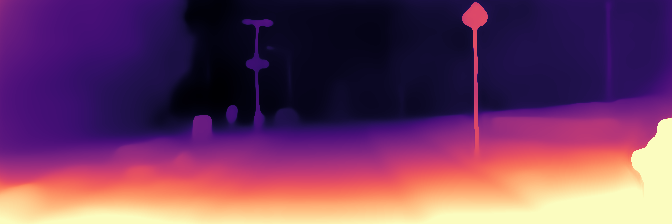}
        \label{figteaser:Label}
    \end{subfigure}
    
    \captionof{figure}{Qualitative results of our method.}
    \label{fig:appendix:qualitative_depth_results}
\end{center}
}]

\twocolumn[{
\begin{center}
    \captionsetup{type=figure}
    \captionsetup[subfigure]{labelformat=empty}
    \addtocounter{figure}{-1}
    
    \begin{subfigure}[b]{0.24\textwidth}
        \centering
        \includegraphics[width=\textwidth, height=2.5cm]{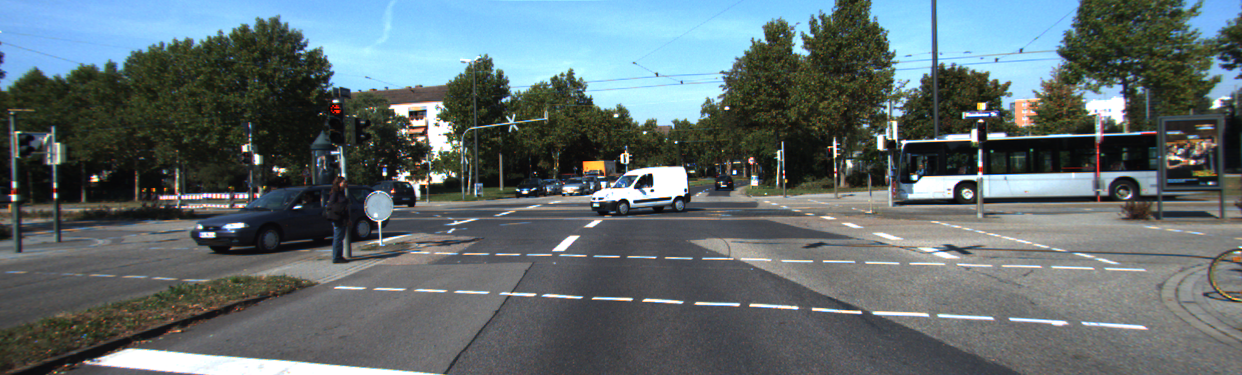}
    \end{subfigure}
    \hfill
    \begin{subfigure}[b]{0.24\textwidth}
        \centering
        \includegraphics[width=\textwidth, height=2.5cm]{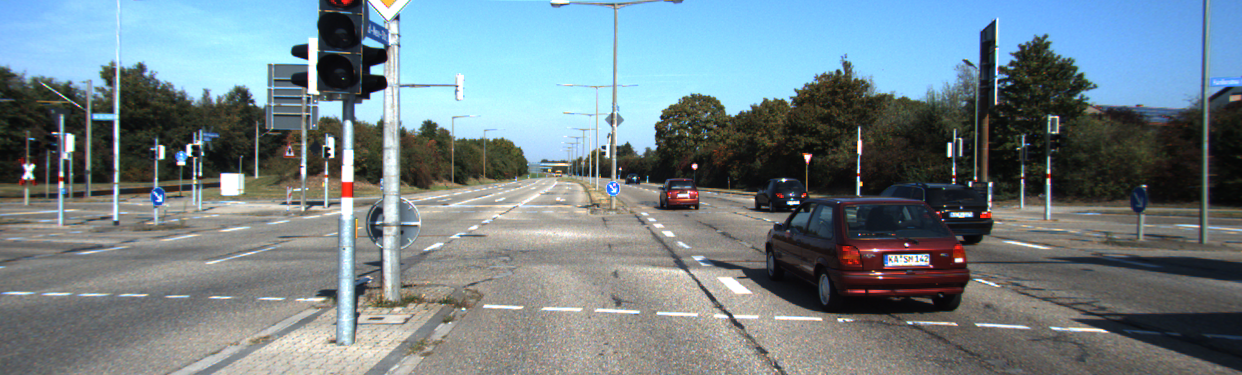}
    \end{subfigure}
    \hfill
    \begin{subfigure}[b]{0.24\textwidth}
        \centering
        \includegraphics[width=\textwidth, height=2.5cm]{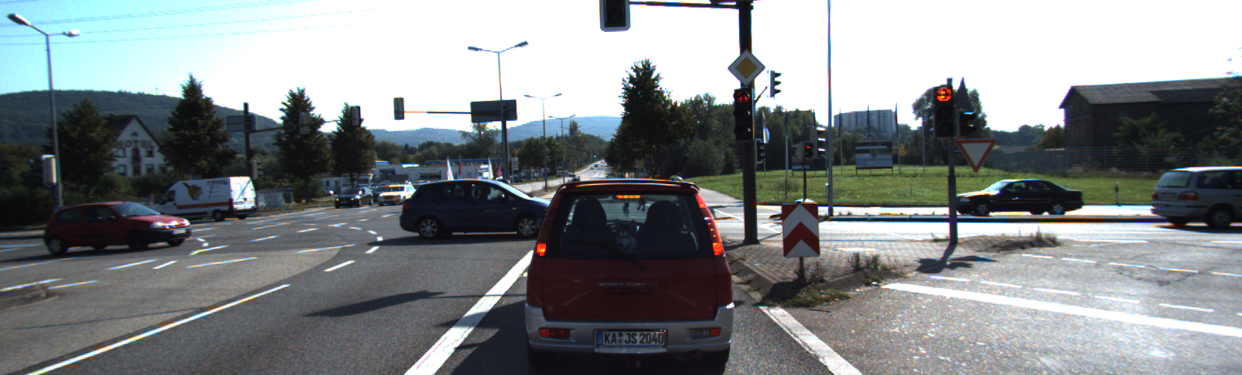}
    \end{subfigure}
    \hfill
    \begin{subfigure}[b]{0.24\textwidth}
        \centering
        \includegraphics[width=\textwidth, height=2.5cm]{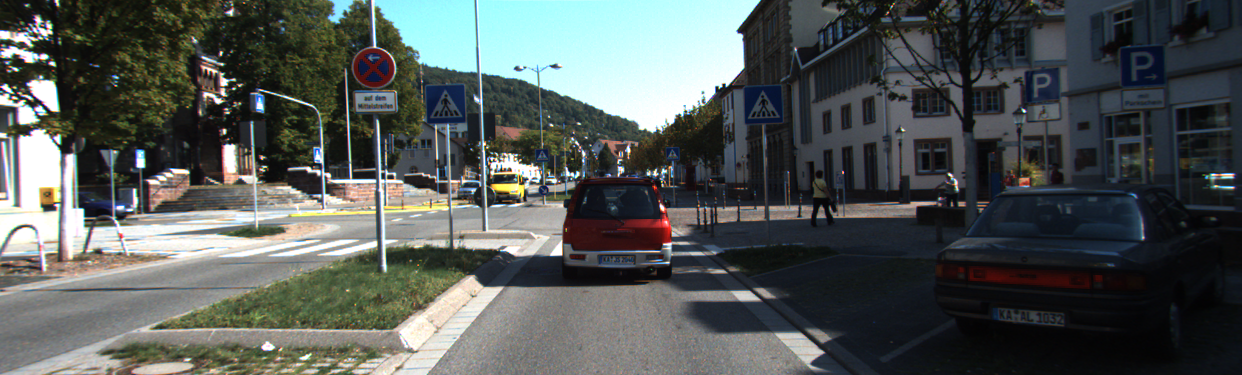}
    \end{subfigure}

    \vspace{0.2cm}  

    \begin{subfigure}[b]{0.24\textwidth}
        \centering
        \includegraphics[width=\textwidth, height=2.5cm]{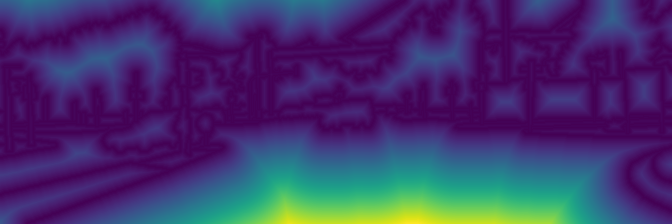}
    \end{subfigure}
    \hfill
    \begin{subfigure}[b]{0.24\textwidth}
        \centering
        \includegraphics[width=\textwidth, height=2.5cm]{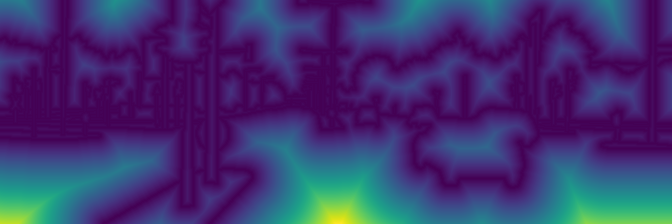}
    \end{subfigure}
    \hfill
    \begin{subfigure}[b]{0.24\textwidth}
        \centering
        \includegraphics[width=\textwidth, height=2.5cm]{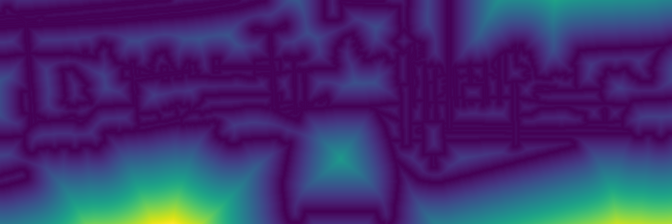}
    \end{subfigure}
    \hfill
    \begin{subfigure}[b]{0.24\textwidth}
        \centering
        \includegraphics[width=\textwidth, height=2.5cm]{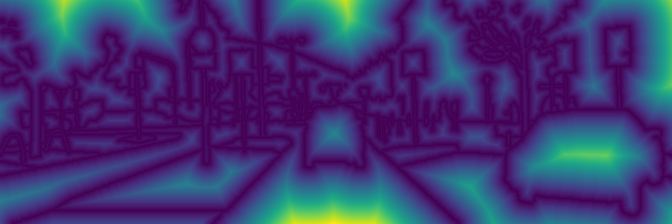}
    \end{subfigure}

    \vspace{0.2cm}  
    
    \begin{subfigure}[b]{0.24\textwidth}
        \centering
        \includegraphics[width=\textwidth, height=2.5cm]{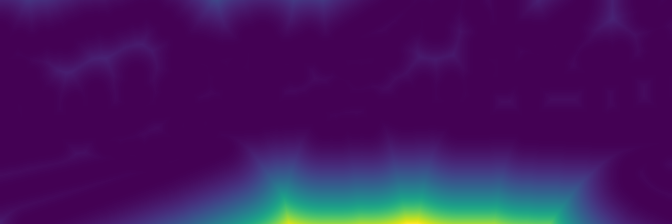}
    \end{subfigure}
    \hfill
    \begin{subfigure}[b]{0.24\textwidth}
        \centering
        \includegraphics[width=\textwidth, height=2.5cm]{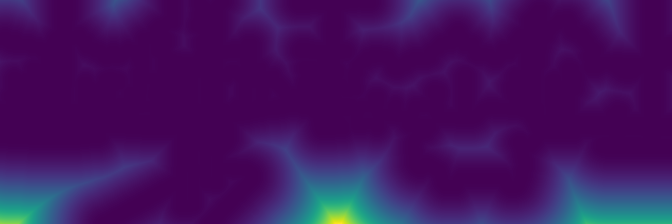}
    \end{subfigure}
    \hfill
    \begin{subfigure}[b]{0.24\textwidth}
        \centering
        \includegraphics[width=\textwidth, height=2.5cm]{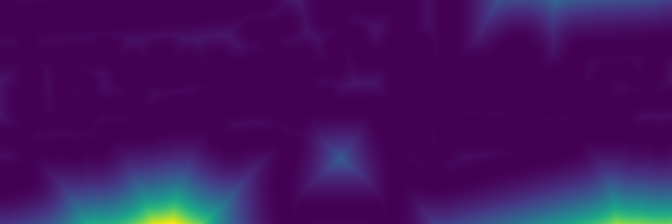}
    \end{subfigure}
    \hfill
    \begin{subfigure}[b]{0.24\textwidth}
        \centering
        \includegraphics[width=\textwidth, height=2.5cm]{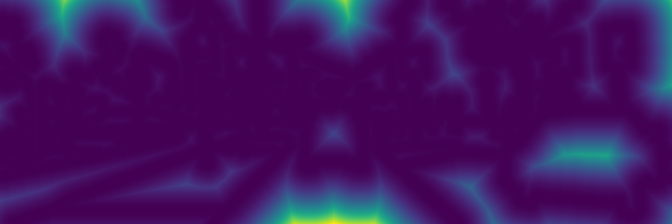}
    \end{subfigure}

    \vspace{0.2cm}  
    
    \begin{subfigure}[b]{0.24\textwidth}
        \centering
        \includegraphics[width=\textwidth, height=2.5cm]{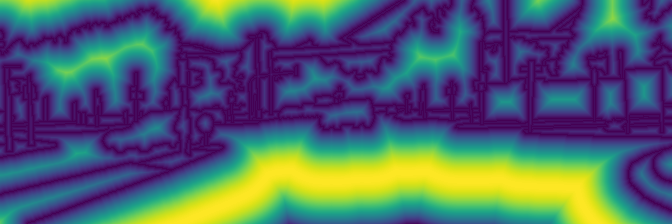}
    \end{subfigure}
    \hfill
    \begin{subfigure}[b]{0.24\textwidth}
        \centering
        \includegraphics[width=\textwidth, height=2.5cm]{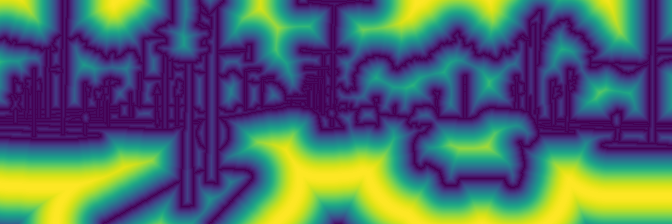}
    \end{subfigure}
    \hfill
    \begin{subfigure}[b]{0.24\textwidth}
        \centering
        \includegraphics[width=\textwidth, height=2.5cm]{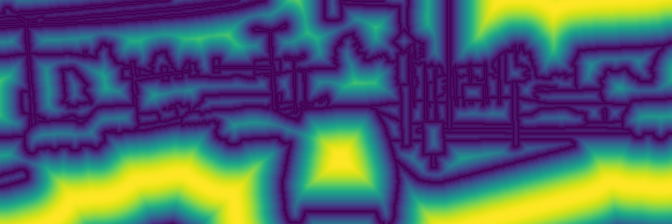}
    \end{subfigure}
    \hfill
    \begin{subfigure}[b]{0.24\textwidth}
        \centering
        \includegraphics[width=\textwidth, height=2.5cm]{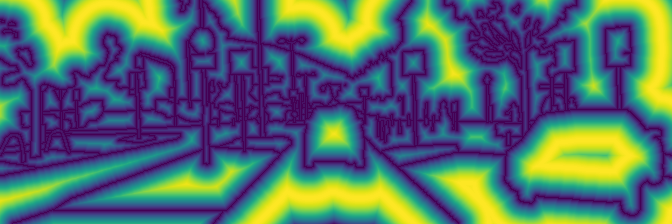}
    \end{subfigure}

    \vspace{0.2cm}  

        \begin{subfigure}[b]{0.24\textwidth}
        \centering
        \includegraphics[width=\textwidth, height=2.5cm]{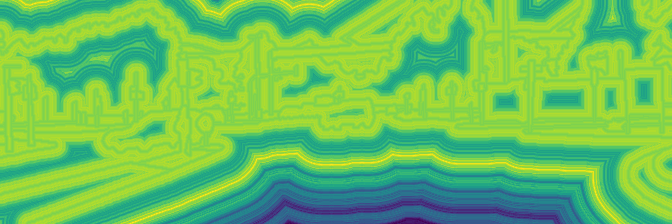}
    \end{subfigure}
    \hfill
    \begin{subfigure}[b]{0.24\textwidth}
        \centering
        \includegraphics[width=\textwidth, height=2.5cm]{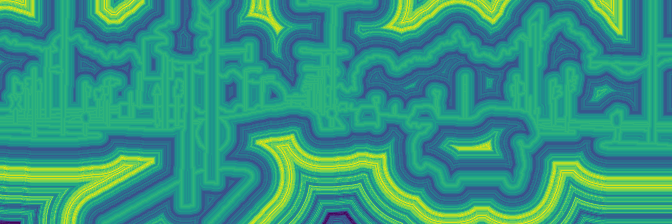}
    \end{subfigure}
    \hfill
    \begin{subfigure}[b]{0.24\textwidth}
        \centering
        \includegraphics[width=\textwidth, height=2.5cm]{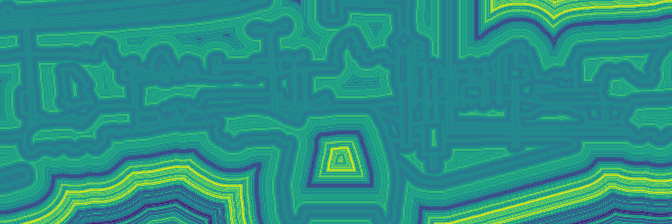}
    \end{subfigure}
    \hfill
    \begin{subfigure}[b]{0.24\textwidth}
        \centering
        \includegraphics[width=\textwidth, height=2.5cm]{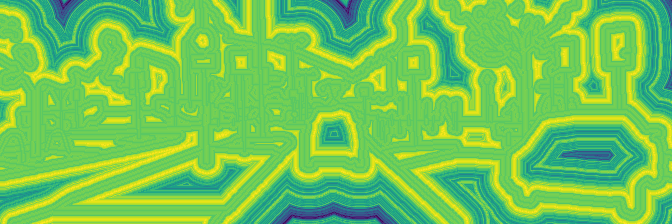}
    \end{subfigure}


    
    \captionof{figure}{Qualitative results of 
    \AMsup{different}
    functions of the distance transform. First row: \(x
    \AMsup{~\mapsto}
    x\). Second row: \(x
    \AMsup{~\mapsto}
    x^2\). Third 
    \AMsup{row:}
    \(x
    \AMsup{~\mapsto}
    \sin(\pi x)\). Last row: 
    first dimension of \(\text{RW}_{3}\)}
    \label{fig:appendix:qualitative_dist_results}
\end{center}
}]

\twocolumn[{
\begin{center}
    \captionsetup{type=figure}
    \captionsetup[subfigure]{labelformat=empty}
    \addtocounter{figure}{-1}
    
    \begin{subfigure}[b]{0.32\textwidth}
        \centering
        \includegraphics[width=\textwidth, height=2.5cm]{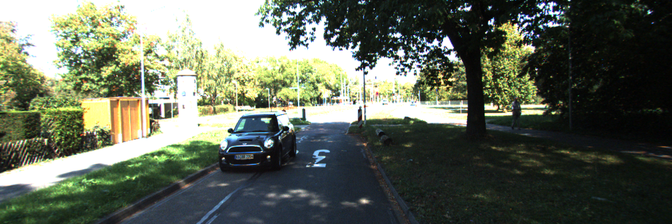}
    \end{subfigure}
    \begin{subfigure}[b]{0.32\textwidth}
        \centering
        \includegraphics[width=\textwidth, height=2.5cm]{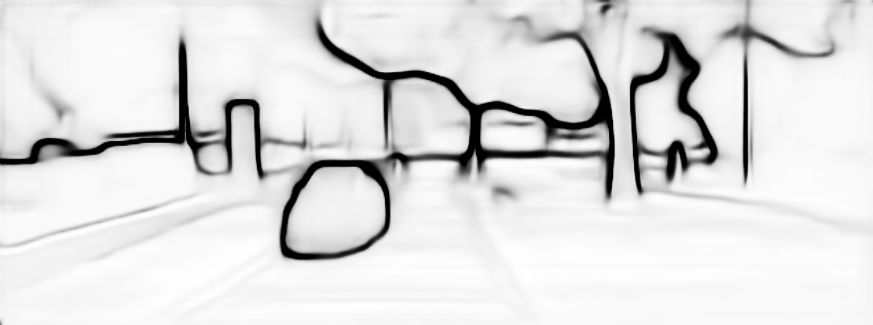}
    \end{subfigure}
    \begin{subfigure}[b]{0.32\textwidth}
        \centering
        \includegraphics[width=\textwidth, height=2.5cm]{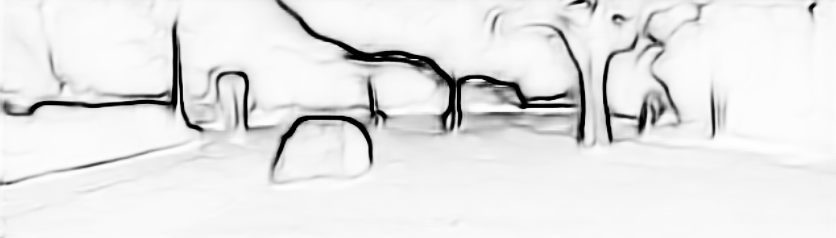}
    \end{subfigure}

    \vspace{0.2cm}  

    \begin{subfigure}[b]{0.32\textwidth}
        \centering
        \includegraphics[width=\textwidth, height=2.5cm]{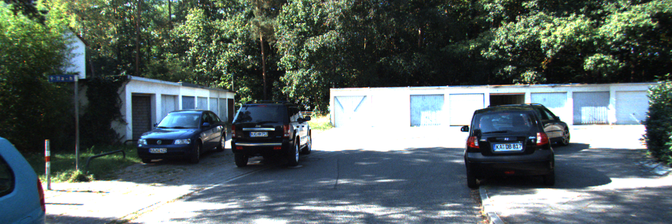}
        \label{figteaser:Orginalimage-anom}
    \end{subfigure}    
    \begin{subfigure}[b]{0.32\textwidth}
        \centering
        \includegraphics[width=\textwidth, height=2.5cm]{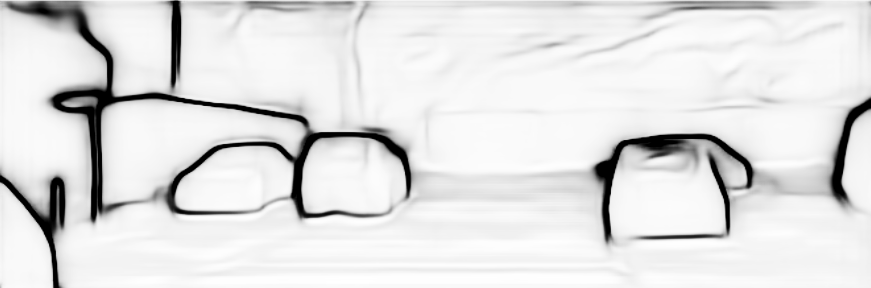}
        \label{figteaser:Orginalimage-anom}
    \end{subfigure}
    \begin{subfigure}[b]{0.32\textwidth}
        \centering
        \includegraphics[width=\textwidth, height=2.5cm]{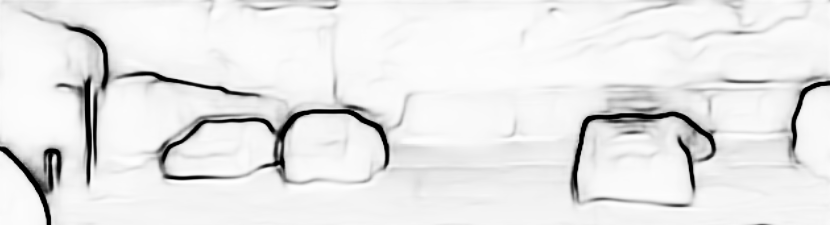}
        \label{figteaser:RobustaGeneration-anom}
    \end{subfigure}

    \vspace{0.2cm}  

    \begin{subfigure}[b]{0.32\textwidth}
        \centering
        \includegraphics[width=\textwidth, height=2.5cm]{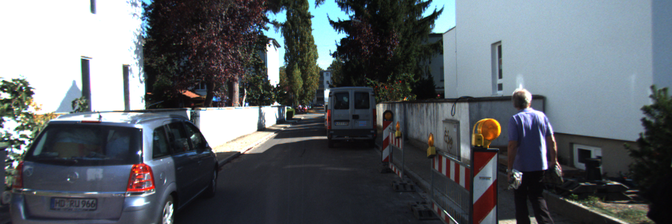}
        \label{figteaser:Orginalimage-anom}
    \end{subfigure}
    \begin{subfigure}[b]{0.32\textwidth}
        \centering
        \includegraphics[width=\textwidth, height=2.5cm]{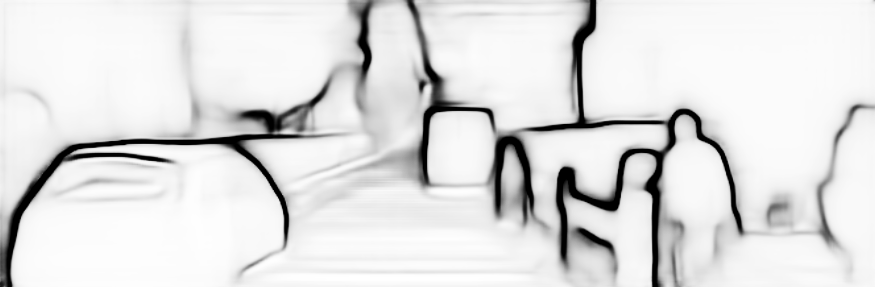}
        \label{figteaser:Orginalimage-anom}
    \end{subfigure}
    \begin{subfigure}[b]{0.32\textwidth}
        \centering
        \includegraphics[width=\textwidth, height=2.5cm]{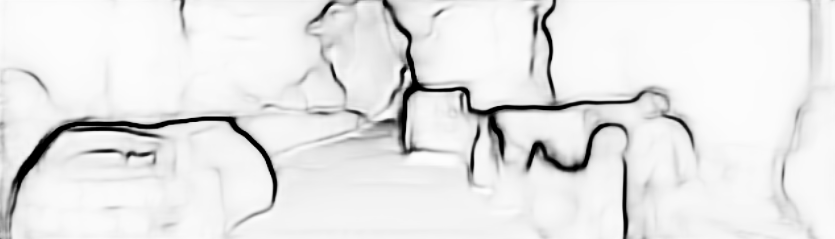}
        \label{figteaser:RobustaGeneration-anom}
    \end{subfigure}

    \vspace{0.2cm}
    
    \begin{subfigure}[b]{0.32\textwidth}
        \centering
        \includegraphics[width=\textwidth, height=2.5cm]{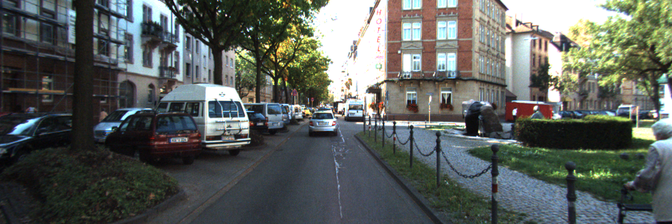}
        \label{figteaser:Orginalimage-anom}
    \end{subfigure}
    \begin{subfigure}[b]{0.32\textwidth}
        \centering
        \includegraphics[width=\textwidth, height=2.5cm]{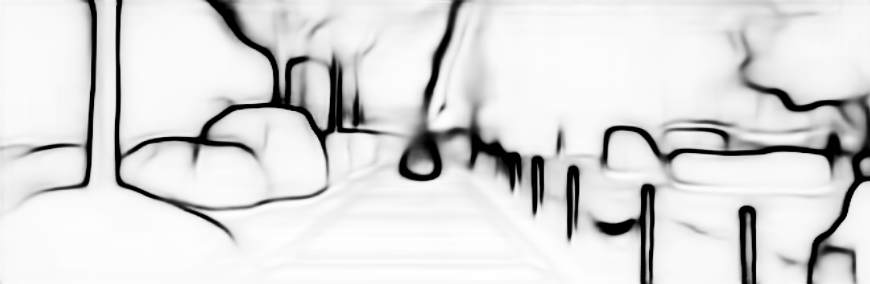}
        \label{figteaser:Orginalimage-anom}
    \end{subfigure}
    \begin{subfigure}[b]{0.32\textwidth}
        \centering
        \includegraphics[width=\textwidth, height=2.5cm]{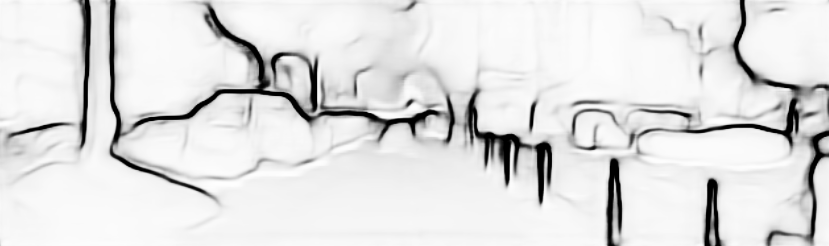}
        \label{figteaser:RobustaGeneration-anom}
    \end{subfigure}

    \vspace{0.2cm}  
        \begin{subfigure}[b]{0.32\textwidth}
        \centering
        \includegraphics[width=\textwidth, height=2.5cm]{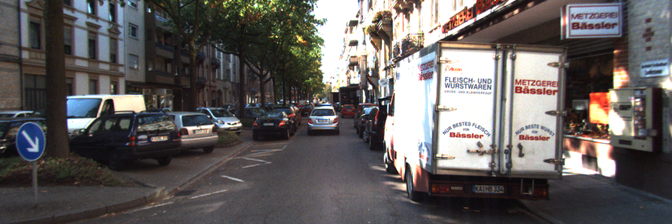}
        \label{figteaser:Orginalimage-anom}
    \end{subfigure}
        \begin{subfigure}[b]{0.32\textwidth}
        \centering
        \includegraphics[width=\textwidth, height=2.5cm]{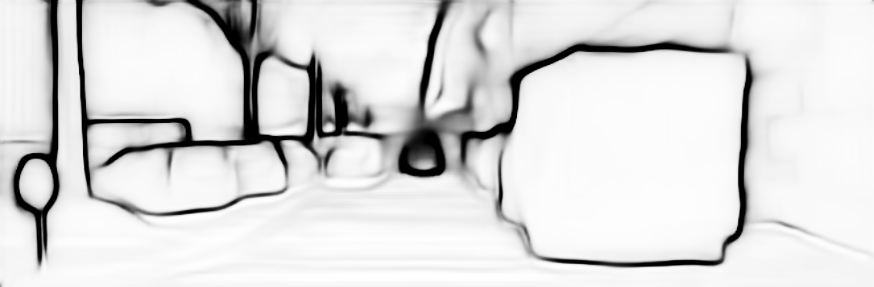}
        \label{figteaser:Orginalimage-anom}
    \end{subfigure}
    \begin{subfigure}[b]{0.32\textwidth}
        \centering
        \includegraphics[width=\textwidth, height=2.5cm]{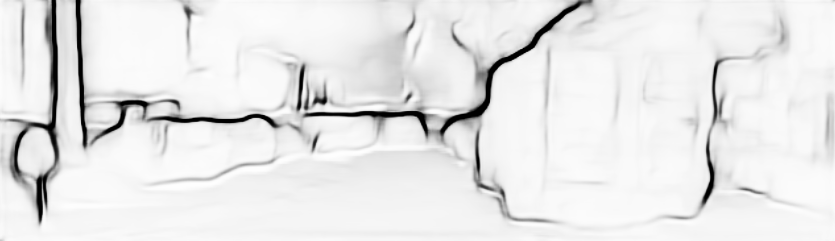}
        \label{figteaser:RobustaGeneration-anom}
    \end{subfigure}


    \captionof{figure}{Contour estimation results for our method (left) and Lego (right) are shown. The results presented here are before post-processing. The Lego method was re-implemented by us, with the rest of our approach remaining unchanged.}
    \label{fig:appendix:qualitative_contour}
\end{center}
}]

\twocolumn[{
\begin{center}
    \captionsetup{type=figure}
    \captionsetup[subfigure]{labelformat=empty}
    \addtocounter{figure}{-1}
    
    \begin{subfigure}[b]{0.32\textwidth}
        \centering
        \includegraphics[width=\textwidth, height=2.5cm]{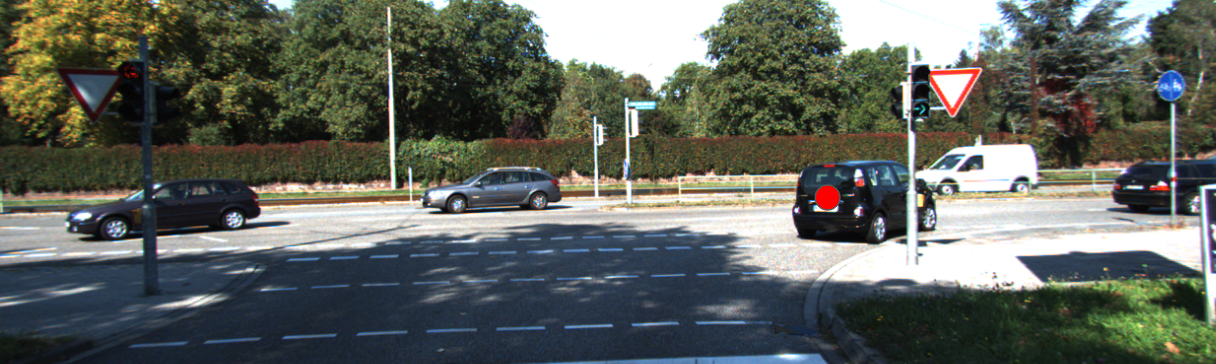}
    \end{subfigure}
    \hfill
    \begin{subfigure}[b]{0.32\textwidth}
        \centering
        \includegraphics[width=\textwidth, height=2.5cm]{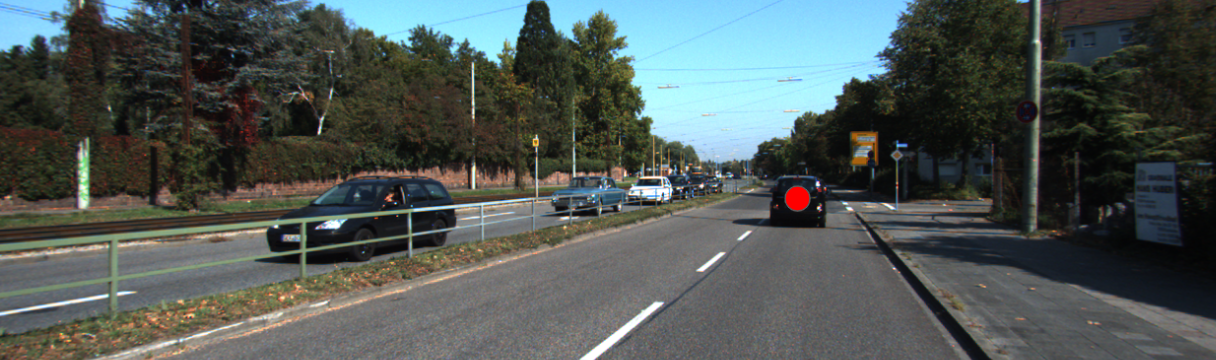}
    \end{subfigure}
    \hfill
    \begin{subfigure}[b]{0.32\textwidth}
        \centering
        \includegraphics[width=\textwidth, height=2.5cm]{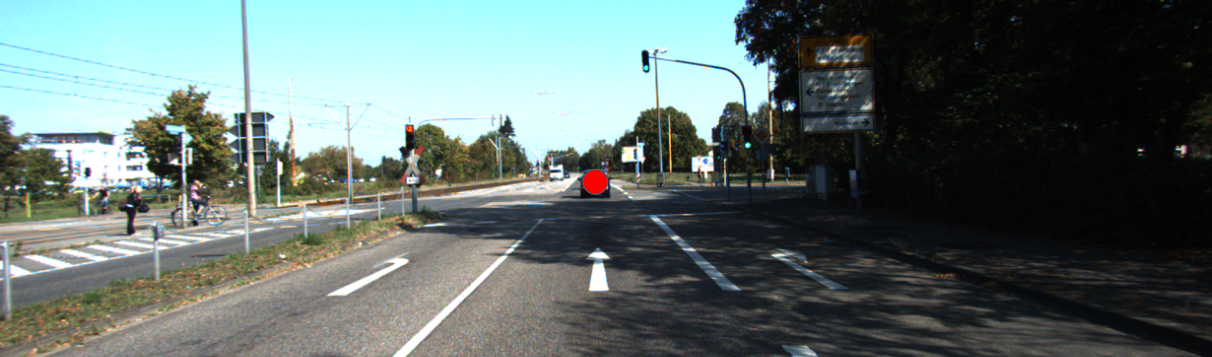}
    \end{subfigure}

    \vspace{0.2cm}  

    \begin{subfigure}[b]{0.32\textwidth}
        \centering
        \includegraphics[width=\textwidth, height=2.5cm]{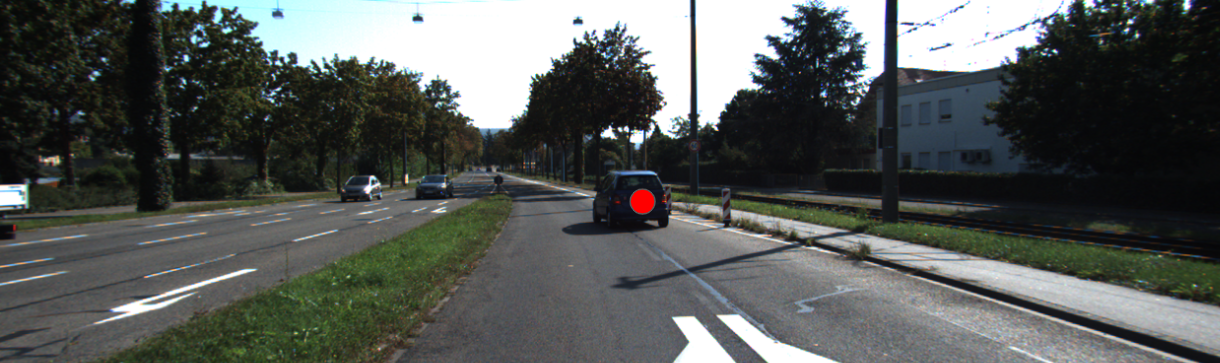}
        \label{figteaser:Orginalimage-anom}
    \end{subfigure}
    \hfill
    \begin{subfigure}[b]{0.32\textwidth}
        \centering
        \includegraphics[width=\textwidth, height=2.5cm]{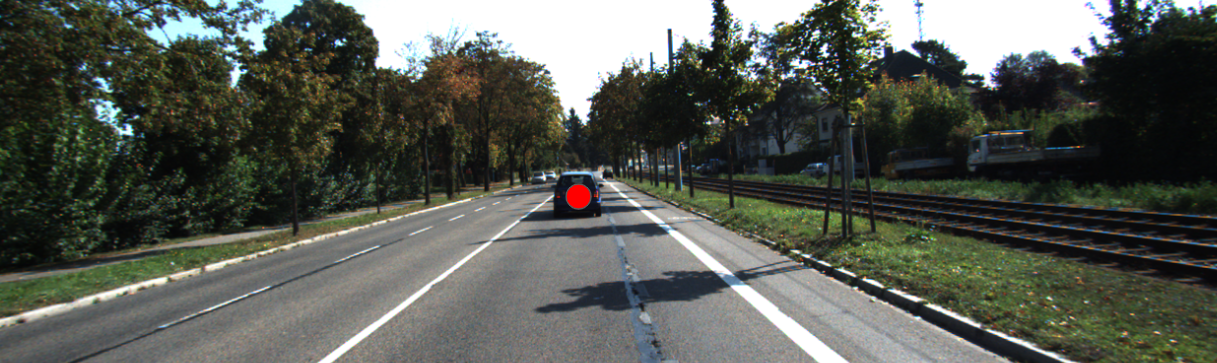}
        \label{figteaser:RobustaGeneration-anom}
    \end{subfigure}
    \hfill
    \begin{subfigure}[b]{0.32\textwidth}
        \centering
        \includegraphics[width=\textwidth, height=2.5cm]{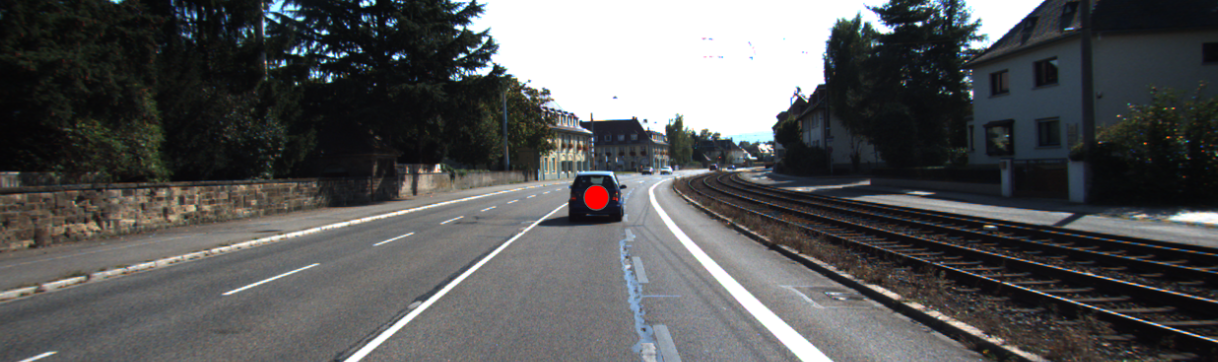}
        \label{figteaser:OasisGeneration-anom}
    \end{subfigure}

    \vspace{0.2cm}  

    \begin{subfigure}[b]{0.32\textwidth}
        \centering
        \includegraphics[width=\textwidth, height=2.5cm]{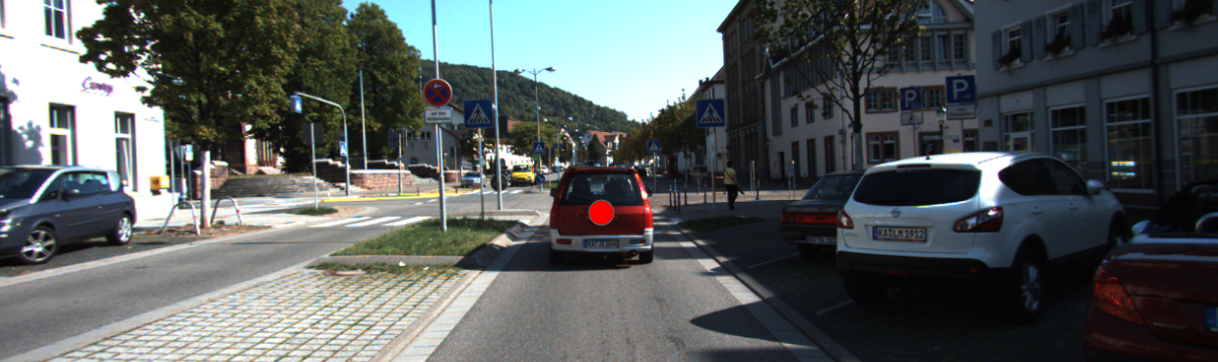}
        \label{figteaser:Orginalimage-anom}
    \end{subfigure}
    \hfill
    \begin{subfigure}[b]{0.32\textwidth}
        \centering
        \includegraphics[width=\textwidth, height=2.5cm]{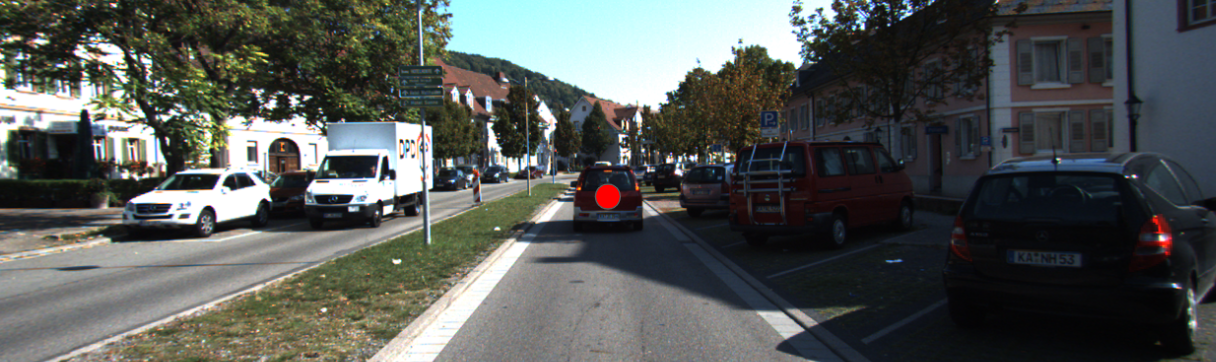}
        \label{figteaser:RobustaGeneration-anom}
    \end{subfigure}
    \hfill
    \begin{subfigure}[b]{0.32\textwidth}
        \centering
        \includegraphics[width=\textwidth, height=2.5cm]{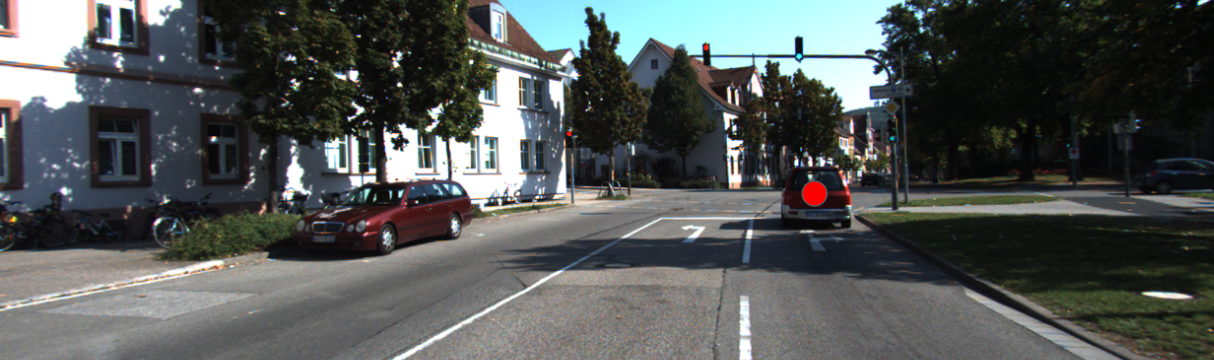}
        \label{figteaser:OasisGeneration-anom}
    \end{subfigure}
    
    \captionof{figure}{Images of KITI MOTS\cite{Voigtlaender19CVPR_MOTS}. Each row shows a red point on the object tracked along the sequence. First row is sequence 4, second row is sequence 10 and last row is sequence 11. Timeline is arranged from left to right. Red circle is the point tracked and considered for our experiments}
    \label{fig:appendix:qualitative_constancy}
\end{center}
}]


\end{document}